%% file: main.tex
\documentclass[11pt]{article}
\usepackage[T1]{fontenc}
\usepackage{lmodern}
\usepackage[margin=1in]{geometry}
\usepackage{amsmath,amssymb}
\usepackage{graphicx}
\usepackage{booktabs}
\usepackage[numbers]{natbib}
\usepackage[hidelinks]{hyperref}
\graphicspath{{figures/}}

\title{Instruction Capacity: Behavior and Degradation of Language Models under Controlled Load}
\author{Hennadii Kutomanov\\
  Independent researcher\\
  \texttt{kutomanov.hennadii@gmail.com}}
\date{}

\begin{document}
\maketitle
\begin{abstract}
\input{sections/00-abstract}
\end{abstract}

\input{sections/01-introduction}
\input{sections/02-related-work}
\input{sections/03-instrument}
\input{sections/04-analysis}
\input{sections/04b-results}

\input{sections/06-outlook}
\input{sections/07-conclusion}
\input{sections/08-availability}
\input{sections/09-acknowledgments}

\bibliographystyle{plainnat}
\bibliography{references}

\appendix
\input{sections/90-appendix}
\end{document}

%% file: sections/00-abstract.tex
% !TEX root = ../main.tex
The quality with which language models follow instructions falls as the load grows, yet there is
still no quantity that expresses the limit of a model in units of load with an uncertainty. We
propose an instrument in which the load consists of rules that replace single-token words in a
text without meaning: such an instruction is atomic, the rules are independent by the construction
of the task, and the outcome of every application of a rule is determined by a computable
reference answer. Every occurrence of a word to be replaced becomes a Bernoulli trial with a known
position in the answer and a known rule number. The statistical model first measures the shape of
the dependence of the probability of a right replacement on the text length, the word position,
the replacement density, the number of rules and the rule number, then describes it by smooth laws
and restores a reference level that does not depend on the choice of the coding. The model yields
the instruction capacity---the number of rules held at a given share of right
replacements---with an uncertainty. A measurement of GPT-4o mini (2958 requests, 642\,430 trials)
gave regular laws for every load factor and for the collapse of the answer; the instruction
capacity was about 37--38 rules at a share of right replacements of 0.95, and about 18--19 rules at
the full replacement density. The code of the instrument, the measurement data and the passport
are open, and every number of the passport is recomputed from the published points. The passport
of a model serves as the calibration of a future meter of the instruction load of prompts.

%% file: sections/01-introduction.tex
% !TEX root = ../main.tex
\section{Introduction}\label{sec:introduction}

Language models follow complex instructions, but this ability is limited. As the number of
simultaneous conditions grows, a model begins to lose some of them: it skips individual
requirements, mixes up correspondences, and continues a plausible text that no longer answers the
statement of the task. The phenomenon is well known and has long been studied: the quality of
execution falls with the number of instructions and with the length of the input and of the
answer, and it depends on the place of an instruction in the prompt
(Section~\ref{sec:related-work}). Yet there is still no quantity that would express the limit of a
model in units of load, carry an uncertainty and be measured in the same way for all models.

The reason lies in the way the load is set. In existing works the instructions differ in nature
and difficulty, share one coherent text or are checked by a judge model, so the number of
instructions is confounded with the composition of the set and with the growing difficulty of the
task itself (Section~\ref{sec:methodology}). For the number of instructions to serve as a measure
of load, one needs an instruction that is atomic, independent of the others, homogeneous, checked
by a program, and that allows every application to be observed at its place in the answer. None of
the approaches reviewed satisfies these requirements at once.

In the present work this instruction is a rule that replaces one single-token word with another in
a text without meaning. A rule admits exactly one right outcome at every place of its application,
the rules are not linked to each other by the construction of the task, and the reference answer
is computed by a program. One rule becomes a conventional unit of load, and every occurrence of a
word to be replaced becomes a Bernoulli trial with a known position in the answer and a known
number of the rule in the list. On these data we build a statistical model that first measures the
shape of the dependence of the probability of a right replacement on every load axis and then
describes it by a smooth law. The model yields the instruction capacity---the number of rules a
model holds at a given share of right replacements---with an uncertainty, and together with it
the laws by which the quality of execution falls with every kind of load.

The contributions of the work are as follows:
\begin{itemize}
\item an instrument with the simplest atomic instruction---the replacement of a single-token word:
  a task is generated from three seeds and reproduced byte for byte, the outcome of every trial is
  determined by a computable reference answer, and the answer is checked line by line,
  distinguishing three kinds of failure (Section~\ref{sec:instrument});
\item a statistical model without a common level that depends on the choice of reference bins: the
  shape of the contributions of the axes is measured by bins and described by smooth laws, and an
  approximation shift correction restores a reference level that does not depend on the coding
  (Sections~\ref{sec:reference} and~\ref{sec:bias});
\item a definition of the instruction capacity as the solution of an inverse problem, with an
  uncertainty computed by the delta method and with the spread of request difficulty taken into
  account (Section~\ref{sec:quantities});
\item the passport of GPT-4o mini: the laws of text length and word position, replacement density,
  number of rules and rule number (Section~\ref{sec:laws}), the laws of the rate and the place of
  the collapse of the answer (Section~\ref{sec:collapse}), and the instruction, fatigue and
  guaranteed instruction capacities (Section~\ref{sec:passport});
\item an open public part of the instrument from which every number of the passport is recomputed
  from the published points (Section~\ref{sec:availability}).
\end{itemize}

The value of the work is twofold. For the study of language model behavior it provides degradation
laws with parameters and uncertainties that can be compared between models and between
generations of models. For practice it lays the foundation of a meter: the passport of a model
serves as a calibration with which the instruction load of an arbitrary prompt can be expressed in
the same units as the capacity of the model (Section~\ref{sec:outlook}).

The paper is organized as follows. Section~\ref{sec:related-work} describes the history of research
on the behavior of language models under load and evaluates the approaches to measuring it.
Section~\ref{sec:instrument} describes the instrument, Section~\ref{sec:analysis} the statistical
model, and Section~\ref{sec:results} the measurement and the passport of GPT-4o mini.
Section~\ref{sec:outlook} is devoted to perspectives, Section~\ref{sec:conclusion} sums up, and
Section~\ref{sec:availability} describes access to the code and data.

%% file: sections/02-related-work.tex
% !TEX root = ../main.tex
\section{Related Work}\label{sec:related-work}

\subsection{History of research on language model behavior under controlled load}\label{sec:history}

The notion that the ability to hold and process information is limited, and that this limit can be
measured, came from cognitive psychology. \citet{miller1956magical} reduced experimental data to an
estimate: in short-term memory tasks a person holds about seven meaningful units---``the magical
number seven, plus or minus two''. Miller himself paid more attention to how grouping elements into
meaningful units extends the memory span, and regarded the number seven rather as a coincidence
\citep{cowan2001magical}. More than forty years later \citet{cowan2001magical} brought together an
extensive body of data and showed that the true limit is smaller: the central capacity of
short-term memory is on average about four units, within the range from three to five. Thus the
capacity of short-term memory became a quantity that is measured, not only described.

The second line, without which the very subject of this study would not exist, is the turning of
instruction following into a property of a language model. \citet{wei2021flan} showed that
instruction tuning---fine-tuning a model on a set of tasks, each of which is described by an
instruction in natural language---sharply improves zero-shot task solving: the 137-billion-parameter
model FLAN, fine-tuned on more than 60 tasks, surpassed the 175-billion-parameter GPT-3 on unseen
task types in 20 tasks out of 25. \citet{ouyang2022instructgpt} fine-tuned GPT-3 on demonstrations
of the desired behavior written by people, and then with reinforcement learning from human ratings.
People preferred the answers of the resulting 1.3-billion-parameter model InstructGPT to the answers
of the 175-billion-parameter GPT-3, although it has a hundred times fewer parameters. Since then
instruction following has been a goal of training mass-market models, and how well a model follows
instructions has been a subject of measurement. By 2023 the field had taken shape to the point of
receiving its first detailed survey \citep{lou2023survey}: what an instruction is and what kinds
there are, how to model instructions, on which datasets and with which metrics to test instruction
following, and what affects it.

The first systematic measurements answered the question of whether a model follows a given
instruction. So that the check would depend neither on expensive and irreproducible human
evaluation nor on biased evaluation by another model, \citet{zhou2023ifeval} proposed verifiable
instructions---requirements whose fulfilment is checked by a program, for example ``write more than
400 words'' or ``mention the word AI at least three times''. The IFEval set contains 25 types of such
instructions and about 500 prompts, each with one or more instructions.

Subsequent works began to increase the complexity. In FollowBench \citep{jiang2023followbench}, at
every level one constraint of one of five types---content, situation, style, format,
example---is added to the initial instruction, from one to five constraints; for almost all of the
13 models studied the quality falls from the first level to the fifth, and the fulfilment of every
constraint is evaluated by a strong judge model. InFoBench \citep{qin2024infobench} decomposes each
of 500 instructions into simple criteria---2250 ``yes/no'' questions in total---and counts the share
fulfilled; the authors entrust the labeling of answers to GPT-4 as an accurate and cheap
evaluator. ComplexBench \citep{wen2024complexbench} builds a taxonomy of complex instructions---4
types of constraints, 19 of their dimensions and 4 types of composition, that is, of the ways in
which constraints are related to each other: simultaneously, in a chain, by selection on a
condition, by nesting---and derives the final score from the structure of the dependencies between
the constraints; the check is performed by an evaluator model complemented by rules.

In 2024 requirements began to be accumulated over the course of a dialogue. In MMMT-IF
\citep{epstein2024mmmtif} a model answers questions about images in a dialogue of up to 20 turns,
and between the questions it receives requirements on the form of the answer---on the length of the
answer, on the first letter and the last character of a sentence, on the length of a sentence, on a
required word, on the parity of a number---which stay in force until the end of the dialogue; up to
six requirements are in force at once, and the fulfilment of each is checked by a program. The
share of fulfilled requirements, averaged over three models, falls from 0.81 at the first turn to
0.64 at the twentieth and decreases with the number of requirements in force: for Claude~3.5 Sonnet
from one with one requirement to 0.74 with six, for Gemini~1.5 Pro from 0.68 to 0.4. If all
requirements are repeated at the end of the input, the share of fulfilled requirements rises by
22.3 points on average: the difficulty lies not only in fulfilling the requirements but also in
finding them scattered over the dialogue. Multi-IF \citep{he2024multiif} stretches IFEval over three
turns: at the second and the third turn one requirement is added, and the earlier ones remain in
force; the set contains 4501 dialogues in eight languages. For all 14 models studied the fulfilment
falls with every turn---for o1-preview from 0.877 to 0.707---the models increasingly forget
requirements that they fulfilled at earlier turns, and they make more errors in Hindi, Russian and
Chinese.

By the end of 2024 the evaluation of language models came to be regarded as an experiment with all
its statistical requirements. \citet{miller2024errorbars} showed that when questions are related to
each other---for example, several questions about one text---the standard errors must be computed
with a correction for this relation, and on real datasets they turn out to be up to three times
larger than the naive ones (3.05 on DROP).

In parallel ran yet another line---the limits of how a model uses a long input. \citet{liu2023lost}
showed on questions over a set of documents and on key-value retrieval that the accuracy depends on
where in the input the needed item stands: it is highest when the needed item is at the very
beginning or at the very end, and it drops noticeably when it is in the middle, even for models
built for long context. RULER \citep{hsieh2024ruler} extended the simple ``needle in a haystack''
probe with multiple needles, multi-hop tracing and aggregation: almost all of the 17 models find a
single needle flawlessly, but the quality drops sharply with length, and although all of them claim
a window of at least 32 thousand tokens, only half hold satisfactory quality at this length. NoLiMa
\citep{modarressi2025nolima} removed the lexical overlap between the question and the needle, so
that the needle can be found only through a semantic link: 13 models with a claimed window of at
least 128 thousand tokens work well on inputs shorter than a thousand tokens, but at 32 thousand the
accuracy of 11 of them falls below half of their own accuracy on a short input; even for
GPT-4o---from 99.3 to 69.7\%.

At the same time researchers carried the notion of working memory over from psychology to models.
\citet{gong2023workingmemory} gave ChatGPT (gpt-3.5-turbo) the n-back task---a standard measure of
human working memory in which, for every element of a stream, one has to say whether it coincides
with the element n steps back---and found in the model a capacity limit strikingly similar to the
human one; changing the wording of the instruction did not remove this limit, and the authors
proposed n-back as a reference probe of the working memory of language models.
\citet{gong2024selfattention}, training small transformers from scratch on this task, found that the
total entropy of the attention weight matrix grows with n, and linked the limit to the dispersion of
attention.

\citet{wang2025unable} carried over from psychology another classic technique as well---proactive
interference, that is, interference from what was learned earlier. A model reads a stream of
updates of ``key---value'' pairs and must name the last value of every key. The accuracy falls
approximately log-linearly with the number of interfering updates---for all models studied---and
this regularity does not reduce to the length of the input: at a fixed length, a growing number of
tracked keys gives the same decline (llama4-maverick---almost 100\% on two keys and below 5\% on 46),
and in a regression across models the model size is significant, but not the length of its context
window.

When a prompt contains several instructions, it turned out that their order matters as well.
\citet{li2024divide} asked models to write a text with given required words---from 3 to 20 words,
100 sets for each number, with permutations of the words and 20 repetitions---and found that the
quality falls sharply as the number of constraints grows, and that fulfilment depends on the place
of the word in the list for all models. The sign of this bias depends on the model: GPT-3.5, GPT-4
and LLaMA2-13b fulfil the early words of the list more often, LLaMA2-7b and LLaMA3-8b the late
ones. \citet{zeng2025order}, permuting the constraints of one instruction and measuring how close
the order is to a ``hard-to-easy'' ordering, found that models of different architectures and sizes
perform better precisely with this order, and linked this to how a model distributes attention
among the constraints.

\citet{jaroslawicz2025ifscale} measured the same phenomenon with a large number of instructions: the
ratio of the error share in the last third of the list to the error share in the first third. With
few instructions it is close to one, at 150--200 instructions it reaches a peak---late instructions
are violated noticeably more often than early ones---and at extreme densities it levels out again:
the errors are spread evenly over the list.

Works on long context measured how a model uses a long input. LongGenBench
\citep{wu2024longgenbench} asked a different question: does a model keep following instructions
while it itself writes a long text. The prompt sets requirements on particular places of the future
answer of 16 or 32 thousand tokens, and it is checked whether the requirement relating to each place
of the answer is fulfilled there. Fulfilment falls as the answer grows longer: the deviation from the
instructions is noticeable after 4 thousand tokens and grows towards 16 thousand. Requirements on
places are fulfilled rarely: Qwen2-72B at a length of 16 thousand tokens fulfils 24.0\% of them. On
the transition to 32 thousand the models moreover more often fail to finish the answer: Qwen2-72B
writes 66.2\% of the prescribed parts against 94.3\%.
Around 45\% of the long answers degenerated into repetitions.

In 2025 the number of simultaneous instructions became an independent load axis. IFScale
\citep{jaroslawicz2025ifscale} gives a model from 10 to 500 instructions of one kind---include a
given word in a business report---and measures the share fulfilled. Of 20 models, even the best
fulfil only 68\% at 500 instructions. The authors distinguish three forms of decline: threshold,
when a model is almost flawless up to a critical density and then quickly loses quality and becomes
unstable (the reasoning models o3 and gemini-2.5-pro), linear (gpt-4.1, claude-sonnet-4) and
exponential (gpt-4o, llama-4-scout); at high density the errors are mostly omissions of
instructions.

\citet{harada2025multiply} gave models up to ten instructions from IFEval at once and showed that
the share of answers in which all instructions are fulfilled falls steadily with their number. A
logistic regression with a single feature---the number of instructions---predicts this share with an
error of about 10\%, including for combinations of instructions that were absent from the training
data. The same work shows that evaluation by a judge model inflates the result: with ten
instructions, 0.657 against 0.213 by the programmatic check.

Load by many instructions was also studied on material from practice. AgentIF \citep{qi2025agentif}
takes the instructions of real agent applications---the system prompt of an agent together with a
user request. Unlike the short synthetic instructions of earlier sets, such an instruction is long
and contains many constraints of different nature: on the form of the answer, on its content and
style, on compliance with tool descriptions, and conditional ones (``if~\ldots, then~\ldots''). Even
the best of the models studied fully fulfils fewer than 30\% of such instructions; models cope worst
with a complex structure of constraints and with tool descriptions. RECAST \citep{guo2025recast}
starts from the observation that models stop coping when there are more than ten explicit
requirements, while existing datasets contain no more than ten per example; the authors build
training data with a much larger number of constraints extracted from real prompts and show that
fine-tuning on them improves the following of complex instructions. AdvancedIF
\citep{he2025advancedif} collects 1645 expert-written dialogues of three kinds---complex
instructions in a single turn, requirements carried over many turns, and requirements of the system
prompt---with a rubric of up to 20 criteria for each; the fulfilment of every criterion is evaluated
by a judge model, and an answer counts only if all of them are fulfilled. The best of the models
studied pass about three quarters of the dialogues (GPT-5 in reasoning mode---77.9\% on average over
the three kinds), and reinforcement learning in which the reward is the fulfilment of the whole
rubric raises the result of Llama~4 Maverick by 6.7 points.

At the same time it became clear that the measurements themselves are fragile.
\citet{dong2025revisiting} test robustness: one and the same task is presented in several close
variants---paraphrased, with a restructured constraint, with a distracting addition---and a model is
credited with the task only if it solves all the variants. By this measure the quality turns out
noticeably lower than by the usual evaluation on IFEval: the drop reaches 61.8\% of the original,
and even for the most robust model, GPT-5, it is 18.3\%.

In 2026 load began to be set not only by the number of instructions. \citet{mittal2026prospective}
loads a model with a parallel task: requirements on the form of the answer from IFEval are combined
with tasks of growing difficulty---from answering a question to a chain of arithmetic problems and
the retelling of a long text. Under this load the fulfilment of form requirements falls by 2--21\%;
requirements that must be fulfilled at the end of the answer suffer most (up to 50\%), while
prohibitions hardly suffer. The influence is mutual: a form requirement can bring down the solution
of the task itself---for one model the accuracy on the GSM8K arithmetic problems fell from 93 to
27\%. As requirements accumulate, their joint fulfilment falls sharply, while repeating the
instruction at the end of the prompt in many cases brings it back to 90--100\%. The check in all
experiments is programmatic.

SEQUOR \citep{canaverde2026sequor} stretches the load in time further: constraints taken from real
conversations---on style, language, format, numerical limits---must be observed over a dialogue of
50 turns; compliance is evaluated by a judge model. Even a single constraint is observed worse and
worse as the conversation grows longer---from the first turn to the last the accuracy falls by more
than 11\%; with several constraints at once, by more than 38\%; with their sequential introduction,
by more than 40\%. Replacing a constraint in the middle of a conversation restores the initial
level, but then the decline is faster.

\citet{eliav2026promptdesign} tested whether the fulfilment of many rules depends on how they are
presented: in what format to write the rule list and where to place it---in the system prompt or in
the user message. The number of rules checked by a program varied from 10 to 160. The number of
rules itself proved decisive: by 80 rules the share of answers without a single violation falls
practically to zero for all five models, however the rules are presented. The way of presentation
has a weaker effect: the place of the rules matters no less than the format, but in different
directions for different models, and there is no universally best format.

Within a few years the behavior of a language model under load has turned from separate
observations into a subject of systematic measurement. It has been established that the quality of
instruction following falls with the number of instructions, with the length of the input and with
the length of the answer, depends on the place of an instruction in the list and on the order of
instructions, falls under a parallel task and over the course of a dialogue, and that the working
memory of models has a limit similar to the human one. These results were obtained in different
ways. The load was set by the number of instructions, their complexity, the text length, a
parallel task, the duration of a dialogue; fulfilment was checked by a program or by a judge model;
the result was expressed by the share of fulfilled instructions, the share of answers without
violations or the shape of the decline curve. How suitable these ways are for measuring load is
examined in the next subsection.

\subsection{Evaluation of the testing methodology}\label{sec:methodology}

The works described above measure one and the same phenomenon---the decline in quality with growing
load---but in very different ways. To compare them, one must first say what we expect from a
measurement of load. We proceed from a requirement on the meter: the load is set by the
experimenter, the number of instructions serves as its measure, and the result is a number with an
uncertainty in units of this load. Nine criteria follow from this requirement.

\begin{enumerate}
\item \textbf{Atomicity}: every application of an instruction is a Bernoulli trial: the outcome is
  binary and uniquely determined, there is no partial fulfilment, and the instruction does not
  decompose into sub-requirements that can be fulfilled separately.
\item \textbf{Independence}: the requirements of the instructions are separable---each instruction
  determines the correctness only of its own part of the answer, these parts do not overlap, and the
  set of correct answers is the product of the sets admissible for each instruction. Then the
  growth of the load is not confounded with the growth of the difficulty of the task, and the
  decline in quality is attributed to the model. The criterion does not require statistical
  independence of the outcomes: their dependence, arising in the model, is precisely the subject
  of measurement. This is the main criterion.
\item \textbf{Homogeneity}: at equal load the outcomes of different instructions belong to one
  distribution, that is, the instructions are exchangeable and the semantics of an instruction is
  not a factor of the outcome.
\item \textbf{Computable check}: the reference answer is a deterministic function of the task, and
  the outcome of a trial is determined by an algorithm, without an evaluator. The error and the
  bias of the evaluation are zero, and the spread of the outcome belongs to the model alone.
\item \textbf{Absence of extraneous factors}: the outcome does not depend on quantities other than
  the measured one---understanding of the text, knowledge of the world, the wording of the
  instruction; the task is free of variance unrelated to the measured load.
\item \textbf{Resolution by place}: the unit of observation is a single application of an
  instruction; for every trial the outcome and its covariates are known---the position in the
  answer and the number of the instruction in the list---so that the dependence of the success
  probability on the place is estimated, not averaged.
\item \textbf{Scale}: the region in which the experimenter sets the load (the number of
  instructions, the text length); it must cover the transition of the success probability from
  close to one to small, otherwise the parameters of the dependence are not identifiable and the
  capacity is obtained by extrapolation.
\item \textbf{Precision}: when the measurement is repeated---under the same conditions or by an
  independent experimenter on a new random sample of tasks---the results agree within the stated
  uncertainty; for this the procedure is fully specified: the generation of tasks, the model
  version, the generation parameters, the reference answer and the processing.
\item \textbf{Protection against memorization}: tasks are generated at random for every
  measurement and cannot coincide with the training data of the model, so the right outcome cannot
  be obtained by reproducing something memorized.
\end{enumerate}

Two of these criteria---the computable check and the absence of extraneous factors---coincide with
what the practice of evaluating AI agents advises \citep{grace2026demystifying}: to check by a
program wherever possible and by a judge model only where it cannot be avoided, and to start every
trial from a clean state, because a state shared by trials makes their outcomes dependent for a
reason that has nothing to do with the model under test.

By the way the load is set, the works of Section~\ref{sec:history} fall into five approaches: form
instructions checked by a program; composite constraints checked by a judge model; required words in
coherent text; synthetic probes of memory and context; a long answer with placed requirements.
Below each approach is examined against these criteria, and the result is summarized in a table at
the end of the subsection.

\paragraph{Form instructions checked by a program.}
Requirements on the form of the answer are added to a substantive task---to write an essay, a
letter, a program: length, number of paragraphs or sentences, required and forbidden words, letter
case, punctuation, code style rules. The fulfilment of each requirement is checked by a program, and
the load is the number of requirements in one prompt. This is how IFEval \citep{zhou2023ifeval},
the ManyIFEval and StyleMBPP sets \citep{harada2025multiply}, the experiments with a parallel task
\citep{mittal2026prospective} and with up to 160 rules \citep{eliav2026promptdesign} are built, and
in a dialogue, where requirements accumulate over the turns, MMMT-IF \citep{epstein2024mmmtif} and
Multi-IF \citep{he2024multiif}.

The programmatic check is the main merit of the approach, but it serves as a reference only
approximately. The check here is a heuristic over free text, not a function of the task: the
authors of IFEval found false rejections in it (a required phrase highlighted with markup is not
recognized) and introduced a second, loose check that transforms the answer before the comparison.
The outcome of a trial depends on which check is used.

There is no independence, and this is the main point. All requirements apply to one and the same
text: length, number of paragraphs, letter case and punctuation constrain the answer as a whole, so
every new requirement narrows the set of answers admissible for the others. The authors see this
themselves. IFEval notes that a simple combination of instructions yields contradictions---no more
than five paragraphs and fewer than 20 words; \citet{harada2025multiply} compiled a map of conflicts
between 15 instructions (a limit on the number of characters against the number of paragraphs,
upper case against lower case) and combined only compatible ones; \citet{mittal2026prospective} kept
11 types out of 15 for stacking requirements, excluding structural conflicts; in MMMT-IF mutually
exclusive variants of a requirement are gathered into one category, of which at most one is in
force, and in Multi-IF contradicting requirements of different turns are filtered out by a model
and then by people. Selecting compatible requirements removes explicit contradictions but does not
make the task separable: compatible requirements still share one text. The consequence is visible
in the data: the rule ``no more than 79 characters per line'' alone is fulfilled in 97 and 99\% of
cases (Claude~3.5 Sonnet and Gemini~1.5 Pro), and together with five other code style rules in 2
and 20\%. Which part of this decline falls on the limit of the model and which on the increased
difficulty of the joint task cannot be established from such data.

Atomicity is partial. Some requirements are checked by a single feature of the whole
answer---length, number of paragraphs. Others are a conjunction over the whole text: ``no commas'',
``only lowercase letters'', ``every line no longer than 79 characters'' decompose into
sub-requirements on every sentence, word or line and are violated by one place out of many; one
extra comma and twenty give the same outcome. There is no homogeneity: different requirements are
fulfilled with different probabilities already in isolation---for GPT-4o from 0.62 (an exact number
of characters) to 1.0 (a title, quotation marks, a ban on commas), and \citet{harada2025multiply}
excluded instructions with very low success from the set in advance. Therefore the result depends
on which requirements made it into the set, and the number of requirements is confounded with its
composition.

There are extraneous factors: form requirements are imposed on a meaningful text on a given topic,
and the outcome changes noticeably when the same requirement is paraphrased
\citep{dong2025revisiting}; in a dialogue an image that must be understood (MMMT-IF) and the
language (Multi-IF) are added to them.
Resolution by place is partial: the unit of observation is a requirement in the whole answer or the
whole answer, and where in the answer a requirement is violated is not recorded; in a dialogue the
outcome is known for every turn, that is, by place in time \citep{epstein2024mmmtif,he2024multiif}.

Scale---up to 5 simultaneous requirements in \citet{mittal2026prospective}, up to 6 in
\citet{epstein2024mmmtif}, up to 10 in \citet{harada2025multiply}, up to 160 in
\citet{eliav2026promptdesign}. In the experiments of
\citet{harada2025multiply} the probability of fulfilling a single requirement remains high (for
GPT-4o---0.85 with ten requirements against 0.94 with one), and only the share of answers in which
everything is fulfilled covers the transition (from 0.94 to 0.21). In \citet{eliav2026promptdesign}
this share reaches practically zero already by 80 rules, and over the second half of the range the
measure distinguishes nothing.

Precision is unattainable in this approach by construction: requirements differ in difficulty, so a
new sample of tasks gives a different result because of a different composition of the set, not
because of the model. Only a run on the same tasks can be repeated---as \citet{harada2025multiply}
did three times for one model---but this is a check of the repeatability of a run, not of the
measurement.

There is no protection against memorization: IFEval was published in 2023 together with 541
prompts, derivative works take their tasks from it
(\citet{harada2025multiply}---216 task descriptions, \citet{mittal2026prospective}---the original
prompts verbatim, \citet{he2024multiif}---the first turn of every dialogue), and these tasks may end
up in the training data of models released later.

Thus the approach provides a programmatic check and a wide range of the number of requirements, but
the number of requirements does not serve in it as a measure of load: the requirements share one
text, differ in difficulty, and as their number grows the task itself changes.

\paragraph{Composite constraints checked by a judge model.}
An instruction contains several substantive constraints---on content, style, situation, format,
example, on the use of tools---and the fulfilment of each is evaluated by a judge model, and by
rules where the answer is closed. The load is increased by adding constraints to the same
instruction \citep{jiang2023followbench}, by linking them with conditions and nesting
\citep{wen2024complexbench}, by stretching them over the time of a dialogue
\citep{canaverde2026sequor}, by taking long instructions from practice \citep{qi2025agentif} or by
entrusting their writing to experts \citep{he2025advancedif}. The goal of these works is to bring
the measurement closer to real
use, and almost all the deviations from the criteria are the price of this goal.

The difficulty begins already with atomicity: a substantive constraint has no unique outcome.
Whether an answer matches a style, a situation or a given content is a matter of judgment, and the
experts who labeled answers for FollowBench agree with each other in only 85\% of cases. InFoBench
\citep{qin2024infobench} decomposes an instruction into ``yes/no'' questions, but a question about
style or content remains a judgment. If the right outcome is not uniquely defined, a program cannot
provide it; hence the judge model, and with it the error and the bias of the evaluation. They are
large. The FollowBench judge agrees with the experts in 88\% of cases, but without the description
of how the constraints were added the agreement drops by 9 points, and with a template of overall
quality assessment to 67\%: the result depends on how the request to the judge is worded. The
SEQUOR judge wrongly rejects 11.5\% of correct answers---as much as the measured decline over 50
turns with one constraint (more than 11\%), and the data do not allow checking whether the error of
the judge itself grows with the length of the conversation. On form instructions, where a
programmatic check exists, the judge inflates the share of fully fulfilled answers threefold: 0.657
against 0.213 with ten instructions
\citep{harada2025multiply}. The AdvancedIF judge, the o3-mini model, agrees with the expert labels
on individual criteria with an F1 measure of 0.723 \citep{he2025advancedif}. Moreover, the judge is
biased: in the MMMT-IF experiments GPT-4o as a judge rates the answers of GPT-4o higher, and Gemini
the answers of Gemini \citep{epstein2024mmmtif}.

There is no independence, and its absence is built in deliberately. FollowBench obtains each next
level of complexity by adding a constraint to the same instruction. ComplexBench builds
compositions---a chain, a selection by a condition, nesting---that is, dependencies between
constraints; the authors themselves write that because of this coupling the constraints are hard to
evaluate separately, and they assemble the final score from the structure of the dependencies. In
SEQUOR and AgentIF all constraints act on one and the same conversation or answer of an agent. The
growth of the number of constraints here is inseparable from the growth of the difficulty of the
task: this is precisely the subject of study, but the number of constraints does not then serve as
a measure of load.

There is no homogeneity: constraints of different types differ in difficulty---in AgentIF models
cope worst with a complex structure of constraints and with tool descriptions. Extraneous factors
lie at the very core of the approach: a substantive constraint cannot be fulfilled, nor its
fulfilment evaluated, without understanding the text and knowing the world, and this is required
both of the model under test and of the judge.

Resolution by place is partial: InFoBench and ComplexBench give an outcome for every ``yes/no''
question, SEQUOR for every turn of a dialogue, that is, by place in time; the position within the
answer is not recorded. The scale is small, and in some of the works the load is not set by the
experimenter: FollowBench---from one to five constraints, SEQUOR---up to three at once, in existing
datasets, according to the survey of \citet{guo2025recast}---no more than ten per example. AgentIF
has on average 11.9 constraints, but their number is determined by a real instruction, not by the
experimenter, so the load region is not planned at all.

Precision rests on the judge. The judge is a part of the measuring instrument: the result depends on
the choice of the judge model and on the template of the request to it, and the measurement can be
repeated only while the same version of the judge is available. Added to this is the same as in the
previous approach: the constraints differ in difficulty, and a new sample of tasks gives a different
result because of a different composition. There is no protection against memorization: the sets
are fixed and published.

The approach is closest of all to real use, but it pays for this with measurability: the outcome is
not uniquely defined, the constraints are coupled by construction, and they are checked by a model
with its own error comparable with the measured effects.

\paragraph{Required words in coherent text.}
A model writes a coherent text and must include given words in it; the load is the number of words.
IFScale \citep{jaroslawicz2025ifscale} gives from 10 to 500 business terms for a business report;
\citet{li2024divide} take sets of concepts from which a plausible sentence must be composed, and
sets of up to 20 required words. Fulfilment is the presence of the word in the answer, checked by a
program. This is the approach closest to ours.

Atomicity holds: a requirement is one word, the outcome is binary and unique. The presence of a word
is checked by a program by exact match. But which answers enter the count is decided by a model:
IFScale repeats the request if the answer is a plain list of words, a refusal or, according to the
o4-mini model, an incoherent report. The count is taken over the answers selected by the judge.

There is no independence, and the reason is visible in the same rejection. Without the requirement
of coherence the task is trivial---to list the words---and it is precisely such answers that IFScale
discards. The difficulty is created by coherence, and coherence couples all the words: each must be
fitted into one meaningful text together with all the others, and the difficulty of fitting in a
word depends on which words and how many of them must stand in the same text. The growth of the
number of words is therefore confounded with the growth of the difficulty of the task.

There is no homogeneity, and the authors of IFScale account for this themselves. For every word they
estimated the difficulty of generating it by the perplexity of a small model, selected the 500
hardest and chose words stratified by difficulty so that the runs would be comparable.
Stratification is needed precisely because the result depends on which words came up. This
difficulty is semantic: a word must be fitted into a business report, in \citet{li2024divide} into
a plausible scene, and success depends on how well the word suits the topic. For the same reason the
extraneous factors cannot be removed: a model writes a meaningful text, and the outcome depends on
the understanding of the topic and of the meaning of the words.

Resolution by place is only by the number in the list: IFScale compares errors at the beginning and
at the end of the list, \citet{li2024divide} by the position of a word in the prompt. Where a word
stands in the answer is not recorded: presence anywhere counts. The scale is the largest among the
works found: from 10 to 500 words in steps of 10. And yet the transition is not covered for all
models: in the words of the authors themselves, the strongest models do not yet collapse at 500
words.

Precision is better attainable here than in the previous approaches: the tasks are generated by a
rule, and every measurement is repeated on five random samples of words. But the comparability of
the repetitions has to be ensured by selecting the samples by word difficulty, so the result is
defined only relative to the composition of the vocabulary. IFScale has protection against
memorization: the set of words for every request is random, and there is no ready right answer that
could be memorized; part of the experiments of \citet{li2024divide} is built on the published
CommonGen set.

The approach preserves atomicity and the computability of the check and brings the number of
instructions to hundreds. But a coherent text couples the words with each other, and the success of
inserting a word depends on its meaning, so the number of words measures not only the load but also
the difficulty of placing them together.

\paragraph{Synthetic probes of memory and context.}
A model receives a synthetic task with a computable answer: to say whether a letter coincides with
the letter n steps back \citep{gong2023workingmemory,gong2024selfattention}, to name the last value
of a key in a stream of updates \citep{wang2025unable}, to find a value by a key or an answer among
documents \citep{liu2023lost}, to find one or several ``needles'' in a long text
\citep{hsieh2024ruler,modarressi2025nolima}. There is one instruction, and it does not change; the
load is the amount of information that must be held or found: the depth n, the number of updates
and keys, the length of the input and the place of the needed item in it. The answer is short.

By most criteria this is the strictest of the approaches. Every probe is one trial with a unique
answer, and the check is an exact match. Answers to different probes do not constrain each other:
the value of each key and the coincidence of each letter are determined separately, and the
difficulty comes from the input, that is, from the controlled load; the interference in PI-LLM is
set in the input and serves precisely as such a load, not as a link between requirements on the
answer. Homogeneity and the absence of extraneous factors are achieved by the authors deliberately:
\citet{liu2023lost} make the keys and values random identifiers to remove the meaning of language as
a confounding factor, and \citet{wang2025unable} pick words from the vocabulary at random in every
run to exclude the influence of particular semantic combinations. There are two exceptions, and both
are deliberate. NoLiMa requires finding the needle through a semantic link that relies on knowledge
of the world---meaning is the subject of study here. In questions over documents a model also
answers from memory: without any documents GPT-3.5 answers correctly in 56.1\% of cases---better
than when the needed document stands in the middle of the input \citep{liu2023lost}.

Resolution by place exists, but by place in the input: the accuracy is measured as a function of
where the needed item stands. Place in the answer does not arise as a question: the answer is short.
The scale is large and covers the transition: an input of up to 128 thousand tokens in RULER and up
to 32 thousand in NoLiMa, up to 400 updates per key and up to 46 keys in PI-LLM, where
the accuracy of one model falls from almost 100 to less than 5\%; the exception is n-back, where the
depth reaches only three.

Precision is attainable: the material is generated by a rule---random identifiers, words, letter
sequences---and a repetition on a new sample is natural; \citet{wang2025unable} give confidence
intervals over multiple repetitions. For the same reason there is also protection against
memorization. The authors of NoLiMa take special care of it---they assemble the ``haystack'' text
from random short passages of open books---although the needles themselves are a fixed set of 58
``question---needle'' pairs; the questions over documents in \citet{liu2023lost} lack it.

The approach is closest of all to the requirements of measurement, but it measures something else:
holding and finding information in the input with a short answer, not the simultaneous execution of
many instructions. The number of instructions does not serve in it as the load, and how execution
changes over the course of a long answer is not observed.

\paragraph{Long answer with placed requirements.}
LongGenBench \citep{wu2024longgenbench} tests whether a model fulfils requirements tied to particular
places of a long answer. The model writes a text of parts of the same type---a diary with an entry
for every day or week of a year, a description of every floor of a skyscraper, a plan of city
blocks---and the prompt contains seven requirements, each of which names a fact and the part of the
answer where this fact must appear: for example, the wife's birthday in the entry for June 29.
A requirement may also apply to several consecutive parts or repeat with a given step.
The load is the length of the answer: the short variant of the task is designed for 16 thousand
tokens, the long one for 32 thousand. Whether the required fact is in the right part is decided by
the Llama~3.1-8B model.

The strength of the approach is that the requirements are tied to places of the answer and the
outcome of each is known where it must be fulfilled. The authors show how fulfilment changes over
the course of the answer: it drops noticeably after 4 thousand tokens. True, this dependence is
given only as a smoothed curve, and the numbers in the tables are shares over all answers, so
resolution by place is partial. Facts, dates and places are substituted into templates at random,
so the tasks cannot end up in the training data of a model. For the same reason precision is
attainable: a new sample of tasks is drawn from the same templates with the same composition of
requirements; but the authors do not report the uncertainty.

There is no independence, for the same reason as with required words: the requirements are facts
that must be fitted into one coherent text. A diary is the story of one person, and a vacation,
birthdays and regular activities must fit into it together with everything the main task requires:
an entry for every day or week, the weather, work. A range requirement occupies several
consecutive parts, a periodic one repeats over the whole answer, and the dates for all requirements
are chosen at random, so the parts of different requirements may coincide. Every requirement is
also linked to the main task: it can be fulfilled only in a part that the model has written at all,
and in the long variant the models more often fail to finish the answer---Qwen2-72B writes
94.3\% of the prescribed parts at 16 thousand tokens and 66.2\% at 32 thousand.

There is no homogeneity, twice over. Requirements of the three kinds are built differently: a
requirement on one part is fulfilled if the fact is in it, while a range or periodic requirement
only if the fact is in every one of many parts; such requirements are harder than single ones and,
moreover, not atomic. The facts themselves also differ: fitting a birthday into a diary and placing
a gym on the 34th floor of a skyscraper are tasks of different difficulty, and, as with required
words, success depends on how well the fact fits the text. Hence also the extraneous factors:
fulfilment depends on the understanding of the topic and on how the model builds a coherent text.
And a program cannot check such a requirement: whether the fact is in a fragment of free text is
decided by the Llama~3.1-8B model, and although on 300 manually labeled cases it agreed with people
in all of them, the reference remains its judgment.

The scale is narrow. There are always seven requirements, so the approach does not allow loading a
model with the number of instructions; only the length serves as the load, and it has two values.
These are, moreover, two different tasks---a diary by weeks and by days, a skyscraper of 100 floors
and of 300---so together with the length the number of parts and their content change, and the
actual length is determined by the model, not by the experimenter.

The approach places requirements on places of a long answer and knows the outcome of each at its
place. But the requirements in it are substantive facts of one coherent text, different in kind and
meaning, they are checked by a model, and the load is set by two different tasks.

\paragraph{What these measurements provide.}
The result in the works reviewed is expressed in one of three ways. Most often it is the share of
fulfilled instructions at the loads that the authors tested. Sometimes it is the share of answers in
which all instructions are fulfilled. The shape of the decline, finally, is described in words: three
forms in IFScale, an ``approximately log-linear'' decline in PI-LLM, a ``knee'' in
\citet{eliav2026promptdesign}.
Some authors estimate the uncertainty by repetitions, others construct confidence intervals, as
MMMT-IF does for the share of fulfilled requirements at every turn \citep{epstein2024mmmtif}, and
still others do not report it at all.

For scientific research these data are valuable but insufficient. They show that the phenomenon
exists: the quality of execution falls with the number of instructions, with the length of the
input and of the answer, and depends on the place of an instruction. They do not allow studying it
comprehensively. The load is tested at a few points, and a share obtained at a chosen load depends
on which points the experimenter chose and cannot be compared with the result of another work.
Closest of all to describing the dependence itself came \citet{harada2025multiply}: their logistic
regression covers all the tested numbers of instructions, but the authors do not derive from it the
load that a model withstands with a given reliability. Moreover, when outcomes are related to each
other, like the outcomes of the instructions of one answer, the uncertainty must be computed with a
correction for this relation; without the correction the interval comes out narrower than the true
one, on real datasets up to three times \citep{miller2024errorbars}.

A model maker needs a parameter of the model: one number with an uncertainty, measured in the same
way for all models, by which they can be compared with each other. Model makers already compare
models by one number when it comes to compliance with given rules of behavior: OpenAI publishes for
each model the share of answers that comply with the behavior specification of its models---from
72\% for GPT-4o to 89\% for GPT-5 Thinking \citep{guo2026modelspec}. For load by instructions no
work provides such a parameter. IFScale makes it possible to rank models by the share fulfilled at
500 instructions, but this is a single point, and the strongest models do not yet collapse at it. The
regression of \citet{harada2025multiply} contains no load parameter. For another phenomenon---the use
of a long input---such a parameter already exists. RULER and NoLiMa define the effective length: the
largest of the tested input lengths at which the quality of a model stays above a threshold. It can
be compared between models; for GPT-4, for example, with a claimed window of 128 thousand tokens it
equals 64 thousand. This parameter is coarse: the tested lengths go in doublings, so it is known
only to within a factor of two, and it has no interval.

The assessments of the five approaches against the nine criteria are summarized in
Table~\ref{tab:approaches}.

\begin{table}[htbp]
\centering
\small
\caption{Approaches to measuring model behavior under load by the criteria of
Section~\ref{sec:methodology}. Column numbers are the numbers of the criteria: 1---atomicity,
2---independence, 3---homogeneity, 4---computable check, 5---absence of extraneous factors,
6---resolution by place, 7---scale, 8---precision, 9---protection against
memorization.}\label{tab:approaches}
\begin{tabular}{@{}p{5.4cm}*{9}{c}@{}}
\toprule
Approach & 1 & 2 & 3 & 4 & 5 & 6 & 7 & 8 & 9 \\
\midrule
Form instructions checked by a program
  & part. & no & no & part. & no & part. & part. & no & no \\
Composite constraints checked by a judge model
  & no & no & no & no & no & part. & no & no & no \\
Required words in coherent text
  & yes & no & no & part. & no & part. & part. & part. & yes \\
Synthetic probes of memory and context$^{a}$
  & yes & yes & yes$^{b}$ & yes & yes$^{b}$ & part. & yes$^{c}$ & yes & yes$^{b}$ \\
Long answer with placed requirements
  & part. & no & no & no & no & part. & no & yes$^{d}$ & yes \\
\bottomrule
\end{tabular}

\smallskip
\begin{minipage}{\linewidth}
\footnotesize
$^{a}$~What is measured is holding and finding information in the input with a short answer; the
number of instructions does not serve as the load.\\
$^{b}$~Except NoLiMa and the questions over documents.\\
$^{c}$~Except n-back.\\
$^{d}$~Attainable by construction, but the authors do not report the uncertainty.
\end{minipage}
\end{table}

\paragraph{Summary of the review.}
The review shows two directions that have so far developed separately. Works on instruction
following study the needed phenomenon and bring the tasks ever closer to real use: the instructions
in them are substantive, linked by a common text and increasingly checked by a judge model. But the
closer a task is to real use, the less in it lends itself to measurement. Synthetic probes of
memory and context, on the contrary, are built as a measuring instrument: the outcome in them is
unique, the tasks are generated at random, and the load is set by the experimenter. They study,
however, another phenomenon---holding and finding information in the input.

In sum, the behavior of a language model under load by instructions has been established as a
phenomenon but not measured as a quantity. It is known that the quality of execution falls with the
number of instructions and with the length of the answer and depends on the place of an
instruction. But it cannot be said what load a given model withstands, how the number of
instructions, the length and the place act together, and how models differ in this ability. To
answer these questions, one needs a test built after the model of synthetic probes but loading the
model precisely with instructions.

%% file: sections/03-instrument.tex
% !TEX root = ../main.tex
\section{The Instrument}\label{sec:instrument}

To measure the degradation of a language model under various kinds of load we built an
instrument---a software system that generates tasks, presents them to a model and checks its
answers. By construction the instrument satisfies the criteria of Section~\ref{sec:methodology}: the
load is set by the experimenter, the material of a task is generated at random and carries no
meaning, and the outcome of every application of an instruction is unique and checked by a program.

The load is formed by instructions that the model must execute simultaneously. A task consists of a
text and a list of instructions for it. The text is a stream of words from a fixed vocabulary of
single-token nouns, without meaning and punctuation, broken into lines of ten words
(Sections~\ref{sec:vocabulary} and~\ref{sec:task}). An instruction is a rule that replaces one word
of the text with another word of the vocabulary (Section~\ref{sec:instruction}). The model must
apply all the rules to the text, keeping it otherwise unchanged.

A task is determined by six numbers:
\begin{itemize}
\item the number of rules $m$;
\item the number of distinct words of the text $D$;
\item the number of repetitions of each word $n$;
\item three seeds of the random number generator---the rule seed, the text seed and the rule order
  seed (Section~\ref{sec:task}).
\end{itemize}

The answer of the model is compared with a reference answer computed by a program, and every
occurrence of a word to be replaced becomes a trial with the outcome 1 or 0. For every trial the
number of its rule in the list and the position of the word in the answer are known
(Section~\ref{sec:validation}).
The result of a measurement is the outcomes of the trials; the degradation of the model shows itself
in a decrease of the share of right replacements as the load grows
(Section~\ref{sec:instrument-summary}).

\subsection{The atomic instruction---word replacement}\label{sec:instruction}
A rule is written as \texttt{lake -> stone}: the model must replace every occurrence of the word
\texttt{lake} in the text with the word \texttt{stone}. Such a rule is an atomic instruction. It
applies to one word and does not decompose into simpler requirements, and at every place of
application it admits exactly one right outcome: the word \texttt{stone} must stand in place of the
word \texttt{lake}.

These properties make a rule suitable for measurement. The correctness of every application is
established by comparison with the reference answer, without a judgment about the quality of the
answer. The rules are not linked to each other: each applies to its own word, and executing one does
not change what is required to execute another. Therefore the load can be increased to hundreds of
simultaneous rules without changing the nature of the task. Finally, executing a rule does not
depend on understanding the content of the text. Both words of a rule are one token for each of the
tokenizers compared, so a rule maps one input token to one output token
(Section~\ref{sec:vocabulary}).

One rule serves as a conventional unit of load, and the number of rules $m$ as the measure of load.
The unit is conventional: it is defined by the protocol of the instrument---the prompt template, the
structure of the text and the rules of answer checking---and by the vocabulary. Values expressed in
rules are comparable between models measured under the same protocol.

The task requires of the model neither reasoning, nor knowledge of the world, nor arithmetic, nor
understanding of the text. Therefore an error of the model reflects, with high probability, the
limit on the number of independent correspondences that the model can hold and apply
simultaneously, not a misunderstanding of the task.

\subsection{Construction of the vocabulary}\label{sec:vocabulary}
The text of a task and the rules are composed exclusively of words of a fixed vocabulary, and the
composition of the vocabulary determines what exactly the instrument measures.

The vocabulary serves three purposes:
\begin{itemize}
\item standardization of the material: tasks for all models are built from the same words, and the
  data obtained on different models are comparable with each other;
\item reproducibility: a task is built by a random choice of words from a fixed list and can
  therefore be restored from the seeds of the generator;
\item interchangeability of words: one word differs from another only by its role in the task, not
  by its own properties, which would otherwise enter the measurement; the residual differences of
  particular words, including semantic ones, are smoothed out because every load is measured on
  different random sets of words and rules (Section~\ref{sec:task}).
\end{itemize}

Four requirements on the words of the vocabulary follow from these purposes:
\begin{itemize}
\item a word is encoded as one token by each of the tokenizers compared---then a replacement remains
  an atomic operation, a partial replacement of a word is impossible, and the length of the prompt
  in tokens does not depend on the choice of words;
\item a word is a monosyllabic noun---this ensures the homogeneity of the material, and replacing a
  noun with a noun preserves the grammar of the stream;
\item a word is in the singular, which excludes pairs such as \texttt{arm}/\texttt{arms} that would
  turn a replacement into a test of morphology;
\item a word is at least three letters long, which excludes the names of letters and function words.
\end{itemize}

The initial list for the vocabulary consisted of all entries of the CMU Pronouncing Dictionary
(\texttt{cmudict.0.7a})---123\,455 words. The CMU Pronouncing Dictionary, like the WordNet database
used further on, is distributed as a fixed version, so rebuilding the vocabulary gives the same
result.

From the initial list we kept the words written only in lowercase Latin letters and at least three
letters long. We determined the number of syllables from the phonetic transcription of a word in the
CMU Pronouncing Dictionary: in it every vowel phoneme is marked with a stress digit, and the number
of such phonemes equals the number of syllables. Monosyllabic words remained.

We determined the part of speech and the number of a word from the WordNet~3.0 database. The words
that are nouns and coincide with their base form (lemma), that is, stand in the singular, remained.

The last step was the single-token check. We encoded every remaining word, with a space before it,
with three tokenizers: \texttt{o200k\_base} (the gpt-4o and gpt-4o-mini models), \texttt{cl100k\_base}
(gpt-3.5-turbo) and the tokenizer of the Qwen2.5 model. The space was added because in the text of a
task and in the rules a word stands after a space, and tokenizers encode the word together with it.
The vocabulary includes the words that give exactly one token with all three tokenizers.
Table~\ref{tab:vocabulary} shows the number of words remaining after each step.

\begin{table}[htbp]
\centering
\small
\caption{Selection of the words of the vocabulary \texttt{v1\_intersection}: the number of words
after each step.}\label{tab:vocabulary}
\begin{tabular}{@{}p{8.4cm}r@{}}
\toprule
Step & Words \\
\midrule
Entries of the CMU Pronouncing Dictionary & 123\,455 \\
Only lowercase Latin letters & 115\,533 \\
At least three letters & 115\,318 \\
One syllable & 14\,884 \\
WordNet nouns & 5750 \\
Singular: the word coincides with its lemma & 3581 \\
One token for each of the three tokenizers & 2220 \\
\bottomrule
\end{tabular}
\end{table}

The current version of the vocabulary has 2220 words from three to eight letters long. This makes it
possible to form tasks with $D$ up to 2048 words and more. The single-token property of the words of
this version is guaranteed only for the three tokenizers listed. For a model with a different
tokenizer the vocabulary is built anew by the same rules, and the single-token property is required
only for the tokenizer of the model under study: a common vocabulary for all models is not needed,
because the comparability of measurements is ensured by averaging over random sets of words and
rules. The vocabulary, together with the code for building and validating it, like the other
materials, is published in the GitHub repository
(Section~\ref{sec:availability}).

\subsection{Construction of a task}\label{sec:task}
The load of a task is determined by three parameters: the number of rules $m$, the number of
distinct words of the text $D$ and the number of repetitions of each word $n$. The triple
$(D, n, m)$ is called a point and sets the position of the task in the load space. The parameters
are bounded by the conditions $1 \le m \le D$ and $D + m \le 2220$, which follow from the size of
the vocabulary (Section~\ref{sec:vocabulary}).

A task is formed by a rule list and a text substituted into a fixed prompt template. A point
determines only the load parameters; the specific material of a task is determined by three seeds
of the pseudorandom number generator:
\begin{itemize}
\item the rule seed sets the choice of the words of the rules and their division into pairs;
\item the text seed sets the choice of the background words and the order of all words in the text;
\item the order seed sets the order of the lines in the rule list.
\end{itemize}
A task at a point, together with the answer of the model to it, is called a request. In the
published data of the instrument a request corresponds to a \texttt{point} record.

Requests at one point with different seeds realize one and the same load on different material;
averaging over them smooths out the residual differences related to particular words. The prompt is
a deterministic function of the point, the three seeds and the vocabulary and is reproduced byte for
byte, so the task of every published request can be restored and checked.

To build the rule list, the generator initialized with the rule seed chooses $2m$ words without
replacement from the vocabulary sorted alphabetically. The first $m$ words become the words to be
replaced, the last $m$ their replacements, and the pairs are formed in order. Sorting the vocabulary
makes the result of the choice independent of the internal storage order of the words. The sets of
words to be replaced and of replacements do not overlap, so the rules do not form chains such as
\texttt{lake -> stone}, \texttt{stone -> fish}, and the execution of each rule does not depend on the
others.

The position of a rule in the list, the rule number $r$, is recorded for every trial. With a zero
order seed the lines follow in the order of the choice; with any other value a random permutation is
applied to them that preserves the pairs themselves. This makes it possible to present the same pairs
in a different order and to separate the contribution of the position of a rule in the list from the
difficulty of a particular pair of words.

The text is built from $D$ distinct words: $m$ words to be replaced and $D - m$ background words. The
background words are chosen by the generator initialized with the text seed from the vocabulary
minus the $2m$ words of the rules; therefore replacements are absent from the text, and in the right
answer every occurrence of a replacement corresponds to a firing of a rule. Each of the $D$ words
enters the text exactly $n$ times, after which the sequence is shuffled by the same generator. Thus
all words of the text have the same frequency, and none of them stands out by the statistics of the
stream. The stream is broken into lines of ten words without punctuation; the lines serve as anchors
in checking the answer (Section~\ref{sec:validation}).

The rule list and the text are substituted into the prompt template fixed by the protocol of
version~0.1: the rules are written one per line in the order of the list, the text in the form built.
\begin{verbatim}
Rewrite the text below by applying the substitution rules.

Rules (replace every occurrence of the left word with the right word):
{rules}

Strict requirements:
- Replace only the words listed in the rules.
- Do not change, add, or remove any other words.
- Keep the exact word order, spaces, and line breaks.
- Output only the rewritten text, nothing else.

Text:
{text}
\end{verbatim}
The template is not a parameter of the experiment: changing the wording means a new version of the
protocol, so the wording of the instruction does not act as a factor of the outcome.

As an example we give a published request at the point $(D, n, m) = (4, 5, 2)$---a text of 20 words
and two rules:
\begin{verbatim}
Rules:
james -> hang
reign -> steal

Text:
reign cab reign luck cab luck reign reign cab reign
james luck james james luck james cab james cab luck
\end{verbatim}
Each of the two words to be replaced occurs in the text five times, so the request contains 10
trials. The full prompt and the answer of the model are given in Appendix~\ref{sec:appendix}.

Tasks whose right answer exceeds the answer length limit of the model are not included in the
measurement: a cutoff of the answer at this limit would be indistinguishable from the degradation of
the model. Requests not executed because of a technical failure of the provider are also excluded
and produce no trials.

\subsection{Answer validation}\label{sec:validation}
Validation turns the answer of the model into a sequence of Bernoulli trials: for every occurrence of
a word to be replaced in the text, the corresponding position of the answer is compared with the
reference answer computed by a program. Neither an evaluator---a person or a model---nor measures of
text similarity enter the measurement.

The reference answer is built in a single pass: every occurrence of a word to be replaced is replaced
by the word assigned to it by a rule, and the other words are kept. A single pass is enough, because
the sets of words to be replaced and of replacements do not overlap, and replacements do not occur
in the text.

The answer is matched to the reference answer line by line, and the lines serve as synchronization
anchors. A line with the reference number of words is scored position by position; in a line with a
different number of words the matching initial segment is counted, and the remaining positions are
considered unmatchable. The next line is matched anew, so the consequences of a local failure are
confined to one line. A wrong replacement thus remains a measured event rather than a cause of lost
synchronization.

Three kinds of deviation of the answer from the reference answer are distinguished---failures of the
first, second and third kind:
\begin{itemize}
\item a failure of the first kind is a wrong replacement with the structure of the answer preserved;
\item a failure of the second kind is a local disruption of a line (a missing or inserted word),
  after which the structure of the answer recovers;
\item a failure of the third kind is a loss of the structure of the answer, after which matching
  with the lines of the reference answer becomes impossible.
\end{itemize}
The kind of a failure characterizes the degree of damage to the answer and is unrelated to errors of
the first and second kind in statistical hypothesis testing.

Line-by-line matching meets two requirements. First, a local failure must not affect the scoring of
the rest of the answer: if the answer were compared with the reference answer as a single stream, the
omission of one word would shift all subsequent positions. Second, the framing of the answer must not
shift the matching. Models often supplement the rewritten text with an introductory or closing
phrase; without excluding it, a flawless answer would be scored as completely wrong. Therefore the
lines added before and after the text are excluded, provided that after this the number of lines of
the answer matches the reference.

The outcome of a trial is 1 if the expected replacement stands at the corresponding position of the
answer, and 0 in all other cases, including when the position cannot be matched with the reference
answer because of a failure of the second or third kind. A damaged answer is not considered a
malfunction of the instrument: it gives a full measurement, and unmatchable positions are counted as
unsuccessful, so the score of the model cannot be overestimated.

\subsection{What the instrument measures}\label{sec:instrument-summary}
The result of a measurement for every request is the sequence of the outcomes of its trials, taking
the values 1 or 0. Every trial is characterized by the number of its rule in the list and the
position of the word in the answer, and a request by the parameters of the point $(D, n, m)$. These
data are primary observations and do not depend on the statistical model used for their analysis.

The instrument is defined by the protocol---the vocabulary, the procedure of task construction and
the procedure of answer checking---and does not depend on the statistical model. Statistical models
are built over the data obtained and can be refined without changing the instrument, and the
accumulated data allow repeated analysis. The statistical model used in this work is described in
Section~\ref{sec:analysis}.

This design makes it possible to study the response of a model to overload in an arbitrary load
range and the influence of each of the load parameters. The points are chosen by the experimenter
within the limits set by the size of the vocabulary and the answer length limit of the model---from
a single rule on a short text to loads at which the model stops executing the rules entirely.

The plan of the experiment is not predetermined by the instrument: varying one parameter with the
others fixed, building a grid of points and choosing points sequentially from the results of
previous measurements are all admissible.

Repeated requests are possible at one point. Requests with new seeds realize the same load on
different material, which makes it possible to estimate the spread between requests and to smooth
out the influence of particular words. Requests with the same pairs of words in a different order,
set by the order seed, make it possible to separate the contribution of the position of a rule in
the list from the difficulty of a pair of words.

Since the task of a request is fully determined by six numbers, it can be presented to different
models; a comparison of models is then carried out on identical tasks, not only at equal load
parameters. For the same reason the tasks of published requests are reproducible: any researcher can
repeat the measurement on another model or check the published results.

%% file: sections/04-analysis.tex
% !TEX root = ../main.tex
\section{The Statistical Model}\label{sec:analysis}
The outcome of an individual trial is random: at one and the same load a replacement is executed in
some trials and not in others. Therefore the measured quantity is the probability of a right
replacement as a function of the load. It cannot be observed directly: a trial gives only 0 or 1,
and the probability is estimated only from many trials.

The load is multidimensional and is described by several axes. A load axis is a quantity that is
computed from the conditions of a trial---the point of the request, the rule number and the position
of the word in the answer---and describes one side of the load.

The relation between the probability and the load is set by a statistical model---a function of the
load axes with a finite number of parameters, which are estimated from the outcomes of all trials at
once. From the estimated statistical model the characteristics of a language model are computed,
for example the load at which the share of right replacements falls to a given quality level.

For such
characteristics to make sense, the statistical model must satisfy three requirements:
\begin{itemize}
\item \textbf{Comparability}: a characteristic is intended for comparing language models with each
  other. Since the probability of a right replacement depends on the load, characteristics obtained
  under different conditions would differ not only because of the properties of the language models
  but also because of the difference in conditions; therefore for all language models a
  characteristic is computed under the same fixed conditions. Moreover, a characteristic describes
  the language model, not the plan of the experiment: a measurement of the same model on a different
  set of points must give the same value within the uncertainty.
\item \textbf{Interpretability}: the dependences described by the statistical model must admit an
  explanation by the properties of the language model. In particular, an increase of the load must
  not raise the probability of a right replacement: such a rise, especially in the region of large
  loads, contradicts the nature of the phenomenon. If the statistical model shows it, this is a
  reason to look for an error in the statistical model itself---in the choice of the axes, in the
  form of the dependence or in the confounding of the contributions of different axes---before
  attributing it to the language model.
\item \textbf{Applicability beyond the instrument}: the axes of the statistical model make sense
  also for tasks that language models solve outside the instrument. The prompt of an applied task
  contains a set of instructions of different complexity and the input data to which these
  instructions are applied; the answer of the model is formed on the basis of the input data. The
  task of the instrument reproduces the same structure in an extremely simplified form: the role of
  an instruction is played by a replacement rule serving as a conventional unit of load, the role of
  the input data by a text of vocabulary words, the role of the answer by the transformed text.
  Consequently, only quantities defined for any prompt of this structure can serve as axes of the
  statistical model, for example the number of instructions, the position of an instruction in the
  list, the length of the input text or the position in the answer. Parameters defined only within
  the instrument, for example the number of repetitions of each word, do not serve as axes.
\end{itemize}

The statistical model must also take into account the properties of the data that follow from the
design of the instrument:
\begin{itemize}
\item the outcome of a trial is binary;
\item the load axes are linked by the construction of the task: the position of a word does not
  exceed the length of the text, and the rule number does not exceed the number of rules;
\item the trials of one request are linked by the common difficulty of the task;
\item the requests are distributed unevenly over the load space.
\end{itemize}

The form of the dependence of the probability of a right replacement on each axis is not known in
advance. Therefore the statistical model first measures it from the data, without imposing a form,
and then describes the measured dependence by smooth laws, from which the characteristics of the
language model are computed.

\subsection{The Bernoulli trial and the logit}\label{sec:bernoulli}
The outcome of a trial $z$ is binary, and every trial is treated as a Bernoulli trial:
$z = 1$ with probability $\pi$ and $z = 0$ with probability $1 - \pi$, where $\pi$ is the probability
of a right replacement in the given trial. The statistical model describes how this probability
depends on the load.

The probability itself cannot be described by a sum of contributions: the probability is bounded by
the interval from 0 to 1, and a sum of contributions can go beyond it. Therefore the logit is
modeled---the logarithm of the odds of a right replacement, that is, of the ratio of the probability
of a right replacement to the probability of a wrong one:
\[
\eta = \ln\frac{\pi}{1 - \pi}, \qquad \pi = \frac{1}{1 + e^{-\eta}}.
\]
The logit takes any real value, and the contributions of the load add up in it
\citep[\S\,4.4]{hastie2009elements}. This meets the requirement of interpretability: every
contribution is read separately in a common unit---the logit---and a change of the logit by $\Delta$
multiplies the odds of a right replacement by $e^{\Delta}$.

The logit of trial $i$ of request $j$ is represented as the sum of the contributions of the
quantities describing the load and the shift of the request:
\[
\eta_{ij} = \sum_k \theta_k x_{ijk} + b_j, \qquad b_j \sim N(0, \sigma^2),
\]
where $x_{ijk}$ is the value of the $k$-th quantity describing the load of the trial and $\theta_k$
is its coefficient.

The shift $b_j$ is one number for all trials of a request, reflecting the difficulty of the
particular task, that is, of its material. It is introduced because the trials of one request are
linked by the common difficulty of the task: without the shift, the spread of task difficulty would
not be separated from the contributions of the load. The shift is considered a random variable with
a normal distribution, and its variance $\sigma^2$ characterizes the spread of request difficulty. A
logistic model with a random shift common to the observations of one group belongs to the
generalized linear mixed models (GLMM) \citep{rabehesketh2002reliable}.

The coefficients $\theta_k$ describe the request that is median in difficulty, with $b_j = 0$, that
is, the action of the load on an individual task, as interpretability requires. The share of right
replacements taken over all requests is obtained from the same model by averaging over the
distribution of the shift; the characteristics comparable between language models are defined
through it. The coefficients describing an individual request and the coefficients describing the
share over all requests have different scales and are not comparable with each other
\citep[Sec.~VII.C]{cameron2015practitioner}.

The shifts of individual requests are not estimated. The likelihood of every request---the
probability of all its outcomes---is averaged over the normal distribution of the shift, and the
coefficients $\theta_k$ together with the variance $\sigma^2$ are found by maximizing this marginal
likelihood \citep{rabehesketh2002reliable}.

\subsection{Variables and bins}\label{sec:binned}
The behavior of a language model under load is not known in advance. Therefore the statistical
model cannot rely on a dependence specified before the measurement---linear or following another
law chosen in advance: such a model would fit the parameters of a given form instead of measuring
the form itself. Instead, the values of each axis are divided into intervals---bins---and each bin
is assigned its own coefficient, which enters the logit only if the value of the axis falls into
this bin. The contribution of the axis to the logit is then piecewise constant, and the form of the
dependence is determined by the data \citep[\S\S\,5.1--5.2]{hastie2009elements}.

The dependence of the probability on the load axes is poorly conveyed by bins of equal width, for
three reasons:
\begin{itemize}
\item the values of the axes span several orders of magnitude, and bins of equal width would merge
  the small values into one or two bins and split the rest of the range into many bins;
\item there are fewer data on large values than on small ones: the beginning of the answer and the
  first rule are present in every request, while distant positions and late rules only in requests
  with a long text and a long rule list;
\item the load acts relatively: what matters is by how many times it has grown, not by how much, and
  in works on load the decline in quality is sometimes close to linear in the logarithm of the load
  \citep{wang2025unable}.
\end{itemize}
Therefore logarithmic bins are used---bins of equal width on the logarithmic scale of the axis: the
upper boundary of each next bin is twice that of the previous one. Each bin corresponds to one
doubling of the value of the axis, the relative precision is the same over the whole range, and the
width of the bins grows where there are fewer data, so every bin contains a sufficient number of
trials.

The logit of trial $i$ of request $j$ in the binned model is written through bin indicators:
\[
\eta_{ij} = \sum_{u} \sum_{k} \theta_{uk}\, I_{uk}(x_{u,ij}) + b_j, \qquad
I_{uk}(x) =
\begin{cases}
1, & x \in B_{uk},\\
0, & x \notin B_{uk},
\end{cases}
\]
where $u$ is the number of the axis, $k$ the number of a bin of this axis, $B_{uk}$ the $k$-th bin of
axis $u$, $x_{u,ij}$ the value of axis $u$ in the trial, $\theta_{uk}$ the coefficient of the bin,
and $b_j$ the shift of the request. The bin indicators play the role of the quantities $x_{ijk}$ of
the formula of Section~\ref{sec:bernoulli}: of the sum over the bins of each axis, the coefficient of
the bin into which the value of the axis fell remains.

The binned representation describes the dependence of the probability on each axis
separately---by the profile of the coefficients of its bins. Such a profile is unique only if the
axes are orthogonal: with orthogonal regressors the estimates of the coefficients of one regressor
do not depend on the others, while with strongly correlated ones they become unstable
\citep[\S\,3.2.3]{hastie2009elements}. The load axes are in general not orthogonal, since they are
linked by the construction of the task. Therefore a requirement analogous to orthogonality is
imposed on their set: the statistical model is built on axes whose bin estimates are minimally
correlated with each other. The measure is the correlation of the bin estimates of two axes,
computed from the covariance of the fit; the values for all pairs of axes are collected in a table
of axis correlations, and a correlation close to one means that the contributions of the two axes
are not separated by the data.

\subsection{Reference bins and the level}\label{sec:reference}
In the conventional notation of a binned model the logit contains a free term $\alpha$ common to all
trials:
\[
\eta_{ij} = \alpha + \sum_{u} \sum_{k} \theta_{uk}\, I_{uk}(x_{u,ij}) + b_j.
\]
The bins of each axis form a partition of its values, so $\sum_k I_{uk}(x) = 1$ for any $x$, and
for any constant $c$ the substitution
\[
\alpha \to \alpha + c, \qquad \theta_{uk} \to \theta_{uk} - c \quad \text{for all } k
\]
does not change the logit of any trial. The parameters of such a notation are not uniquely
determined. The redundancy is usually removed by setting the coefficient of one bin of each axis to
zero: $\theta_{u k_u^0} = 0$; the bin $k_u^0$ is called the reference bin. Then $\alpha$ is the logit
of the median request in the combination of the reference bins of all axes, that is, the level of
the model, and the coefficients of the other bins are counted from the reference ones. Only the
differences of the coefficients of the bins of one axis are then determined by the data, and the
table of estimates gives only comparisons with the reference bin \citep{firth2000quasi}.

The choice of the reference bin is arbitrary, and the level depends on it. If instead of bin $k^0$ of
axis $u$ the bin $k^1$ is made the reference one, the estimates are transformed as follows:
\[
\alpha' = \alpha + \theta_{u k^1}, \qquad \theta'_{uk} = \theta_{uk} - \theta_{u k^1},
\]
while the logit of every trial remains the same. The data do not distinguish these notations, but
the level in them is different. Consequently, $\alpha$ and any quantity counted from it are
determined not only by the language model but also by the choice of the coding, which contradicts
the requirement of comparability.

The second weakness concerns bins in which all outcomes are of one sign. Let all replacements in the
bin $B_{uk}$ be right. The derivative of the log-likelihood with respect to the coefficient of this
bin, for any value of the shift of the request,
\[
\frac{\partial \ell}{\partial \theta_{uk}}
  = \sum_{x_{u,ij} \in B_{uk}} \bigl(z_{ij} - \pi_{ij}\bigr)
  = \sum_{x_{u,ij} \in B_{uk}} \bigl(1 - \pi_{ij}\bigr) > 0
\]
is positive at any finite value of the coefficient; therefore the likelihood, including the marginal
one, grows as $\theta_{uk} \to \infty$, and no finite maximum-likelihood estimate exists. For a bin
where all replacements are wrong, the same holds as $\theta_{uk} \to -\infty$. This is a special case
of the general condition for the existence of maximum-likelihood estimates in logistic regression
\citep{albert1984existence}. Such a bin is excluded from the fit together with its trials, and the
check is repeated until there is nothing left to exclude. It is natural to choose the bin of the
smallest loads as the reference one, but it is precisely in the combination of the smallest loads
that there may be no errors at all; then the reference bin is excluded, and the level $\alpha$ is not
determined by the data.

Therefore our statistical model has no free term: the logit has the form given in
Section~\ref{sec:binned}. The level in it is not tied to any individual bin, and bins in which all
outcomes are of one sign are excluded without harm to the determinacy of the model. Between the axes
the same redundancy remains as between $\alpha$ and the bins: the common constant of the logit can be
moved from the bins of one axis to the bins of another without changing the logit of any trial, and
it is removed by an arbitrary condition. Consequently, the data determine only the shape of the
profile of each axis, that is, the differences between the coefficients of its bins, but not its
shift. The shifts of all profiles are established later from the smooth laws drawn through the bins,
and only then is the level determined.

\subsection{Approximation shift correction}\label{sec:bias}
The profile of the coefficients of the bins serves as an estimate of the dependence of the
probability on the axis that is free of assumptions about its form. A smooth law is fitted to this
estimate as a model with fewer parameters: the parameters of the law are chosen so that the values of
the law in the bins are closest to the bin estimates in the metric given by the covariance of these
estimates. Such a two-step scheme---first free, ``reduced-form'' parameters, then ``structural''
parameters linked to them by a known mapping---is classical minimum distance estimation
\citep[\S\S\,1, 2.2.4]{newey1994large}. It makes it possible to choose the form of the law from the
measured profile rather than to specify it in advance.

The parameters of the law of axis $u$ are found by minimizing the distance
\[
\hat\psi_u = \arg\min_{\psi_u}\,
\bigl[\hat\theta_u - h_u(\psi_u)\bigr]^{\mathsf{T}}\, \hat\Omega_u^{-1}
\bigl[\hat\theta_u - h_u(\psi_u)\bigr],
\]
where $\hat\theta_u$ is the vector of estimates of the coefficients of the bins of the axis,
$\hat\Omega_u$ is their covariance obtained in the fit, and $h_u(\psi_u)$ is the vector of values of
the law with parameters $\psi_u$ in the same bins. The weight matrix inverse to the covariance of the
reduced-form parameters is optimal: with it the estimates of the structural parameters have the
smallest asymptotic variance among minimum distance estimates \citep[\S\,5.2]{newey1994large}. Bins
excluded from the fit (Section~\ref{sec:reference}) have no estimate and do not enter the distance.
The parameters of the law are uniquely determined by the remaining bins if different parameter values
give different values of the law in the bins; for this the number of bins with an estimate must be no
smaller than the number of parameters of the law \citep[\S\,2.2.4]{newey1994large}.

The fit of the bins determines the shape of the profile of each axis, but not the level at which the
profile lies. As shown in Section~\ref{sec:reference}, the coefficients of the bins of one axis can be
increased and those of another decreased by the same amount without changing the logit of any trial.
Therefore the level at which the bins of each axis end up is set by an arbitrary condition of the
fit, not by the data. An additional restriction that fixes such a constant and makes the parameters
uniquely determined is called a normalization \citep[\S\,2.2, footnote~11]{newey1994large}.

The law of an axis is fitted with a free constant term---a normalization constant. While the other
parameters describe the shape of the profile, the normalization constant takes the level at which the
bins ended up. Then the arbitrary normalization is replaced by a meaningful one: the law, together
with the bins of the axis,
is shifted so that the law vanishes at the smallest load along this axis. The size of this shift is
called the shift of the profile. If at vanishing load the law converges asymptotically to a
horizontal line, the shift equals the level of this asymptote. If the law is defined at the minimum
load, the shift equals its value at this load. After the shift, the profile of each axis shows the
change of the logit relative to the smallest load along this axis.

The shift $\delta_u$ of axis $u$ is subtracted from the coefficients of all its bins, and the sum of
the shifts of all axes forms the reference level of the statistical model:
\[
\tilde\theta_{uk} = \theta_{uk} - \delta_u, \qquad \alpha_{\mathrm{phys}} = \sum_u \delta_u,
\]
\[
\eta_{ij} = \alpha_{\mathrm{phys}} + \sum_u \sum_k \tilde\theta_{uk}\, I_{uk}(x_{u,ij}) + b_j.
\]
A trial falls into exactly one bin of each axis, so the shift of each axis is subtracted from its
logit, and the level returns their sum: the logit of no trial changes. The normalization constant of
the law is free, so the shift of an axis moves together with its bins: however the common constant of
the logit was distributed between the axes in the fit, the corrected bins and the level do not depend
on it.

The parameters of the laws, the shifts and the level are functions of the estimates of the
coefficients of the bins, and their uncertainties are determined by the covariance of the fit. The
covariance of these quantities is obtained by the delta method---linearization of the function at the
point of the estimate:
\[
\operatorname{Cov}\bigl(g(\hat\theta)\bigr) \approx \nabla g\,\Sigma\,\nabla g^{\mathsf{T}},
\]
where $\hat\theta$ are the estimates of the coefficients, $\Sigma$ their covariance, and $\nabla g$
the matrix of derivatives of $g$ at the point $\hat\theta$. For the parameters of a law with the
optimal weight matrix the linearization gives the asymptotic covariance
$\bigl(G_u^{\mathsf{T}}\hat\Omega_u^{-1}G_u\bigr)^{-1}$, where $G_u$ is the matrix of derivatives of
the values of the law in the bins with respect to its parameters \citep[Theorem~3.2]{newey1994large}.

The approximation correction removes the indeterminacy left by the fit of the bins. A level appears
again in the statistical model---the reference level $\alpha_{\mathrm{phys}}$, the logit of a right
replacement for the median request at the smallest load along all axes. Unlike the level $\alpha$ of
the conventional notation, which depends on the choice of reference bins
(Section~\ref{sec:reference}), the reference level does not depend on the coding: it is determined by
the data through the laws of the axes. All bins are corrected by the approximation curves: each curve
either converges asymptotically to zero at vanishing load along its axis or crosses zero at the point
of minimum load $(D, n, m) = (4, 5, 1)$. The contributions of all axes are now counted from one
state---the smallest load---and express in logits the price of the load along each axis. Therefore
the contributions of different axes are comparable with each other, and their sum shows what part of
the reference level a given load spends.

\subsection{The notion of instruction capacity}\label{sec:quantities}

\paragraph{Definition.}
The statistical model describes how the probability of a right replacement depends on the load. A
characteristic of a language model is the inverse quantity---the load at which this probability falls
to a given quality level~$q$. The probability of a right replacement depends on all load axes at
once, so such a load is defined only at fixed values of the other axes. For it to be comparable
between language models, the other axes are fixed at the smallest load~$x^0$---in the same state in
which the reference level is defined (Section~\ref{sec:bias}). The instruction execution capacity,
or instruction capacity (Instruction
Execution Capacity, IEC), $\mathrm{IEC}_q$ is the number of rules~$m$ at which the probability of a
right replacement in this state equals~$q$, that is, the solution of the equation
\[
\pi(m;\, x^0) = q
\]
with respect to~$m$.

This is the inverse problem of regression: from the estimated dependence of a response on an
exposure, the exposure at which the response reaches a given value is found. In this way, in
dose--response analysis the effective dose is determined---the dose at which the probability of a
response equals a given value---and the ratio of the effective doses of two curves serves as a measure
for comparing two substances \citep[Sec.~``Inverse regression'']{ritz2015doseresponse}. The capacity
is expressed in conventional units of load---rules. Unlike the effective input length, which is
chosen from the tested lengths (Section~\ref{sec:methodology}), it is a continuous quantity with an
uncertainty. We use two quality levels: $q = 0.95$ corresponds to reliable execution of the rules,
$q = 0.5$ to the level at which only half of the replacements are right.

The quality level~$q$ refers to the share of right replacements over all requests, whereas the
coefficients of the statistical model describe the request that is median in difficulty
(Section~\ref{sec:bernoulli}). The share over all requests at the logit~$\eta$ of the median request
is obtained by averaging over the distribution of the shift of a request:
\[
\bar\pi(\eta) = \int \frac{\varphi(b;\, 0, \sigma^2)}{1 + e^{-(\eta + b)}}\, db,
\]
where $\varphi(b;\, 0, \sigma^2)$ is the density of the normal distribution of the shift. The
transition from the probabilities of an individual group of trials to the probabilities over the
whole population of groups by integrating over the random effect is a known technique for
random-effects models \citep[Sec.~3]{hedeker2018marginalization}. Therefore on the logit scale the
level~$q$ corresponds to a threshold---the logit of the median request $\eta^*(q, \sigma^2)$ at which
the share over all requests equals~$q$:
\[
\bar\pi\bigl(\eta^*(q, \sigma^2)\bigr) = q,
\]
and the condition ``the share of right replacements is not below~$q$'' is equivalent to the
condition $\eta \ge \eta^*(q, \sigma^2)$.

At $\sigma^2 = 0$ the threshold coincides with the logit of the quality level
\[
L_q = \ln\frac{q}{1 - q}:
\]
$L_{0.95} = \ln 19 \approx 2.94$ and $L_{0.5} = 0$. At $q = 0.5$ the threshold is zero for any
variance of the shift: the logistic function is symmetric, $1/(1 + e^{x}) = 1 - 1/(1 + e^{-x})$, and
with a symmetric distribution of the shift the share over all requests at a zero logit of the median
request equals exactly~$0.5$. At $q = 0.95$ the threshold grows with the variance of the shift.
Consequently, $\mathrm{IEC}_{0.5}$ does not depend on the spread of request difficulty, while
$\mathrm{IEC}_{0.95}$ does.

The same definition applies to any axis along which the load accumulates, for example to the
position of a word in the answer or to the length of the text. The capacity along such an axis is its
value at which the share of right replacements falls to~$q$, and this capacity is expressed in the
units of its axis. Capacities along different axes have different units and do not add up. The
capacity in the number of rules shows how many rules a language model holds at once, and the
capacity in the position in the answer how long it applies them over the course of the answer.

\paragraph{Computation.}
The difference between the reference level and the threshold
\[
\Delta_q = \alpha_{\mathrm{phys}} - \eta^*(q, \sigma^2)
\]
is called the reserve of the level: by this many logits the load can lower the logit of the reference
state before the share of right replacements falls to~$q$. At $q = 0.5$ the whole reference level
serves as the reserve. In a logistic model with a linear predictor $\eta = \beta_0 + \beta_1 x$ and
without the shift of a request, the solution of the inverse problem is written explicitly,
$x_q = (\beta_0 - L_q)/(-\beta_1)$: the exposure spends the reserve $\beta_0 - L_q$ at $-\beta_1$ per
unit. The capacity is built in the same way, but the load is rules, and each next rule has its own
price.

The price $a(r)$ of the rule with number~$r$ is the decrease of the logit brought by adding this rule
to the table; what it consists of is determined by the axes of the statistical model. The first rule
is present in any task, so it enters the reference level and has no price. Rules with different
numbers cost differently, and their prices add up one by one: the cumulative price of a table of~$m$
rules equals
\[
C(m) = \sum_{r=2}^{m} a(r).
\]
The capacity is the largest number of rules whose cumulative price fits into the reserve, with the
fraction of the rule at which the reserve is exhausted:
\[
\mathrm{IEC}_q = m_0 + \frac{\Delta_q - C(m_0)}{a(m_0 + 1)}, \qquad
m_0 = \max\{\, m : C(m) \le \Delta_q \,\}.
\]
The fraction of the last rule makes the capacity a continuous quantity. If the cumulative price is
given by a smooth law, the capacity is the root of the equation $C(m) = \Delta_q$; such equations are
in general solved numerically \citep[Sec.~``Inverse regression'']{ritz2015doseresponse}. The capacity
along another axis~$u$ is computed in the same way---as the root of the equation
$C_u(x) = \Delta_q$, where $C_u$ is the cumulative price of the load along this axis.

Within the region of loads represented in the measurement, the capacity is measured. If the reserve is
not exhausted by the edge of this region, the capacity by the smooth law is obtained by extrapolation
and is marked as an extrapolation. By the bins there is no price of a rule beyond the last bin, and
only a lower bound of the capacity is known---the edge of the measured region.

The capacity is computed in two ways: by the corrected bins, where the prices are given by the
coefficients of the bins, and by the smooth laws. The first computation shows what is measured, the
second what the accepted form of the dependence gives, and the discrepancy between them serves as a
measure of the quality of the approximation.

\paragraph{Uncertainty.}
The capacity is a function of the estimates of the statistical model, and its uncertainty, like the
uncertainty of an effective dose \citep[Sec.~``Inverse regression'']{ritz2015doseresponse}, is
obtained by the delta method (Section~\ref{sec:bias}). The capacity is given by the equation
$C(m) = \Delta_q$ implicitly, so its derivative with respect to the estimates~$\theta$, which here
include the variance of the shift~$\sigma^2$, is found by differentiating this equation:
\[
\frac{\partial\, \mathrm{IEC}_q}{\partial \theta}
  = -\,\frac{\partial C/\partial \theta - \partial \Delta_q/\partial \theta}{\partial C/\partial m}.
\]
The variance of the shift enters the capacity uncertainty only through the threshold:
$\partial \Delta_q/\partial \sigma^2 = -\,\partial \eta^*/\partial \sigma^2$. At $q = 0.5$ this
derivative is zero, and the uncertainty of the estimate of $\sigma^2$ does not enter
$\mathrm{IEC}_{0.5}$.

%% file: sections/04b-results.tex
% !TEX root = ../main.tex
\section{The Passport of GPT-4o mini}\label{sec:results}
The experimental instrument and the statistical model described in the previous sections were used
for a systematic study of the behavior of the GPT-4o mini language model under controlled load. The
results obtained make it possible to identify the factors that determine the gradual degradation of
the quality of instruction following and the occurrence of failures of the third kind, and to
establish quantitative regularities of the influence of each of these factors.

\subsection{The measurement}\label{sec:measurements}

The language model GPT-4o mini, presented by OpenAI on 18 July 2024 as the most cost-efficient of
its small models \citep{openai2024gpt4omini}, was chosen as the object of measurement.
The study was carried out on the version \texttt{gpt-4o-mini-2024-07-18} with a context window of
128\,000 tokens and a maximum answer length of 16\,384 tokens \citep{openai2026gpt4ominidocs}.
These limits were taken into account when forming the tasks: in particular, the full right answer
had to fit within the answer length limit of the model (Section~\ref{sec:task}).

The same generation parameters were used for all requests, chosen so as to minimize the variability
of the answers unrelated to the content of the task. The temperature was set to zero, and the
\texttt{seed} parameter was fixed at the value 20260907, which made it possible to increase the
reproducibility of the results when the same task was presented again. The value
\texttt{top\_p = 1} excluded an additional restriction of the choice of tokens by cumulative
probability. The \texttt{max\_tokens} parameter was set to 16\,384, that is, to the maximum value
admissible for this model, so that the settings of the request would not introduce additional
restrictions on the length of the generated answer.

The scale of the measurements characterizes the computational costs needed to study one model. The
measurement covered 1890 experimental points and included 2958 requests,
each of which received an answer. The points of the measurement were chosen by a closed adaptive
planning technology (Section~\ref{sec:availability}). The length of the presented texts ranged from
20 to 10\,240 words,
and the number of rules in a task from 1 to 256. The total number of trials (words requiring
replacement) was 642\,430. According to the counters of the provider, during the measurements the
model processed 7\,850\,106 input tokens and generated 8\,091\,766 output tokens. The total running
time of the instrument was 5~h 43~min.

\subsection{Choice of the axes of the predictor model}\label{sec:predictor_axes}

The axes of the predictor model were chosen sequentially, by testing alternative sets of axes. The
goal was to find a coordinate system that would make it possible to estimate the main factors of
degradation separately and to predict the probability of correct execution of an individual trial,
including at experimental points that did not take part in the fit of the statistical model.

Initially a parametric statistical model was used in which the probability of a right replacement
depended on five axes: the length of the text \(v=Dn\), where \(D\) is the number of distinct words
and \(n\) is
the number of repetitions of each word; the number of distinct words \(D\); the number of rules
\(m\); the ordinal number of the checked rule \(r\) and the position of the word of the trial in the
generated answer \(s\), counted from one. In addition, two interactions were taken into account: of
the number of rules with the length of the text and with the number of distinct words. The
dependences on \(v\) and \(D\) were set by analytic functions chosen in advance.

The first results cast doubt on the validity of this choice of functional forms. In particular, the
experimental data did not allow reliably distinguishing a linear and a logarithmic dependence on the
length of the text. Using a single slope in \(D\) also led to carrying the regularities measured at
small values of this axis over to the region of large values without sufficient experimental
confirmation. Therefore the next step was the transition from functions set in advance to a free
binned parametrization, which makes it possible to estimate the profile of the influence of each axis
directly from the data.

However, even after this the contribution of the number of distinct words remained hard to
interpret: an increase of \(D\) was accompanied by an increase of the probability of a right
replacement, as if greater lexical diversity made the task easier. An analysis of the construction of
the task showed that at a fixed length of the text an increase of \(D\) simultaneously decreases the
number of repetitions of each word \(n=v/D\). Thus the original axis \(D\) in fact also reflected the
inverse dependence on repetitiveness. For a more natural interpretation of the load it was replaced
by the number of repetitions \(n\). This replacement did not add new information, since for a known
\(v\) the parameters \(D\) and \(n\) are uniquely expressed through each other, but it made it
possible to represent the observed effect differently.

At the same time one more problem of the original coordinate system came to light. Significant
interaction terms showed that the influence of the number of rules depends on the length of the text.
This agreed with the assumption that the rule table and the processed text compete for limited
resources of the model and indicated that the load in the chosen coordinates cannot be
satisfactorily represented by a sum of independent contributions.

To achieve a more stable separation of the factors, we moved from describing a task through its
constructive parameters to describing individual trials with regard to the order of their execution.
The new statistical model used three axes: the number of rules \(m\), the number of the checked rule
\(r\) and the event number \(g\)---the ordinal number of the current trial among all trials of the
request, including itself. These axes described, respectively, the size of the rule table, the
position of a rule in the table and the progress of the model through the sequence of replacements.

The contributions of the three axes were well separated, but the statistical model contained no
explicit characteristic of the total volume of the processed text. As a result, the degradation
related to its increase was partly attributed to the number of rules, since in the experimental data
these quantities changed together. Therefore the length of the text \(v\) was returned to the
statistical model. The opposite approach was also considered---describing a request exclusively by
its original parameters \(D\), \(n\) and \(m\). However, such a statistical model did not take into
account the position of an individual trial in the sequence
of execution of the task and, consequently, did not make it possible to separate the overall load of
the request from its change as the answer is generated.

The next essential step was the introduction of the replacement density

\[
p=\frac{m}{D},
\]

defined as the share of distinct words of the text for which replacement rules are given. The need
for this axis became apparent when comparing tasks with the same number of rules but a different
lexical composition. For example, with eight rules and eight distinct words the model correctly
executed approximately half of the replacements, whereas with the same eight rules and 64 distinct
words practically all of them, even though the first text was shorter than the second. Consequently,
the absolute number of rules does not characterize the load they create without taking into account
what share of the words requires the application of rules.

The introduction of the density made it possible to form a new triple of task coordinates: the length
of the text \(v\), the replacement density \(p\) and the number of rules \(m\). To check the validity
of this choice, we compared it with alternative parametrizations. Returning to the original
parameters \(D\), \(n\) and \(m\) improved the separability of the estimates but worsened the quality
of the description of the experimental data and the accuracy of prediction at points that did not
take part in the fit. Including \(D\) and \(n\) together with \(v\), \(p\) and \(m\) also gave no new
information, since the original parameters are uniquely recovered from the new ones:

\[
D=\frac{m}{p},\qquad
n=\frac{v\,p}{m}.
\]

Additional interaction terms improved the description of the data used in the fit but did not
provide a corresponding improvement of the forecast at new points. Based on the totality of these
results, the triple \(v,p,m\) was chosen to characterize the load of a request as a whole.

After the task coordinates were chosen, it remained to determine which axes are needed to describe an
individual trial. Besides the rule number \(r\) and the event number \(g\) already used, the position
of a word in the answer \(s\) and the application number of a rule \(k\)---the ordinal number of the
current application of the given rule among all its applications in the request---were tested.

It turned out that these quantities contain a considerable share of common information. The position
of a word approximately corresponded to the ratio \(g/p\), and the application number of a rule to
the ratio \(g/m\). When the axes were included together, even singling out the residual
contributions of \(s\) and \(k\) not explained by these dependences did not improve the forecast at
new experimental points. This created the need to choose the most suitable coordinate of progress
through the answer, rather than keep several axes that mainly describe the same process.

Separately, the possibility of using the position of a word \(s\) instead of the total length of the
text \(v\) was tested. Such a representation worsened the quality of the description of the data and
led to a strong mutual dependence of the position, the event number and the replacement density. This
confirmed the need to distinguish between the total volume of the presented text and the position of
an individual trial in the generated answer.

An additional difficulty in separating the factors was the interrelation of the length of the text
and the replacement density, caused by the very construction of the tasks: very small values of the
density are attainable only with a sufficiently long text. To separate more explicitly the progress
through the sequence of trials from the influence of the intermediate words that do not require
replacement, an auxiliary axis was introduced---the excess background:

\[
Q=s-g\frac{v+1}{H+1},\qquad H=mn,
\]

where \(H\) is the total number of trials in the request. The quantity \(Q\) characterizes the
deviation of the actual position of the checked word from the expected position of the corresponding
trial under a random placement of the trials in the text. It was assumed that such a coordinate would
make it possible to single out better the influence of the background text without mixing it with
the ordinal number of the replacement being executed.

To test this hypothesis, a series of experiments was carried out in which \(Q\), \(g\), \(v\) and
\(s\) were simultaneously included in the statistical model, after which individual axes were
excluded one by one.
It turned out that the profile of the excess background does not retain an independent stable form
and in fact is a mixture of the contributions of the length of the text and the position of the word.
The event number \(g\) and the position \(s\) also described mainly the same progress through the
answer. When \(g\) was excluded, the dependence on \(s\) acquired a more regular profile, whereas the
contribution of the rule number \(r\) remained stable in all the parametrization variants tested.

These results determined the choice of two coordinates of an individual trial: the rule number \(r\)
and the position of the word \(s\). The event number \(g\), the application number of a rule \(k\)
and the excess background \(Q\) did not provide an independent stable contribution that would justify
keeping them alongside the chosen axes.

To check the completeness of the resulting system, one more possible axis was considered---the
position of the checked word in the full context \(c\), defined by the number of tokens from the
beginning of the prompt to the corresponding word in the answer, including the instruction and the
rule table. This axis contained additional information but could replace neither the length of the
text \(v\) nor the position of the word \(s\). Therefore it was not included in the main set of five
axes and remained a candidate for further extension of the statistical model.

Finally, the need to keep the number of rules \(m\) as an independent coordinate was checked again.
When it was excluded, the corresponding contribution was redistributed between the replacement
density and the rule number, which confirmed the need to take \(m\) into account explicitly.
Re-including the number of repetitions \(n\) and the application number of a rule \(k\) revealed no
new factors improving the statistical model.

Thus, as a result of the sequential comparison of alternative parametrizations, a set of five axes of
the predictor model was chosen:

\begin{itemize}
\item \(V\)---the length of the word list \(v = Dn\);
\item \(S\)---the ordinal number of a word in the list \(s\), counted from one;
\item \(P\)---the replacement density \(p = m/D\);
\item \(M\)---the number of rules \(m\);
\item \(R\)---the ordinal number of the checked rule in the table \(r\).
\end{itemize}

For the GPT-4o mini language model, the chosen coordinate system is the most complete of the sets of
axes studied with a low mutual correlation (Table~\ref{tab:axes_correlation}).

\begin{table}[htbp]
\centering
\small
\caption{Correlations of the axes of the predictor model from the data of the GPT-4o mini
measurement: the canonical correlation of the increments of the coefficients of neighboring bins of
two axes.}
\label{tab:axes_correlation}
\begin{tabular}{@{}c*{5}{c}@{}}
\toprule
 & \(V\) & \(S\) & \(P\) & \(M\) & \(R\) \\
\midrule
\(V\) & 1 & 0.33 & 0.76 & 0.53 & 0.15 \\
\(S\) & 0.33 & 1 & 0.41 & 0.22 & 0.36 \\
\(P\) & 0.76 & 0.41 & 1 & 0.45 & 0.10 \\
\(M\) & 0.53 & 0.22 & 0.45 & 1 & 0.42 \\
\(R\) & 0.15 & 0.36 & 0.10 & 0.42 & 1 \\
\bottomrule
\end{tabular}
\end{table}

The chosen coordinate system makes it possible to characterize both the overall load of a task and
the heterogeneity of its action on individual trials. The comparative experiments carried out confirm
the sufficiency of the five-axis parametrization: the additional axes tested did not provide an
independent contribution justifying a further complication of the statistical model.

In the statistical model with separate axes, the contributions of the length of the word list and of
the position of the word are represented additively. Consequently, a doubling of the length should
lead to the same change of the logit regardless of the position of the word in the answer. However,
the measurements showed that the size of this change depends substantially on the position: at the
beginning of the answer the decrease of the logit when the length doubles is
1.4--1.8, whereas deep in the answer it approaches zero. Such behavior cannot be described by a sum
of two independent contributions. Therefore the separate axes \(V\) and \(S\) were combined into a
two-dimensional grid in which each combination of length and position bins has its own free
coefficient.

Thus, in the statistical model of the passport the joint influence of length and position is
represented by a single term:

\[
\eta_{ij} = VS(v, s) + P(p) + M(m) + R(r) + b_j,
\]

where \(VS(v, s)\) is the coefficient of the cell of the two-dimensional grid corresponding to the
given trial, and \(P(p)\), \(M(m)\) and \(R(r)\) are the coefficients of the bins of the other axes.
After the notation of its axes, the model was named VS-PMR.

In this parametrization length and position enter the model jointly, not as two independent
terms.

\subsection{Laws and observations}\label{sec:laws}

From the results of the GPT-4o mini measurement, the fit of the VS-PMR statistical model gave the
profiles of the coefficients of the bins of all load axes. The form of these profiles was not set in
advance, yet each of them turned out to be regular and is described by a smooth law with a small
number of parameters, whose form is determined by the shape of the profile itself. These laws are the
regularities found: each of them expresses how the load factor described by the axis lowers the
probability of a right replacement.

\subsubsection{Text length and word position}\label{sec:laws_vs}

\begin{figure}[htbp]
\centering
\includegraphics[width=\linewidth]{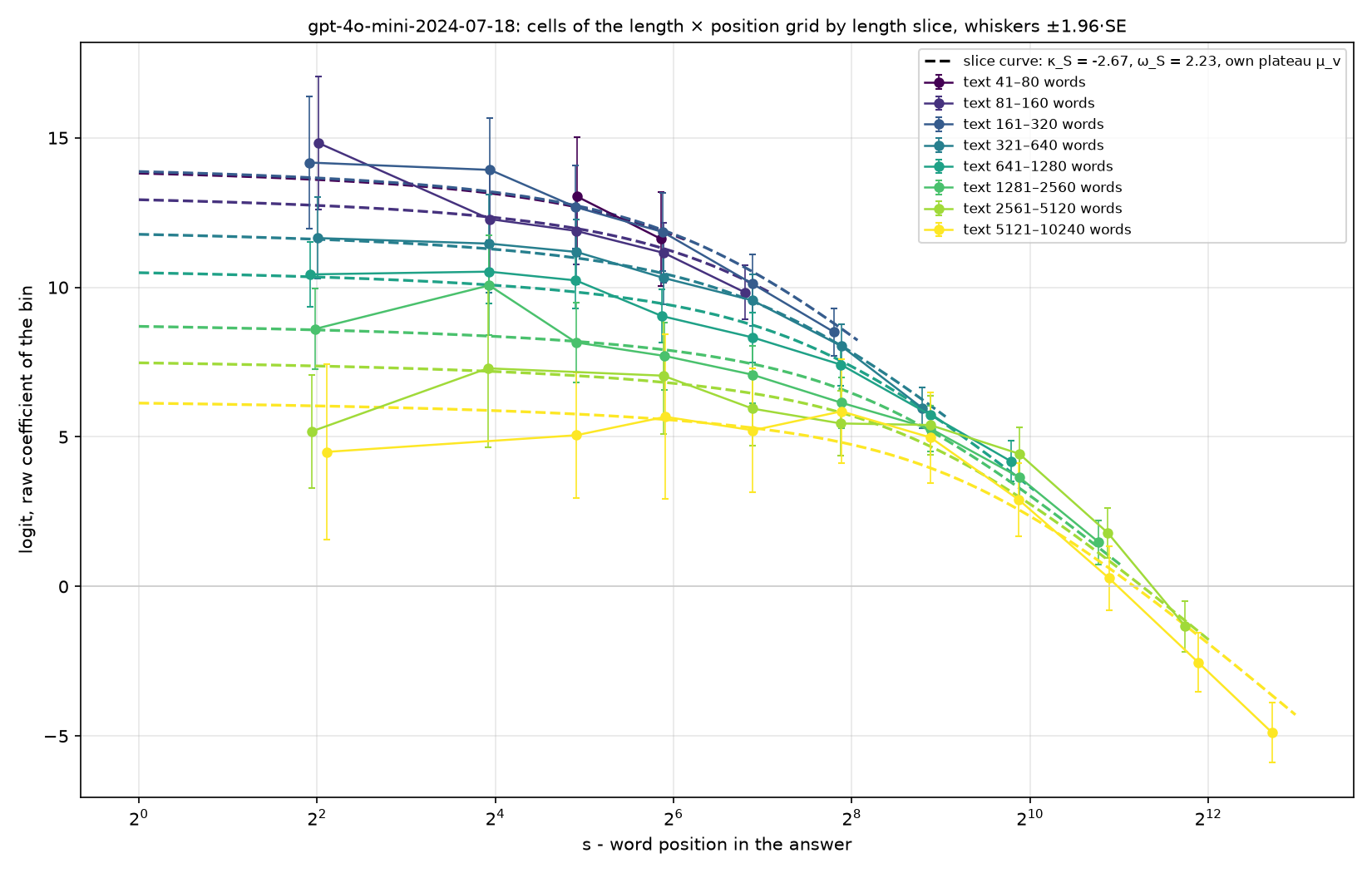}
\caption{Coefficients of the cells of the length and position grid by text length slice before the
shift correction. The points of one slice are connected by a line; the dashed line is the hyperbola
of the slice with the shared parameters $\kappa_S$, $\lambda_S$, $\omega_S$ and its own plateau level
$\mu_v$. Whiskers are the 95\% confidence interval of the coefficient of a cell.}
\label{fig:vs_raw}
\end{figure}

The coefficients of the cells of the length and position grid obtained by the fit of the bins are
shown in Figure~\ref{fig:vs_raw}. The cells of one text length bin form a slice of the grid, and each
slice is represented by a polyline connecting its cells by the position of the word in the answer.
There are no slices of texts up to 40~words in the figure: in all their cells, as in the first two
cells of the slice of 41--80~words, the replacements were right, and by the rule of
Section~\ref{sec:reference} these cells are excluded from the fit. The level at which the
coefficients lie is set by an arbitrary condition of the fit (Section~\ref{sec:reference}), and only
the shape of the profiles is meaningful. Here and below the uncertainty is given as the half-width of
the 95\% confidence interval.

The profiles of all slices have the same shape. At the beginning of the answer the coefficients
hardly change and stay at the plateau level, its own for each slice, and from a certain position on
they decrease approximately linearly in the logarithm of the position, the slope of this decrease
being the same for all slices, and deep in the answer the profiles converge. This shape is set by
two asymptotes---a horizontal one, which determines the plateau level of the slice, and a slanted
one, common to all slices. With a sharp transition between them the profile would coincide with the
smaller of the two asymptotes,
\[
\min\bigl\{\mu_v,\; \lambda_S + \kappa_S \log_2 s\bigr\},
\]
and the observed smooth transition is described by a hyperbola that is a smoothed minimum of these
asymptotes:
\[
VS_v(s) = \frac{1}{2}\Bigl[\lambda_S + \kappa_S \log_2 s + \mu_v
  - \sqrt{(\lambda_S + \kappa_S \log_2 s - \mu_v)^2 + \kappa_S^2\,\omega_S^2}\,\Bigr],
\]
where
\begin{itemize}
\item $\mu_v$ is the plateau level of slice $v$, that is, of the horizontal asymptote to which the
  profile of the slice tends at the beginning of the answer;
\item $\lambda_S + \kappa_S \log_2 s$ is the slanted asymptote, which the profile approaches deep in
  the answer;
\item $\kappa_S$ is the slope of the slanted asymptote in logits per doubling of the position: by
  this much the logit decreases with every doubling of the word number past the corner of the
  hyperbola;
\item $\lambda_S$ is the value of the slanted asymptote at the first word of the answer, at $s = 1$;
\item $\omega_S$ is the width of the transition between the asymptotes in doublings of the position:
  at the intersection point of the asymptotes the curve passes below them by
  $|\kappa_S|\,\omega_S/2$.
\end{itemize}

The hyperbola was fitted to all cells of the grid at once: the parameters $\lambda_S$, $\kappa_S$ and
$\omega_S$ are common to all slices, while the plateau level $\mu_v$ is its own for each slice. The
fit gave $\kappa_S = -2.67 \pm 0.31$ and $\omega_S = 2.23 \pm 0.53$, and the curves of the slices,
shown in Figure~\ref{fig:vs_raw} by dashed lines, pass through the cells within their uncertainties.
Consequently, the slices differ only by the level of the horizontal asymptote.

The weights of this fit were the inverses of the variances of the estimates of the cells, without
their mutual covariance. The cells of one slice are strongly correlated, and with weights inverse to
the full covariance the fit drove the curve of a slice away from all its cells. The uncertainties of
the parameters were then computed from the full covariance of the fit by the delta method
(Section~\ref{sec:bias}).

\begin{figure}[htbp]
\centering
\includegraphics[width=0.85\linewidth]{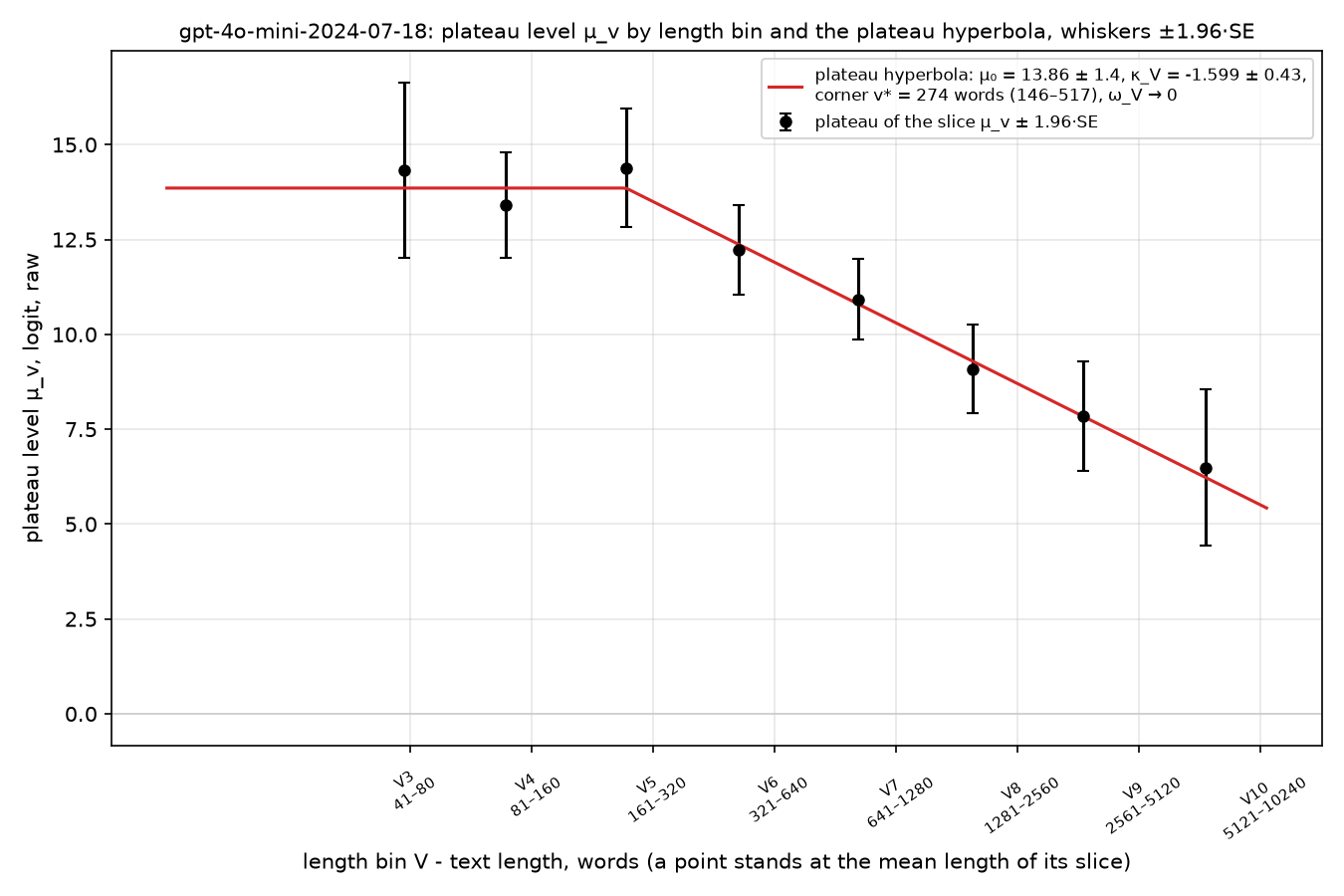}
\caption{Plateau levels of the slices $\mu_v$ before the shift correction and the hyperbola of the
plateau law. The point of a slice stands at the mean length of its texts; whiskers are the 95\%
confidence interval of the plateau level.}
\label{fig:vs_plateau_raw}
\end{figure}

Since the shape of the profile is common to all slices, to describe the whole grid it remains to
establish how the plateau level depends on the length of the text. The plateau levels of the slices
are shown in Figure~\ref{fig:vs_plateau_raw}. For texts from 41 to 320~words the plateau is the same
within the uncertainties---$14.3 \pm 2.3$, $13.4 \pm 1.4$ and $14.4 \pm 1.6$---while for longer
texts it decreases approximately linearly in the logarithm of the length and for texts of
5121--10\,240~words falls to $6.5 \pm 2.1$. The dependence of the plateau on the length has the same
form of a transition from a horizontal asymptote to a slanted one as the profile of a slice in the
position, and it is described by the same hyperbola:
\[
\mu(v) = \mu_0 + \frac{\kappa_V}{2}\Bigl[\log_2\frac{v}{v_*}
  + \sqrt{\log_2^2\frac{v}{v_*} + \omega_V^2}\,\Bigr],
\]
where
\begin{itemize}
\item $\mu_0$ is the level of the horizontal asymptote to which the plateau tends for short texts;
\item $\kappa_V$ is the slope of the slanted asymptote in logits per doubling of the length: by this
  much the plateau decreases with every doubling of the length of a long text;
\item $v_*$ is the text length at which the asymptotes intersect, that is, the position of the
  transition from the plateau to the slanted asymptote;
\item $\omega_V$ is the width of the transition between the asymptotes in doublings of the length.
\end{itemize}
The hyperbola was fitted to the plateau levels of eight slices with weights inverse to their
variances; the position of a slice on the length axis is given by the mean logarithm of the text
length of its cells. The fit gave $\mu_0 = 13.86 \pm 1.37$, $\kappa_V = -1.60 \pm 0.43$ and
$v_* \approx 274$~words (from 146 to 517~words). The width of the transition is not determined by
the data, and its estimate converges to zero, so no noticeable rounding of the transition was found;
at the same time $v_*$ estimates the position of the transition within the accepted law, not a sharp
length threshold established by the measurement. The horizontal asymptote is determined by the
slices of texts from 41 to 320~words, whose plateaus coincide within the uncertainties.

Substituting the plateau law into the hyperbola of a slice gives the law for the whole grid:
\[
VS(v, s) = \frac{1}{2}\Bigl[\lambda_S + \kappa_S \log_2 s + \mu(v)
  - \sqrt{(\lambda_S + \kappa_S \log_2 s - \mu(v))^2 + \kappa_S^2\,\omega_S^2}\,\Bigr].
\]
With a sharp transition this law coincides with the smaller of the two quantities, $\mu(v)$ and
$\lambda_S + \kappa_S \log_2 s$, and with a finite width of the transition it is their smoothed
minimum. Thus the non-additivity of the contributions of length and position, because of which these
axes were combined into a grid (Section~\ref{sec:predictor_axes}), receives an analytic description.

At the beginning of the answer, as $s \to 0$, the slanted asymptote grows without bound, since
$\kappa_S < 0$, and the law tends to the plateau $\mu(v)$, while as the text length decreases the
plateau tends to $\mu_0$. The limit $s \to 0$ serves as a mathematical continuation of the law that
sets the reference level, whereas the actual word positions start at $s = 1$. Consequently, at
vanishing load along both axes the law converges asymptotically to the horizontal line $\mu_0$, and
by the rule of Section~\ref{sec:bias} the shift of the profile of the grid equals the level of this
asymptote: $\delta_{VS} = \mu_0 = 13.86 \pm 1.37$. It is precisely $\mu_0$ that serves as the
normalization constant of the law: if the value of the slanted asymptote at the first word and the
plateau are counted from it, $\tilde\lambda_S = \lambda_S - \delta_{VS}$ and
$\tilde\mu(v) = \mu(v) - \delta_{VS}$, the law takes the form
\[
VS(v, s) = \delta_{VS} + \widetilde{VS}(v, s),
\]
\[
\widetilde{VS}(v, s) = \frac{1}{2}\Bigl[\tilde\lambda_S + \kappa_S \log_2 s + \tilde\mu(v)
  - \sqrt{(\tilde\lambda_S + \kappa_S \log_2 s - \tilde\mu(v))^2
  + \kappa_S^2\,\omega_S^2}\,\Bigr],
\]
where the corrected contribution of the grid $\widetilde{VS}$ tends to zero at the beginning of the
answer on a short text, and its shape does not depend on $\delta_{VS}$. The shift $\delta_{VS}$ is
subtracted from the coefficients of all cells of the grid and passes into the reference level
$\alpha_{\mathrm{phys}}$ (Section~\ref{sec:bias}); the logit of no trial changes.

\begin{figure}[htbp]
\centering
\includegraphics[width=\linewidth]{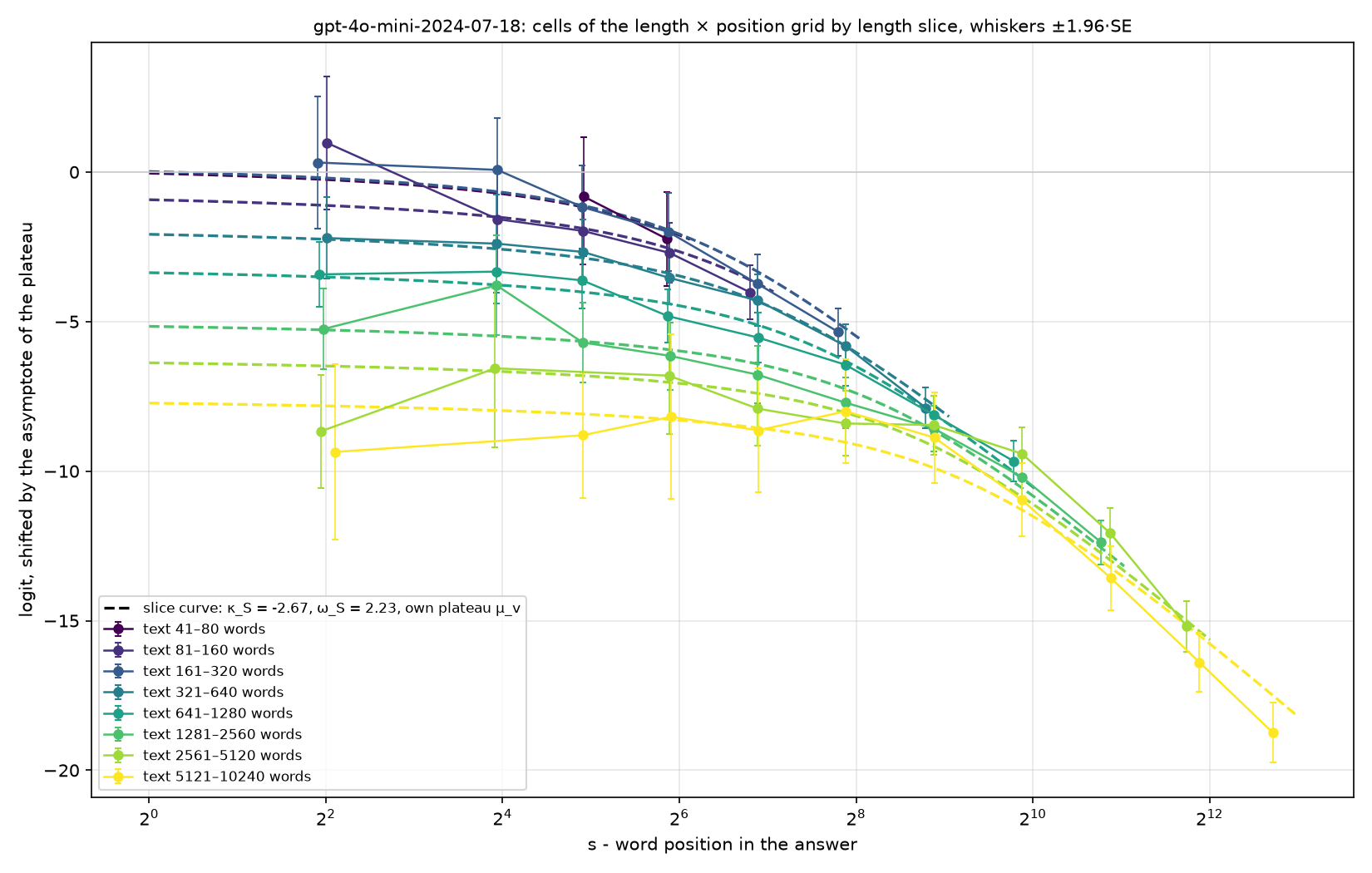}
\caption{Coefficients of the cells of the length and position grid after the shift correction:
$\delta_{VS} = \mu_0$ is subtracted from all cells. The notation is the same as in
Figure~\ref{fig:vs_raw}.}
\label{fig:vs_corrected}
\end{figure}

\begin{figure}[htbp]
\centering
\includegraphics[width=0.85\linewidth]{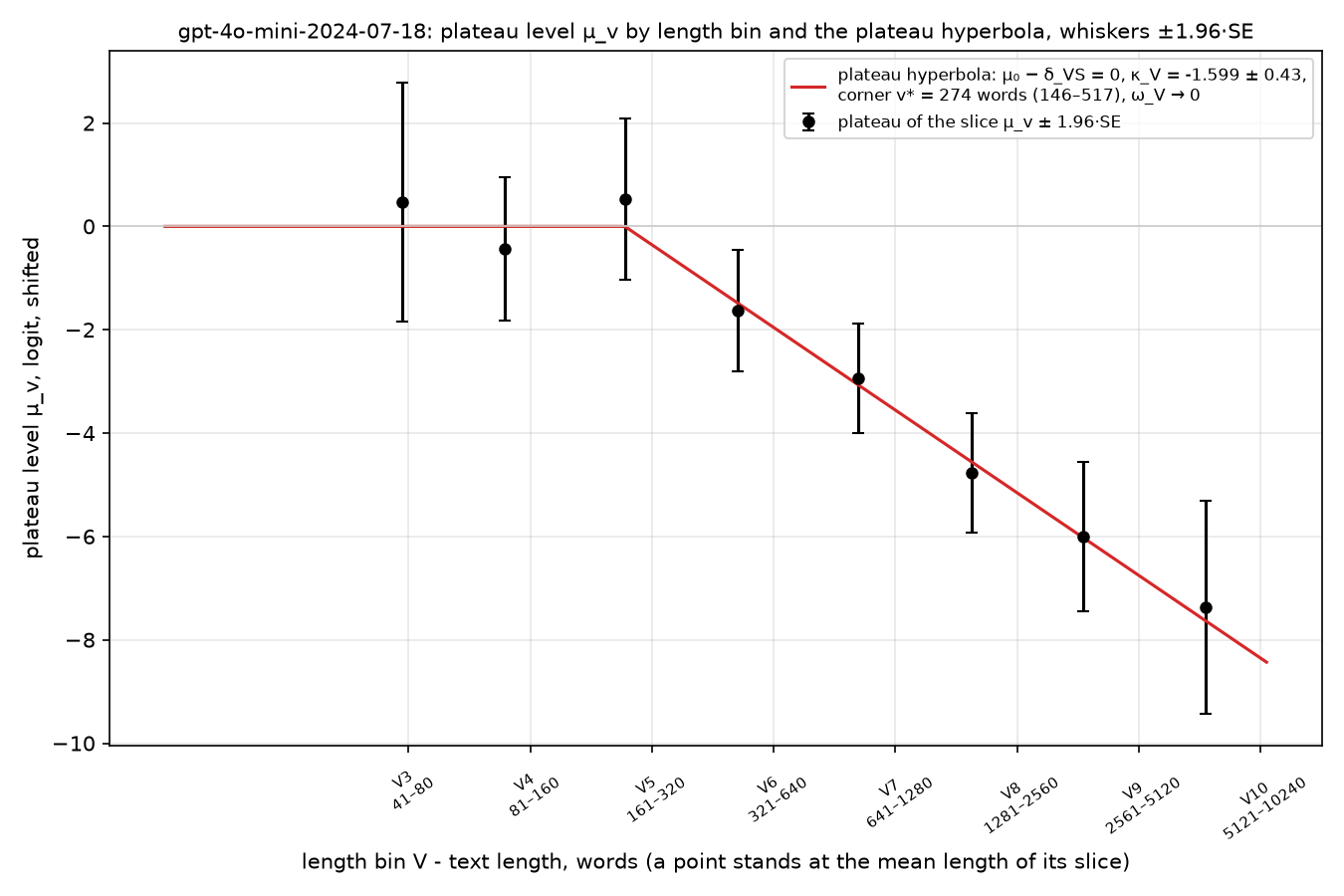}
\caption{Plateau levels of the slices after the shift correction: the hyperbola of the plateau law
tends to zero for short texts. The notation is the same as in Figure~\ref{fig:vs_plateau_raw}.}
\label{fig:vs_plateau_corrected}
\end{figure}

The corrected cells and plateau levels are shown in Figures~\ref{fig:vs_corrected}
and~\ref{fig:vs_plateau_corrected}. After the correction the plateau of short texts lies at zero, and
each cell shows the change of the contribution of the grid to the logit relative to the beginning of
the answer on a short text, with the other terms of the statistical model unchanged. At the beginning
of the answer, increasing the text length up to the transition region, whose position is estimated
by $v_* \approx 274$~words, hardly changes this contribution, while every doubling of the length of a
long text lowers it by $|\kappa_V| = 1.60 \pm 0.43$ logits. Past the corner of the hyperbola every
doubling of the word position lowers the contribution of the grid by $|\kappa_S| = 2.67 \pm 0.31$
logits, and this slope is common to all text lengths.

The grid law obtained admits a clear geometric interpretation. In the coordinates of the
contribution to the logit and the logarithm of the word position, all slices have the form of a
hyperbola with a common slanted asymptote. They differ only by the height of the horizontal plateau:
the load by the text volume $v$ cuts off the upper part of one and the same descending dependence at
the level $\mu_v$, and the hyperbola provides a smooth transition from the plateau to the slanted
part. Thus an increase of the volume does not shift the whole profile down but changes the level of
its upper bound.

The height of this bound is determined by the length of the text. Before the transition region near
$v_*$ the plateau level hardly depends on the length, and beyond it the level decreases by an
approximately constant amount with every doubling of $v$, that is, linearly in the logarithm of the
volume. Since the slanted asymptote remains unchanged, lowering the plateau simultaneously shifts the
intersection point of the asymptotes to later word positions: for the slice of 41--80~words the
asymptotes intersect around the 80th word of the answer, and for the slice of 5121--10\,240~words
around the 600th. Therefore, as the text volume grows, the plateau becomes at the same time lower
and longer. Its greater length does not testify to a greater robustness of the model to progress
through the answer but is a direct geometric consequence of lowering the cutoff level.

Beyond the transition region the profiles of all slices converge to the common slanted asymptote,
and the contribution of the grid is determined practically exclusively by the word position $s$,
regardless of the text length $v$. Consequently, the law found describes not an accumulation of two
independent losses but an interaction of two limits. The text volume sets the upper level of the
quality of instruction following, which shows itself already at the beginning of the answer, whereas
the word position sets the general regularity of its subsequent decrease. As long as the positional
limit is weaker than the volume one, the profile stays on the plateau. When it becomes stricter, the
further decrease is determined by the word position, and the initial differences between texts of
different length gradually disappear. This is precisely what explains the observed convergence of
the profiles deep in the answer.

The dependences found admit a mechanistic interpretation. The limit associated with the axis
$V$ may reflect a decrease in the reliability of processing the task as a whole as the volume of the
input data grows. Presumably, up to a certain region of load the model is able to process the
presented volume without a noticeable additional loss of quality, after which every doubling of it
is accompanied by an approximately equal decrease of the logit. Such an effect may be associated both
with limits on holding the input information and with interference between the elements of a large
input text.

The common dependence on the position $s$ is of a different nature: it shows itself as the answer
progresses and hardly depends on the full volume of the text. One possible explanation is the
growing distance between the current place of generation and the rule list located at the beginning
of the prompt. As the answer is formed this distance grows, which may make it harder to refer to the
needed rule and apply it correctly. This hypothesis agrees with the earlier established independent
informational contribution of the absolute position of the word in the full context beyond the text
length and the position in the answer (Section~\ref{sec:predictor_axes}).

Thus the results point to the existence of two qualitatively different limits of instruction
execution: a volume limit, which determines the available initial level, and a positional limit,
which determines the subsequent decrease. Their interaction is described with good accuracy by a
smoothed minimum, not by a sum of independent penalties. At the same time, the form and the
quantitative characteristics of the dependences have been established experimentally, but not the
mechanisms underlying them. The interpretations proposed require a separate experimental test.

\subsubsection{Replacement density}\label{sec:laws_p}

\begin{figure}[htbp]
\centering
\includegraphics[width=0.85\linewidth]{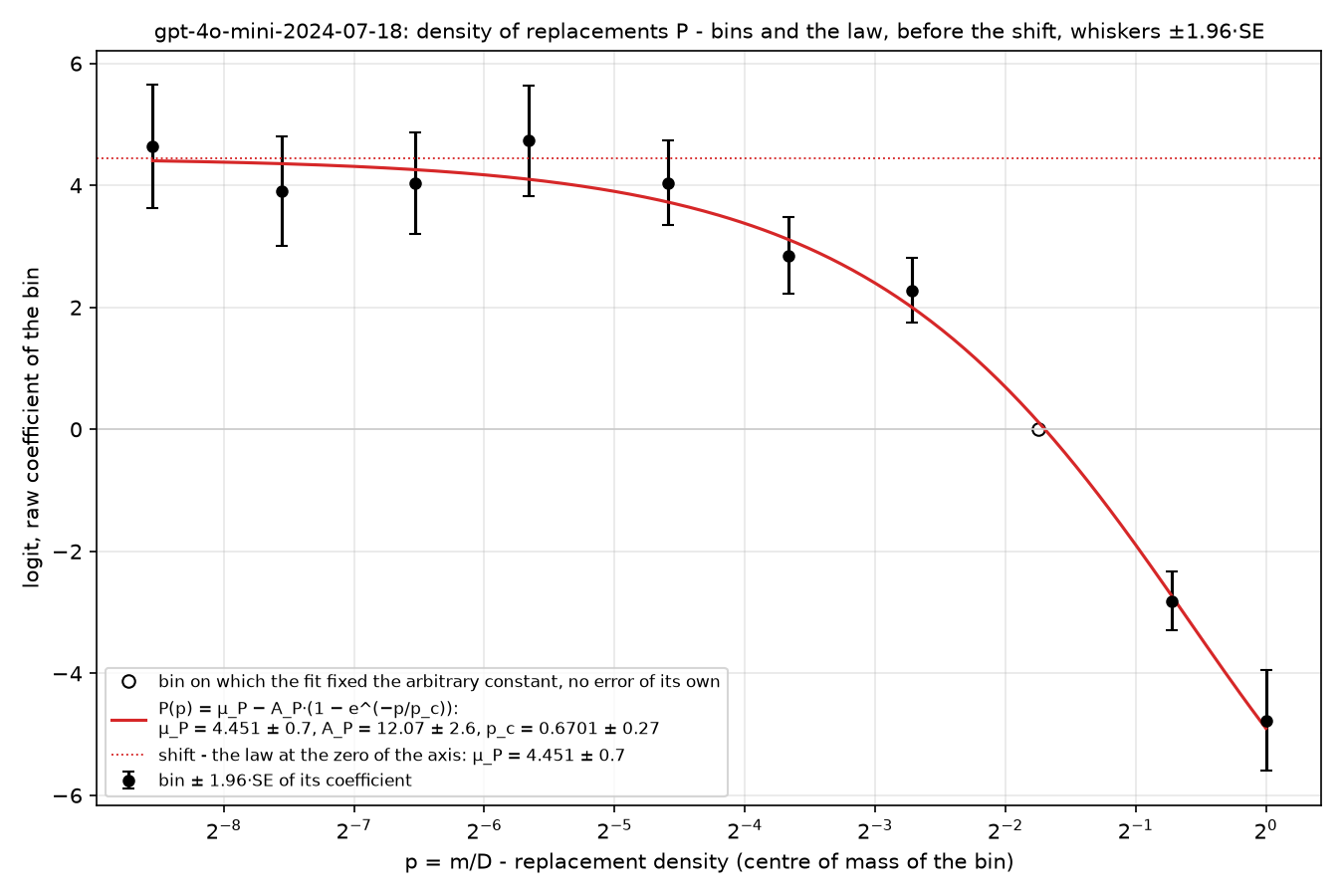}
\caption{Coefficients of the replacement density bins and the density law before the shift
correction. The open marker is the bin on which the fit pinned the arbitrary constant; the dashed
line is the level of the shift $\delta_P = \mu_P$. Whiskers are the 95\% confidence interval of the
coefficient of a bin.}
\label{fig:p_raw}
\end{figure}

The coefficients of the replacement density bins $p = m/D$ obtained by the fit are shown in
Figure~\ref{fig:p_raw}. Each bin is plotted at the mean density of its trials; the bins span the range
from the density 1/512 to the full density $p = 1$, at which every word of the text is replaced, and
none of them is excluded from the fit. As with the grid, the level at which the coefficients lie is
set by an arbitrary condition of the fit
(Section~\ref{sec:reference}): the density bin 1/4--1/2 is pinned at zero by this condition and has
no uncertainty of its own, and only the shape of the profile is meaningful.

At low density the coefficients are constant within the uncertainties: for the bins up to a density
of about 1/16 they lie between 3.9 and 4.7. From this density on the profile decreases, and at full
density the coefficient falls to $-4.8$. On the logarithmic density axis the profile has a horizontal
asymptote at rare replacements and then falls steeply. This shape is described by the law of
exponential saturation
\[
P(p) = \mu_P - A_P\bigl(1 - e^{-p/p_c}\bigr),
\]
where
\begin{itemize}
\item $\mu_P$ is the level of the horizontal asymptote to which the profile tends at rare
  replacements, $p \to 0$;
\item $A_P$ is the limit of the loss of the logit, to which the law tends as the density grows;
\item $p_c$ is the characteristic density: at $p = p_c$ the loss reaches the fraction
  $1 - e^{-1} \approx 0.63$ of $A_P$.
\end{itemize}

The law was fitted to the coefficients of nine bins with weights inverse to their full covariance, as
in Section~\ref{sec:bias}; the bin pinned by the condition of the fit does not enter the distance.
The fit gave $A_P = 12.07 \pm 2.63$ and $p_c = 0.67 \pm 0.27$, and the curve of the law in
Figure~\ref{fig:p_raw} passes through the bins within their uncertainties.

At vanishing density, $p \to 0$, the law converges asymptotically to the horizontal line $\mu_P$. The
smallest measured density is 1/512, and the limit $p \to 0$ serves as a mathematical continuation of
the law that sets the reference level. By the rule of Section~\ref{sec:bias} the shift of the density
profile equals the level of this asymptote: $\delta_P = \mu_P = 4.45 \pm 0.70$.
The normalization constant of the law is $\mu_P$, and after subtracting the shift the law takes the
form
\[
\widetilde{P}(p) = P(p) - \delta_P = -A_P\bigl(1 - e^{-p/p_c}\bigr),
\]
that is, it expresses the loss of the logit relative to rare replacements. The shift $\delta_P$ is
subtracted from the coefficients of all density bins and passes into the reference level
$\alpha_{\mathrm{phys}}$ (Section~\ref{sec:bias}); the logit of no trial changes.

\begin{figure}[htbp]
\centering
\includegraphics[width=0.85\linewidth]{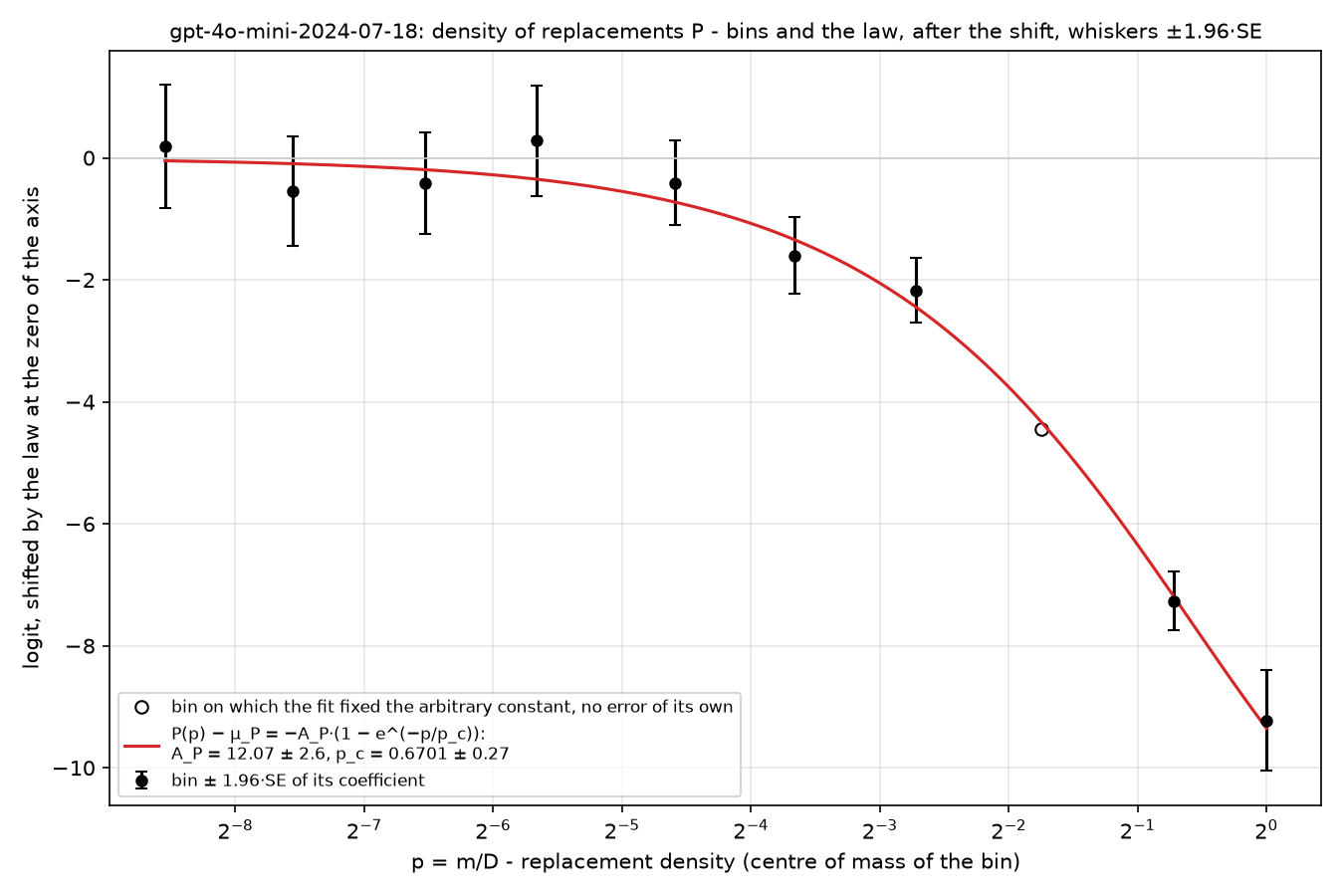}
\caption{Replacement density bins and the density law after the shift correction:
$\delta_P = \mu_P$ is subtracted from all bins. The notation is the same as in
Figure~\ref{fig:p_raw}.}
\label{fig:p_corrected}
\end{figure}

The corrected bins are shown in Figure~\ref{fig:p_corrected}. After the correction the profile at
rare replacements lies at zero, and each bin shows the change of the contribution of the density to
the logit relative to rare replacements, with the other terms of the statistical model unchanged. At
the density 1/16 the loss is about one logit, and at full density, when every word of the text is
replaced, it reaches $A_P\bigl(1 - e^{-1/p_c}\bigr) = 9.35 \pm 1.01$ logits.
This is about 78\% of the limiting loss $A_P$: saturation is not reached in the measured region, and
the loss keeps growing up to full density.

The density law admits a simple reading. At low density
$1 - e^{-p/p_c} \approx p/p_c$, and the loss is proportional to the share of the words of the text
requiring replacement: it grows at the rate $A_P/p_c \approx 18$ logits per unit of density, that is,
about 0.18 logits per percent of replaced words. Therefore on the logarithmic axis rare replacements
cost almost nothing as long as their share does not exceed a few percent, after which the loss grows
rapidly. On the linear density scale the growth of the loss slows down as full density is
approached, and the law tends to the limit $A_P$, which the measured region does not reach.

The form of the law admits an interpretation through a limited throughput. While replacements are
rare, each of them is executed almost without loss, and as the share of the words requiring the
application of rules grows, the replacements begin to compete for the limited ability of the language
model to apply rules in the course of the answer, and the loss grows. The factor $1 - e^{-p/p_c}$ has
a simple probabilistic meaning here. The words of the text are arranged in random order, and the
share of positions requiring replacement equals $p$; therefore the probability of meeting another
word to be replaced among the nearest $1/p_c$ words is approximately $1 - e^{-p/p_c}$. With the value
of $p_c$ found, this window is about one and a half words, from one to two and a half with the
uncertainty taken into account: the loss grows together with the probability that the neighboring
word also requires replacement. The saturation of the law at the level $A_P$ corresponds to the
limit of throughput: when almost every word to be replaced has another one as a neighbor, a further
growth of the density adds few new collisions.

A similar dependence has been studied for flows of many bodies along paths of limited capacity. In
single-file pedestrian movement the speed of each pedestrian is determined by the length
required---the distance to the one walking ahead, which on average is inverse to the density of the
flow---and the speed decreases as the density grows; this dependence of speed on density, called the
fundamental diagram, serves as a quantitative measure of the capacity of walking paths
\citep{seyfried2005fundamental}. In the task of the instrument the role of this distance is played by
the gap between neighboring words to be replaced, on average equal to $1/p$ words, and the quantity
$1/p_c$ sets the scale of the gap at which the loss reaches 63\% of its limit.

Separating the contribution of the density from the contribution of the text length requires care.
Density and length form the most strongly linked pair of axes of the statistical model
(Table~\ref{tab:axes_correlation}): a very small density is attainable only in a long text, so the
horizontal asymptote at rare replacements is determined mainly by long texts, and its separation
from the contribution of the length relies on the additivity of the contributions of the density and
the grid in the statistical model. The instrument measures the form of the dependence, but not its
cause, and the interpretation proposed requires a separate test.

Thus the replacement density acts as an independent load factor: the loss of the logit grows with
the share of the words requiring replacement, becomes noticeable from a density of the order of a
few percent and at full density reaches about nine logits relative to rare replacements.

\subsubsection{Number of rules}\label{sec:laws_m}

\begin{figure}[htbp]
\centering
\includegraphics[width=0.85\linewidth]{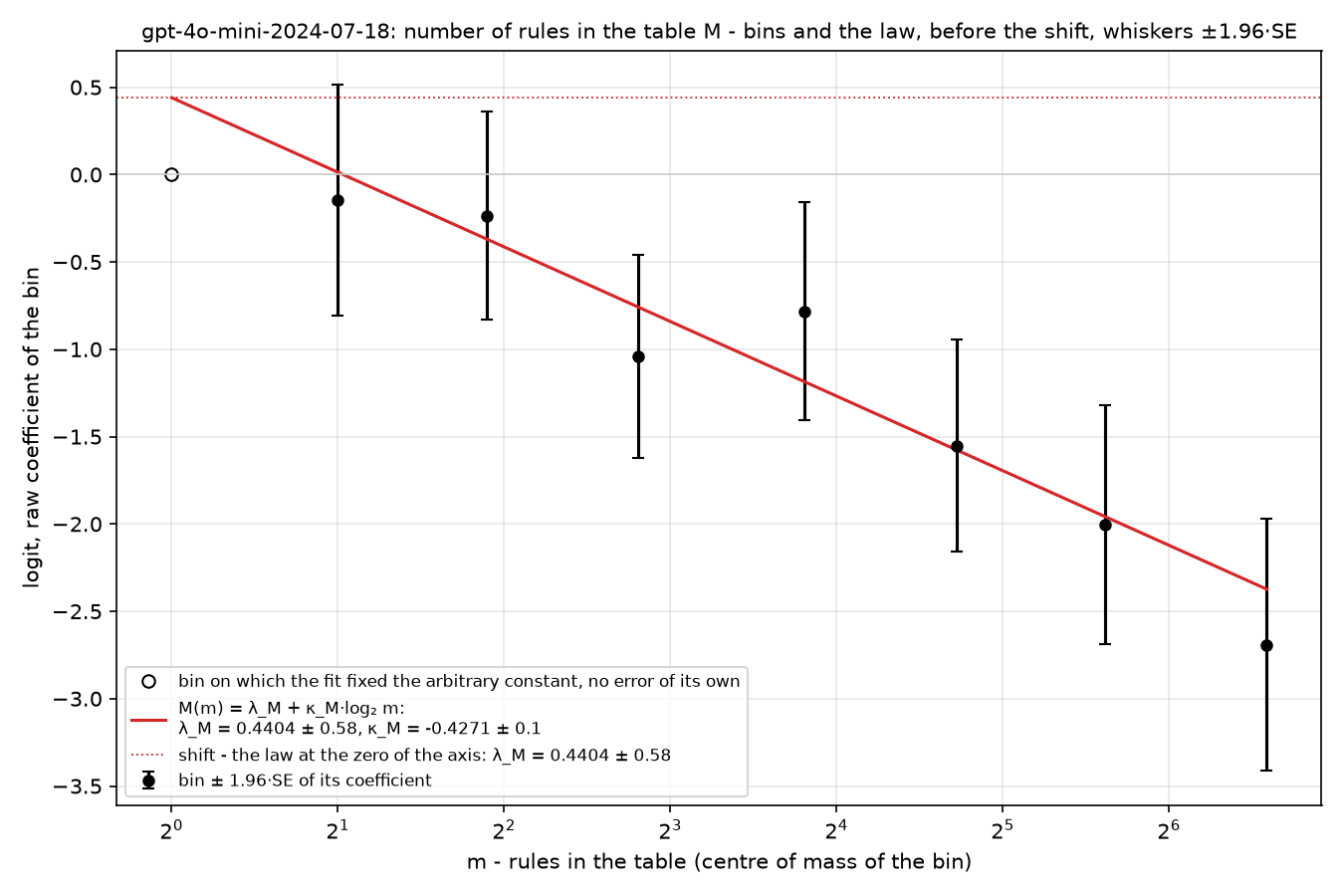}
\caption{Coefficients of the bins of the number of rules and the law of the number of rules before
the shift correction. The open marker is the bin on which the fit pinned the arbitrary constant; the
dashed line is the level of the shift $\delta_M = \lambda_M$. Whiskers are the 95\% confidence
interval of the coefficient of a bin.}
\label{fig:m_raw}
\end{figure}

The coefficients of the bins of the number of rules $m$ obtained by the fit are shown in
Figure~\ref{fig:m_raw}.
Each bin is plotted at the geometric mean of the number of rules of its trials; the bins span the
values from one to 128 rules. All nine requests with 256 rules ended with a failure of the third
kind, that is, with a complete loss of the structure of the answer; such requests describe not the
capacity but the collapse of the answer, did not enter the fit, and the axis ends with the bin of
65--128 rules. The bin of one rule is pinned at zero by an arbitrary condition of the fit
(Section~\ref{sec:reference}) and has no uncertainty of its own, and only the shape of the profile is
meaningful.

Unlike the grid and the density, the profile of the number of rules has no plateau: the
coefficients decrease from the very first bins, from $-0.15$ with two rules to $-2.69$ with 65--128
rules, and on the logarithmic axis they lie approximately on a straight line. Each doubling of the
number of rules thus costs approximately the same, and this shape is described by the law
\[
M(m) = \lambda_M + \kappa_M \log_2 m,
\]
where
\begin{itemize}
\item $\lambda_M$ is the value of the law at one rule, $m = 1$;
\item $\kappa_M$ is the slope in logits per doubling of the number of rules: by this much the logit
  changes when the rule list is twice as long.
\end{itemize}

The law was fitted to the coefficients of seven bins with weights inverse to their full covariance,
as in Section~\ref{sec:bias}; the bin pinned by the condition of the fit does not enter the distance.
The fit gave $\kappa_M = -0.43 \pm 0.10$ and $\lambda_M = 0.44 \pm 0.58$, and the straight line in
Figure~\ref{fig:m_raw} passes through the bins within their uncertainties. The continuation of the
straight line to one rule differs from the pinned bin by $\lambda_M$, which does not exceed the
uncertainty: the case of one rule obeys the same law as lists of larger
size.

The law of the number of rules is defined at the smallest load along its axis---one rule, which is
present in any task. Therefore, unlike for the grid and the density, the shift of the profile equals
not the level of an asymptote but the value of the law at this point, as the second case of the rule
of Section~\ref{sec:bias} prescribes: $\delta_M = \lambda_M = 0.44 \pm 0.58$. The normalization
constant of the law is $\lambda_M$, and after subtracting the shift the law takes the form
\[
\widetilde{M}(m) = M(m) - \delta_M = \kappa_M \log_2 m,
\]
that is, it vanishes at one rule. The shift $\delta_M$ is subtracted from the coefficients of all bins
of the number of rules and passes into the reference level $\alpha_{\mathrm{phys}}$
(Section~\ref{sec:bias}); the logit of no trial changes.

\begin{figure}[htbp]
\centering
\includegraphics[width=0.85\linewidth]{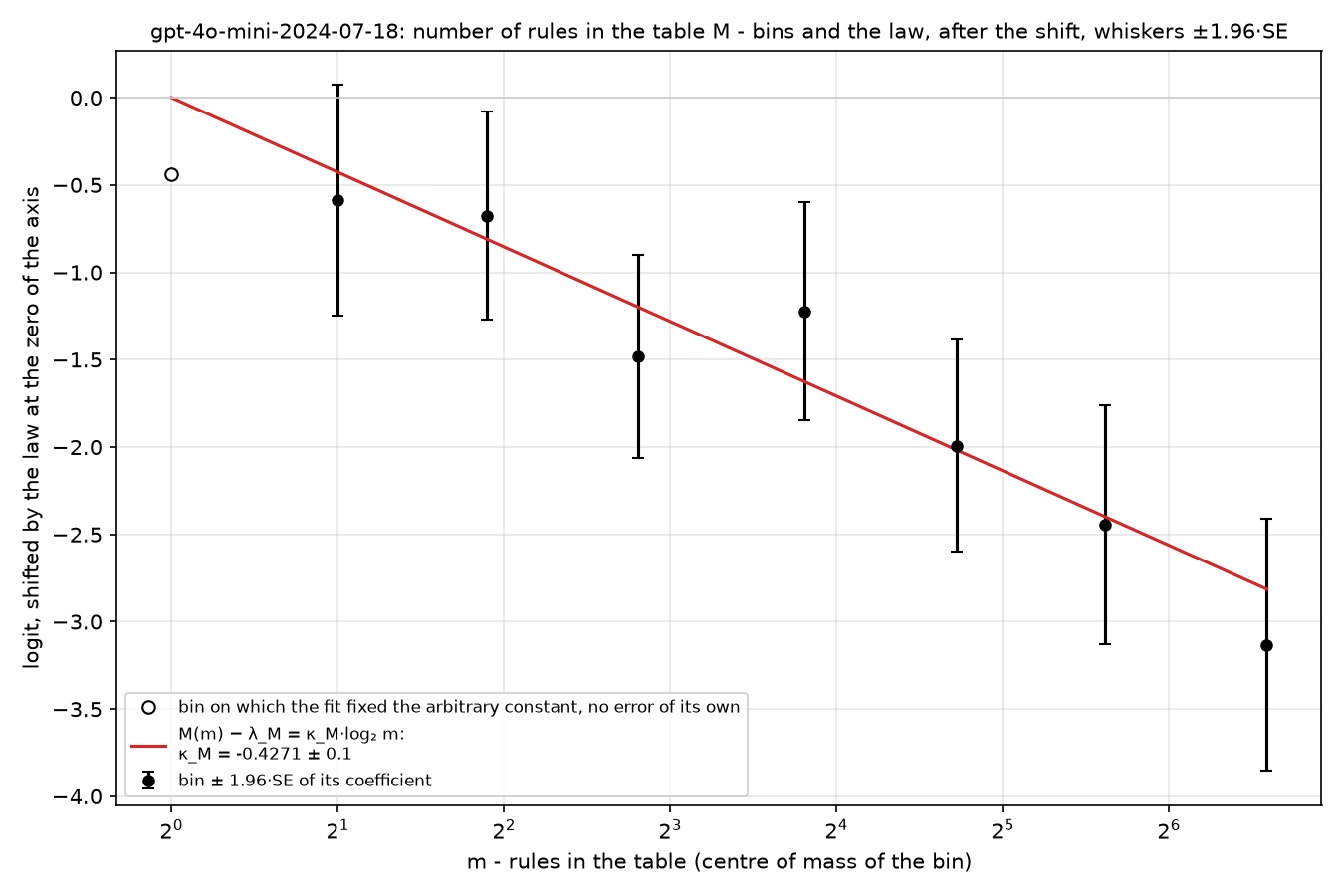}
\caption{Bins of the number of rules and the law of the number of rules after the shift correction:
$\delta_M = \lambda_M$ is subtracted from all bins. The notation is the same as in
Figure~\ref{fig:m_raw}.}
\label{fig:m_corrected}
\end{figure}

The corrected bins are shown in Figure~\ref{fig:m_corrected}. After the correction the law at one
rule lies at zero, and each bin shows the change of the contribution of the number of rules to the
logit relative to a task with one rule, with the other terms of the statistical model unchanged. Each
doubling of the rule list lowers this contribution by $|\kappa_M| = 0.43 \pm 0.10$ logits, and at
65--128 rules the decrease reaches about 2.8 logits.

The axis of the number of rules expresses the price of the size of the rule list as such. The
position of the checked rule in the list is described by the separate axis $R$, and the share of the
words requiring replacement by the axis $P$, so the slope $\kappa_M$ shows what a doubling of the list
costs at an unchanged rule number and replacement density. In works where the quality of instruction
following is measured as a function of the number of instructions
\citep{jaroslawicz2025ifscale, harada2025multiply}, these contributions are not separated: as the
number of instructions grows, both the size of the list and the numbers of the instructions checked
in it grow at the same time.

A dependence linear in the logarithm of the number of elements held simultaneously is known for
holding ``key---value'' correspondences: the recall accuracy decreases approximately linearly in the
logarithm of the number of tracked keys and of the number of their updates, which the authors
associate with a general limit on the resistance of a language model to mutual interference
\citep{wang2025unable}. One may assume that the rules of the list also interfere with each other: the
more rules there are in it, the harder it is to choose the needed one at each replacement. The measure
in that work is the share of correct answers, not the logit, so it is the form of the dependence that
coincides, not its numerical values; the instrument measures the form of the dependence, but not its
cause, and the interpretation proposed requires a separate test.

Separating the contribution of the number of rules from the contributions of other axes relies on the
additivity of the statistical model. The number of rules is moderately linked with the text length,
the replacement density and the rule number (Table~\ref{tab:axes_correlation}), and with the rule
number it is also linked by the construction of the task: the number of the checked rule does not
exceed their number.

Thus the size of the rule list acts as an independent load factor: each doubling of the list lowers
the logit of a right replacement by one and the same amount, about 0.43, from one rule to 128.

\subsubsection{Rule number}\label{sec:laws_r}

\begin{figure}[htbp]
\centering
\includegraphics[width=0.85\linewidth]{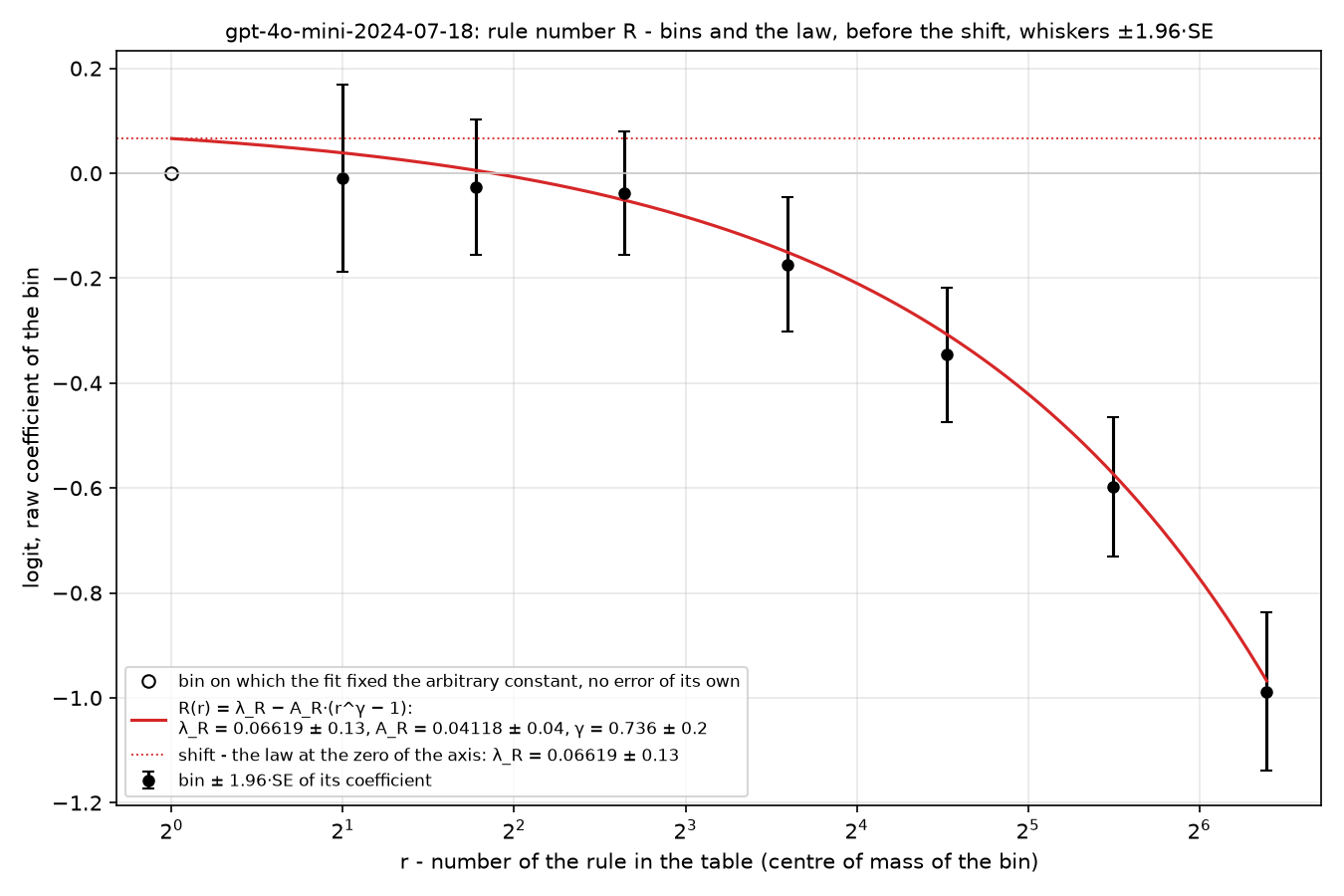}
\caption{Coefficients of the bins of the rule number and the law of the rule number before the shift
correction. The open marker is the bin on which the fit pinned the arbitrary constant; the dashed
line is the level of the shift $\delta_R = \lambda_R$. Whiskers are the 95\% confidence interval of
the coefficient of a bin.}
\label{fig:r_raw}
\end{figure}

The coefficients of the bins of the rule number $r$---the position of the checked rule in the
list---are shown in Figure~\ref{fig:r_raw}. The bins are the same as for the number of rules, from
the first rule to the bin of the 65th--128th, and each bin is plotted at the geometric mean of the
rule numbers of its trials. The order of the lines of the rule list is randomly permuted by the order
seed (Section~\ref{sec:task}), so the contribution of the position of a rule is separated from the
difficulty of a particular pair of words. The bin of the first rule is pinned at zero by an arbitrary
condition of the fit (Section~\ref{sec:reference}) and has no uncertainty of its own, and only the
shape of the profile is meaningful.

The first eight rules of the list cost almost nothing: the coefficients of the bins up to the eighth
rule lie between $-0.01$ and $-0.04$. Further on the decline accelerates and for the rules from the
65th to the 128th reaches $-0.99$. Unlike the number of rules, on the logarithmic axis the profile of
the rule number is not a straight line: each next doubling of the number costs more than the previous
one. This shape is described by the power law
\[
R(r) = \lambda_R - A_R\bigl(r^{\gamma} - 1\bigr),
\]
where
\begin{itemize}
\item $\lambda_R$ is the value of the law at the first rule, $r = 1$;
\item $A_R$ is the scale of the degradation of late rules in logits;
\item $\gamma$ is the exponent of the growth of degradation: at numbers much larger than one, a
  doubling of the number increases the degradation by a factor of $2^{\gamma}$.
\end{itemize}

The law was fitted to the coefficients of seven bins with weights inverse to their full covariance,
as in Section~\ref{sec:bias}; the bin pinned by the condition of the fit does not enter the distance.
The fit gave $A_R = 0.041 \pm 0.040$ and $\gamma = 0.74 \pm 0.20$, that is,
$2^{\gamma} = 1.67 \pm 0.23$, and the curve in Figure~\ref{fig:r_raw} passes through the bins within
their uncertainties. The parameters $A_R$ and $\gamma$ separately are determined coarsely, and the
uncertainty of the scale almost equals its value; the degradation of late rules itself, however, is
clearly visible in the bins.

The law of the rule number, like the law of the number of rules, is defined at the smallest load
along its axis---the first rule, which is present in any task. Therefore by the second case of the
rule of Section~\ref{sec:bias} the shift of the profile equals the value of the law at this point:
$\delta_R = \lambda_R = 0.07 \pm 0.13$, and it does not differ from zero. The normalization constant
of the law is $\lambda_R$, and after subtracting the shift the law takes the form
\[
\widetilde{R}(r) = R(r) - \delta_R = -A_R\bigl(r^{\gamma} - 1\bigr),
\]
that is, it vanishes at the first rule. The shift $\delta_R$ is subtracted from the coefficients of all
bins of the rule number and passes into the reference level $\alpha_{\mathrm{phys}}$
(Section~\ref{sec:bias}); the logit of no trial changes.

\begin{figure}[htbp]
\centering
\includegraphics[width=0.85\linewidth]{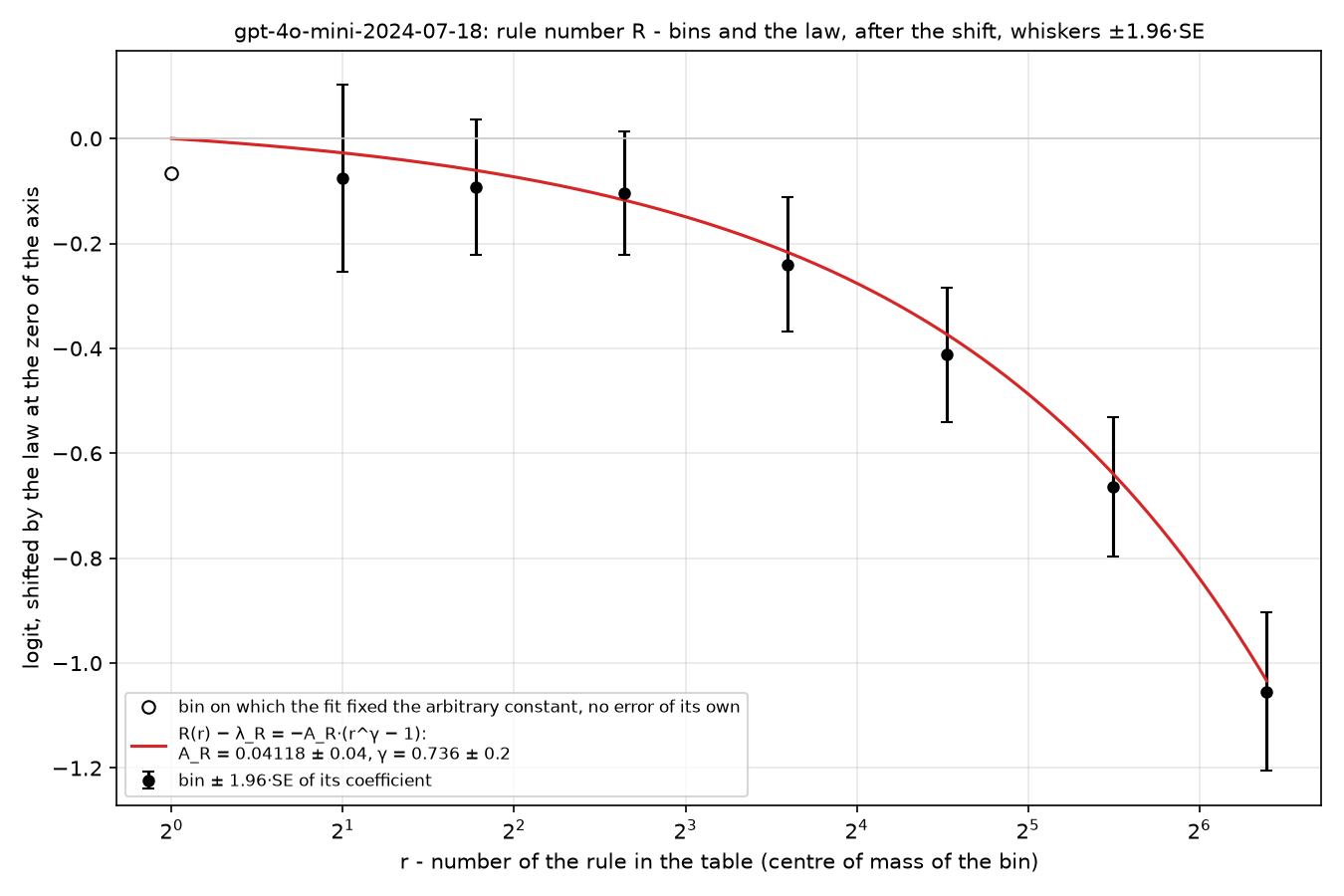}
\caption{Bins of the rule number and the law of the rule number after the shift correction:
$\delta_R = \lambda_R$ is subtracted from all bins. The notation is the same as in
Figure~\ref{fig:r_raw}.}
\label{fig:r_corrected}
\end{figure}

The corrected bins are shown in Figure~\ref{fig:r_corrected}. After the correction the law at the
first rule lies at zero, and each bin shows the change of the contribution of the rule number to the
logit relative to the first rule of the list, with the other terms of the statistical model
unchanged. The rules from the 65th to the 128th are executed worse than the first by
$1.05 \pm 0.12$ logits, and each doubling of the number increases this loss approximately by a
factor of 1.7.

Worse execution of the late instructions of a list has been noted in other works as well. In IFScale
the error share in the last third of the instruction list is higher than in the first, and this
difference is most noticeable at 150--200 instructions \citep{jaroslawicz2025ifscale}; for GPT-3.5
and GPT-4 the keywords standing earlier in the prompt are fulfilled more often, while for some other
models the opposite effect prevails \citep{li2024divide}. The dependence on the position in the input
is not necessarily monotonic: in tasks of retrieving information from a long context the accuracy is
highest when the needed information stands at the beginning or at the end of the input, and lowest in
the middle \citep{liu2023lost}. For GPT-4o mini there is no recovery towards the end of the rule
list: the last rules are executed worst, although in the prompt they stand closest to the text.

One may assume that every rule standing earlier in the list creates interference for the subsequent
ones, and that this interference accumulates with the rule number: the more rules precede the checked
one, the harder it is to apply. The instrument measures the form of the dependence, but not its
cause, and the interpretation proposed requires a separate test. Separating the contribution of the
rule number from the contribution of the number of rules relies on the additivity of the statistical
model: the rule number is the least linked of all axes with the others
(Table~\ref{tab:axes_correlation}), but with the number of rules it is linked by the construction of
the task, since the number of the checked rule does not exceed their number.

Thus the position of a rule in the list acts as an independent load factor: the first rules are
executed almost without loss, while the degradation of late ones grows at an accelerating rate,
approximately by a factor of 1.7 per doubling of the number. Together with the law of the number of
rules, the law of the rule number sets the price of each next rule, through which the instruction
capacity is computed (Section~\ref{sec:quantities}).

\subsection{Collapse of the answer}\label{sec:collapse}

A collapse of the answer is a request whose answer ended with a failure of the third kind
(Section~\ref{sec:validation}), that is, with a loss of the structure of the answer. A failure of the
third kind takes two forms. In the first, the model skips, merges or adds lines in the middle of the
answer, and the lines following them end up not at their numbers. In the second, the answer is cut
off: it is incomplete, and the last lines of the text are missing from it. In both cases we will say
that the request lost the text. All replacements after the place of the failure, including those in
the missing lines, are counted as unsuccessful. In the GPT-4o mini measurement 995 requests out of
2958 lost the text, that is,
33.6\%.

A wrong replacement is an event of an individual trial, whereas a collapse is an event of the request
as a whole: after the place where the text is lost, the outcome of every trial is zero, whatever rule
is checked in it. Therefore a collapse has two measurable characteristics of its own: the probability
that a request loses the text, and the place in the answer at which this happens. Each of them is
described by its own law: the collapse rate in Section~\ref{sec:collapse_rate}, the place of the
collapse in Section~\ref{sec:collapse_place}.

\subsubsection{Collapse rate}\label{sec:collapse_rate}

Let $\psi$ denote the probability that a request loses the text. Each request is treated as one
Bernoulli trial with the outcome ``text lost'' or ``text kept'', and the law of the collapse rate
describes how $\psi$ depends on the load of the request. The fit of the law also included requests
that did not enter the fit of the
statistical model: such requests describe not the capacity but the collapse of the answer
(Section~\ref{sec:laws_m}), and it is precisely for the collapse law that they are informative.

The form of the law was chosen from the binned representation: a logistic model of the outcomes of
requests with bins of the text length and the replacement density (Section~\ref{sec:binned}). The
reference bins are the length bin of 41--80 words, the shortest of the bins in which both outcomes
occur, and the density bin 1/4--1/2, as in the statistical model. Bins all of whose requests have one
outcome have no estimate and are excluded together with their requests: texts up to 40 words (35
requests, all kept the text) and the density 1/1024--1/512 (6 requests, all lost it). The
coefficients of the length bins grow approximately linearly with the bin number, that is, linearly in
the logarithm of the length. The coefficients of the density bins are almost constant at densities
up to 1/32 and then grow steeply towards full density. If bins of the number of rules are added to
the model, their coefficients lie within $\pm 0.5$ logits of zero and do not change systematically
with the number of rules: only for one bin out of eight, 5--8 rules, does the 95\% interval not
include zero, which with eight bins agrees with random scatter.
This shape is described by the law
\[
\ln\frac{\psi}{1 - \psi} = \lambda_\psi + \kappa_\psi \log_2\frac{v}{20} + A_\psi\, p^{\gamma_\psi},
\]
where
\begin{itemize}
\item $\lambda_\psi$ is the logit of the collapse at the shortest text of the measurement, $v = 20$
  words, and vanishing density, $p \to 0$;
\item $\kappa_\psi$ is the slope in logits per doubling of the text length;
\item $A_\psi$ is the contribution of the density to the logit of the collapse at full density,
  $p = 1$;
\item $\gamma_\psi$ is the exponent that sets how fast the contribution of the density grows with
  $p$.
\end{itemize}

The law was built in the same way as the laws of the axes of the statistical model. First, the
coefficients of the length and density bins were found by maximum likelihood from the outcomes of
the requests. The requests are distributed unevenly over the load space
(Section~\ref{sec:analysis}), so each request entered the likelihood with a weight inverse to the
number of requests in its cell---the combination of the bins of length, density and number of rules.
With such weights all cells contribute equally to the fit, and the law describes the language model,
not the plan of the experiment. The uncertainties of the coefficients were computed from the
sandwich covariance, which remains consistent also with a weighted likelihood
\citep{cameron2015practitioner}.

Then the law of each axis was drawn through the coefficients of its bins: the straight line
$V_\psi(v) = \lambda_{V\psi} + \kappa_\psi \log_2(v/20)$ through the length bins and the curve
$P_\psi(p) = \mu_{P\psi} + A_\psi\, p^{\gamma_\psi}$ through the density bins. The density law was
fitted with weights inverse to the full covariance of the bins, as in Section~\ref{sec:bias}. The
length bins are counted from one reference bin and are strongly correlated through it, and with such
weights the line drifted away from all the bins; therefore, as for the grid
(Section~\ref{sec:laws_vs}), the weights of the length law were the inverses of the variances of the
bins, and the uncertainties were computed from the full covariance by the delta method. By the rule
of Section~\ref{sec:bias} the shift of the length equals the value of the law at the smallest length
of the measurement, $v = 20$ words: $\lambda_{V\psi} = -3.64 \pm 2.12$. The shift of the density
equals the horizontal asymptote of the law as $p \to 0$: $\mu_{P\psi} = -5.33 \pm 0.76$. Both shifts
pass into the level $\lambda_\psi = \alpha_\psi + \lambda_{V\psi} + \mu_{P\psi}$, where
$\alpha_\psi$ is the free term of the binned model. The fit gave $\lambda_\psi = -20.00 \pm 1.58$,
$\kappa_\psi = 2.40 \pm 0.19$, $A_\psi = 12.25 \pm 1.01$ and
$\gamma_\psi = 0.71 \pm 0.09$.

\begin{figure}[htbp]
\centering
\includegraphics[width=0.85\linewidth]{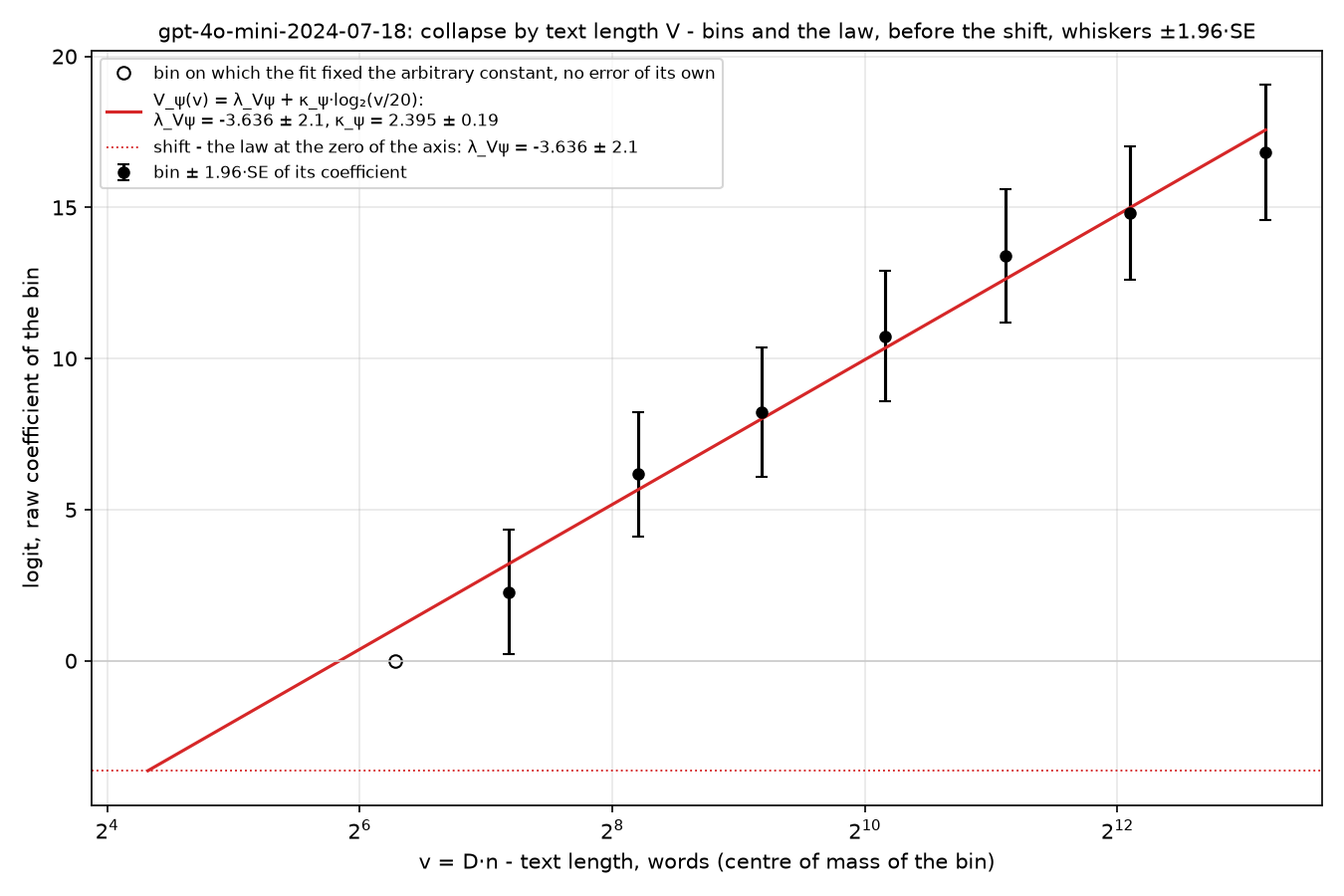}
\caption{Coefficients of the text length bins in the logit of the collapse and the length law before
the shift correction. The open marker is the bin on which the fit pinned the arbitrary constant; the
dashed line is the level of the shift $\lambda_{V\psi}$. Whiskers are the 95\% confidence interval
of the coefficient of a bin.}
\label{fig:collapse_v_raw}
\end{figure}

\begin{figure}[htbp]
\centering
\includegraphics[width=0.85\linewidth]{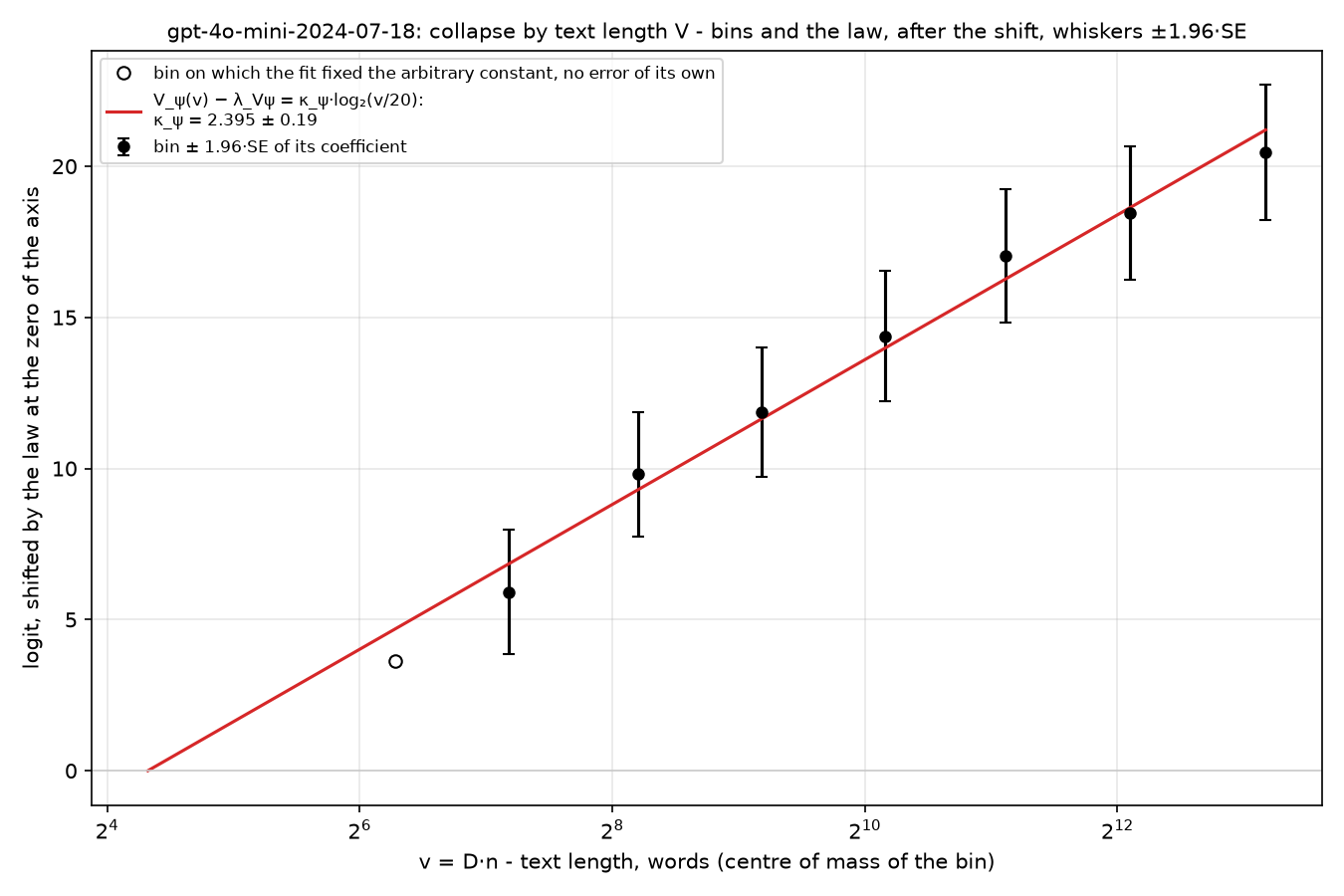}
\caption{Text length bins and the length law after the shift correction: $\lambda_{V\psi}$ is
subtracted from all bins. The notation is the same as in Figure~\ref{fig:collapse_v_raw}.}
\label{fig:collapse_v_corrected}
\end{figure}

\begin{figure}[htbp]
\centering
\includegraphics[width=0.85\linewidth]{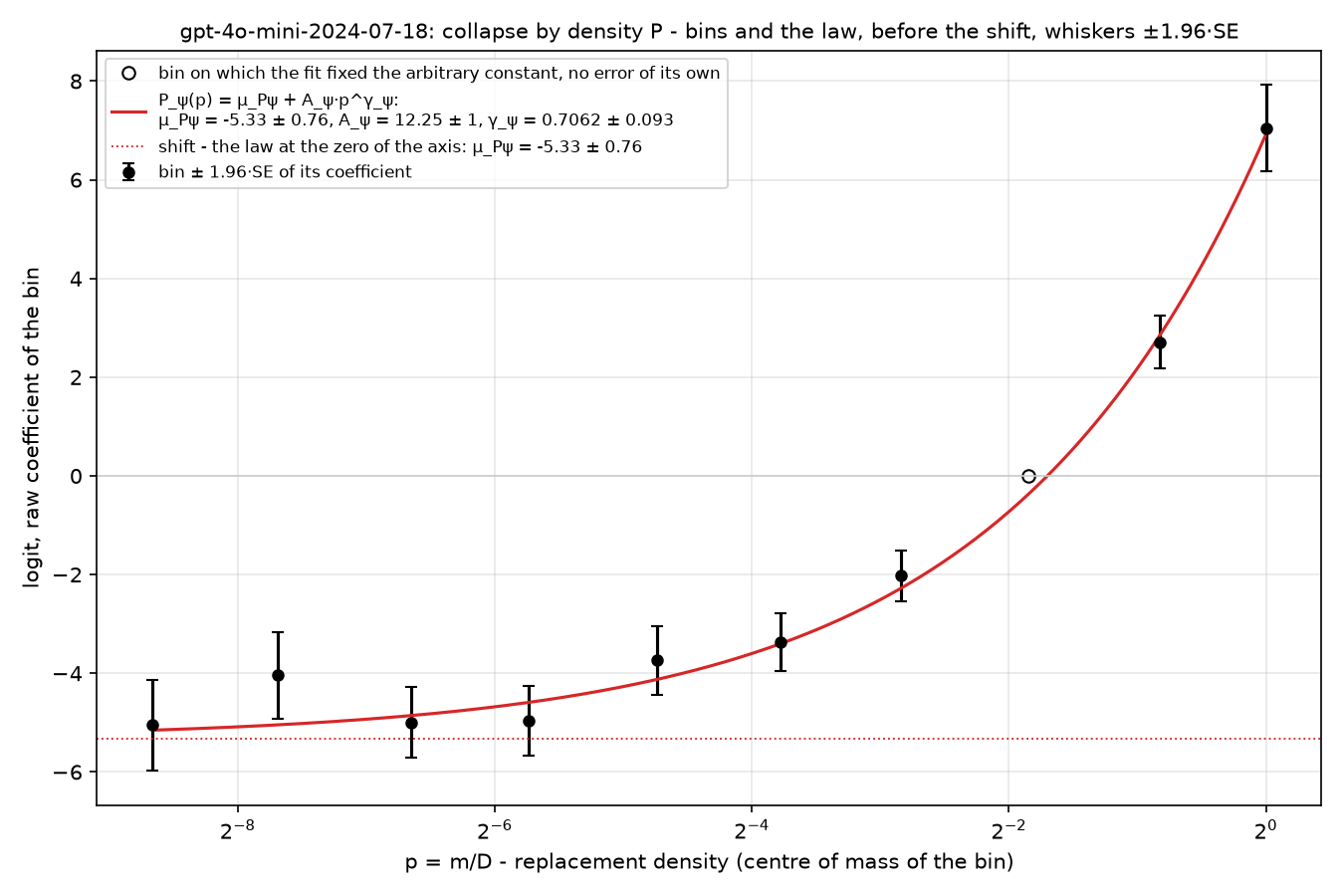}
\caption{Coefficients of the replacement density bins in the logit of the collapse and the density
law before the shift correction. The notation is the same as in Figure~\ref{fig:collapse_v_raw}; the
dashed line is the level of the shift $\mu_{P\psi}$.}
\label{fig:collapse_p_raw}
\end{figure}

\begin{figure}[htbp]
\centering
\includegraphics[width=0.85\linewidth]{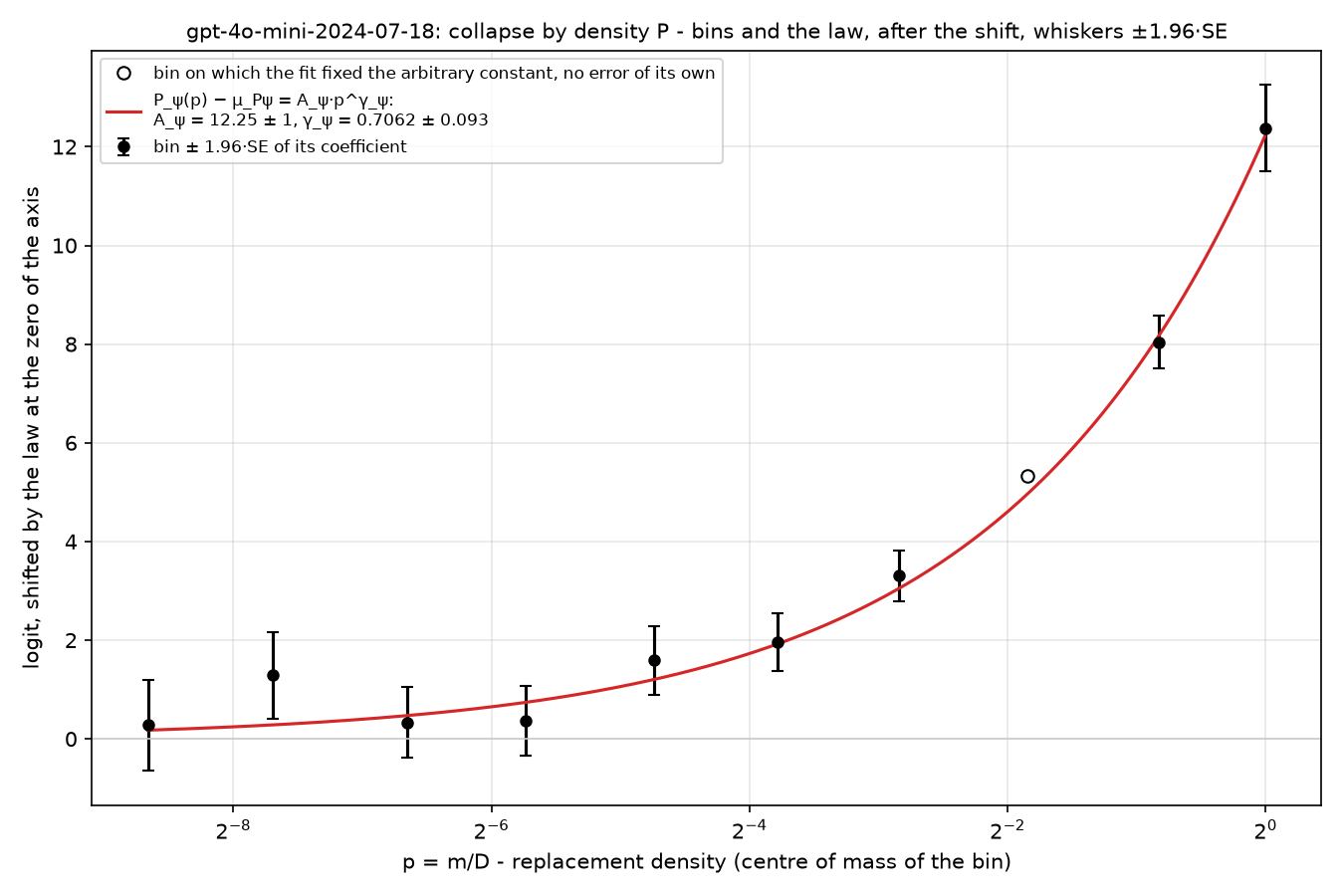}
\caption{Replacement density bins and the density law after the shift correction: $\mu_{P\psi}$ is
subtracted from all bins. The notation is the same as in Figure~\ref{fig:collapse_v_raw}.}
\label{fig:collapse_p_corrected}
\end{figure}

The coefficients of the bins and the laws of the axes are shown before the shift correction in
Figures~\ref{fig:collapse_v_raw} and~\ref{fig:collapse_p_raw}, and after the correction in
Figures~\ref{fig:collapse_v_corrected} and~\ref{fig:collapse_p_corrected}. Each doubling of the text
length increases the logit of the collapse by $\kappa_\psi$, that is, multiplies the odds of the
collapse approximately by $e^{2.40} \approx 11$. The contribution of the density is about 0.15 logits
at the density 1/512 and about 1.7 logits at the density 1/16, and at full density it reaches
$A_\psi = 12.3$ logits. Each doubling of the density multiplies this contribution by
$2^{\gamma_\psi} \approx 1.6$, and the main growth of the contribution falls on densities above 1/16.

The number of rules does not enter the law, although the share of collapses taken over the requests
with a given number of rules grows with the number of rules. The growth is explained by the
construction of the task: at one and the same density a task with more rules contains more distinct
words, $D = m/p$, and therefore has a longer text, $v = Dn$. All nine requests with 256 rules lost the
text. Consequently, at a given text length and replacement density the number of rules does not
affect the probability of the collapse.

The length law admits a simple probabilistic reading. If the model lost the text on each line of the
answer with one and the same small probability, independently of the other lines, the probability of
the collapse, while it is small, would be proportional to the number of lines, and each doubling of
the text length would increase the logit of the collapse by $\ln 2 \approx 0.69$. The measured slope
$\kappa_\psi$ is 3.5 times larger: the odds of the collapse grow approximately in proportion to the
text length to the power 3.5, $(v/20)^{\kappa_\psi/\ln 2} = (v/20)^{3.5}$. Consequently, the risk of
losing the text on the next line is not constant but grows---either as the answer progresses or
with the volume of the whole text. Which of these causes acts is shown by the place of the collapse
(Section~\ref{sec:collapse_place}).

The independence of the collapse from the number of rules separates it from the limit that the
instruction capacity measures. A doubling of the rule list lowers the logit of a right replacement by
$|\kappa_M| = 0.43$ (Section~\ref{sec:laws_m}), but does not change the probability of losing the
text at an unchanged length and density. The collapse is determined by the load of the text---its
length and the share of the words requiring replacement---and not by the number of instructions
held. The rules determine which word to write in place of the word to be replaced, but not where in
the text the model is; apparently, applying rules and keeping the place in the text load different
sides of the work of the model.

The dependences found admit a mechanistic interpretation. The task of the instrument requires
rewriting the text line by line, and the model composes most of the answer by copying the input. In
transformers, copying previously encountered sequences is performed by induction heads: such a head
finds the place in the input where the current token and the few preceding ones occurred, and raises
the probability of the token that stood next there, by the scheme $[A][B] \ldots [A] \to [B]$
\citep{olsson2022induction}. With such a mechanism the place in the text is determined by the
coincidence of the end of the answer already written with a fragment of the input.

This mechanism explains the dependence of the collapse on the density. The replacement words do not
occur in the source text, so each such word written in the answer breaks the coincidence of the last
words of the answer with the corresponding fragment of the input text. In a sparse text such breaks
are isolated: the answer repeats the input almost verbatim, and the place
in the text is found by the neighboring copied words. As the density grows, the share of the words of
the answer absent from the text grows together with $p$, and at the density $p = 1$ not a single word
of the answer occurs in the input, and the place in the text has to be held without relying on a
verbatim match. This may explain the steep growth of the contribution of the density as full density
is approached.

The dependence on the length agrees with observations of long generation. In LongGenBench the
fulfilment of requirements tied to places of the answer worsens as the answer grows longer, and on the
transition from 16 to 32 thousand tokens the models more often fail to finish the answer
\citep{wu2024longgenbench}; an unfinished answer corresponds to the second form of a failure of the
third kind. For
GPT-4o mini the loss of the text becomes frequent already at text lengths of the order of a thousand
words, and the slope $\kappa_\psi$ serves as a quantitative measure of this effect. The instrument
measures the form of the dependence, but not its cause, and the interpretations proposed require a
separate test.

Thus the collapse rate is determined by the load of the text: the odds of losing the text grow
approximately in proportion to its length to the power 3.5 and rise steeply with the replacement
density above
1/16, while the number of rules
at a given length and density does not affect them. The collapse of the text and the instruction
capacity are different limits of a language model.

\subsubsection{Place of the collapse}\label{sec:collapse_place}

Let $F$ denote the probability that for a request that lost the text, the text is already lost by
the word with number $s$. The place where the text is lost is the line from which the answer stops
following the text line by line: the first line of the text after the last line of the answer that
stands at its own number, if further on in the answer there are lines of the text with shifted
numbers, and for a cut-off answer the first missing line. Each word to be replaced of a request that
lost the text is treated as a Bernoulli trial with the outcome 1 if its line does not precede the
place where the text is lost, and 0 otherwise. The fit included all 995 requests that lost the text
and their 365\,955 words to be replaced, of which 49.2\% stand at or after the place where the text
is lost.

The probability $F$ is described in the same way as the collapse rate. The logit of $F$ is decomposed
by the bins of three axes: the word number $s$ (the position bins of the grid,
Section~\ref{sec:laws_vs}), the replacement density $p$ and the share of the text passed before the
word, $(s - 1)/v$, divided into ten equal bins. The reference bins are the bin of words 641--1280,
which has the most trials, the density bin 1/4--1/2 and the first bin of the text share; there are no
bins with one outcome. Each request entered the likelihood with a weight inverse to the number of
requests that lost the text in its cell, and the uncertainties of the coefficients were computed from
the cluster sandwich covariance with the cluster ``one request'' \citep{cameron2015practitioner}.

The law of each axis was drawn through the coefficients of its bins. The profile of the word number
has a horizontal plateau at the beginning of the answer and a slanted rise after the corner, that is,
the same form as the plateau level of the grid as a function of the text length
(Section~\ref{sec:laws_vs}), but with a rise instead of a decline. It is described by the hyperbola
\[
S_F(s) = \mu_{SF} + \frac{\kappa_{SF}}{2}\Bigl[\log_2\frac{s}{s_*}
  + \sqrt{\log_2^2\frac{s}{s_*} + \omega_{SF}^2}\,\Bigr],
\]
where
\begin{itemize}
\item $\mu_{SF}$ is the level of the horizontal asymptote to which the profile tends at the beginning
  of the answer;
\item $\kappa_{SF}$ is the slope of the slanted asymptote in logits per doubling of the word number;
\item $s_*$ is the word number at which the asymptotes intersect;
\item $\omega_{SF}$ is the width of the corner in doublings of the word number.
\end{itemize}
The profile of the density is described by the same power curve as in the law of the collapse rate,
$P_F(p) = \mu_{PF} + A_F\, p^{\gamma_F}$, and the profile of the text share by the straight line
$Q_F = \lambda_{QF} + \kappa_{QF}\,(s - 1)/v$. All three laws were fitted with weights inverse to the
full covariance of the bins, as in Section~\ref{sec:bias}. The width of the corner, like the width of
the transition of the plateau of the grid, is not determined by the data: its estimate converges to
the lower boundary of the search region and is held there, so no noticeable rounding of the corner
was found.

By the rule of Section~\ref{sec:bias} the shifts of the laws of the word number and the density equal
their horizontal asymptotes: $\mu_{SF} = -5.15 \pm 1.37$ at the beginning of the answer and
$\mu_{PF} = -3.99 \pm 1.53$ as $p \to 0$. The shift of the text share equals the value of the straight
line at the beginning of the text: $\lambda_{QF} = -0.13 \pm 0.64$. All three shifts pass into the
level $\lambda_F = \alpha_F + \mu_{SF} + \mu_{PF} + \lambda_{QF}$, where $\alpha_F$ is the free term of
the binned model, and the law of the place of the collapse takes the form
\[
\ln\frac{F}{1 - F} = \lambda_F + \frac{\kappa_{SF}}{2}\Bigl[\log_2\frac{s}{s_*}
  + \sqrt{\log_2^2\frac{s}{s_*} + \omega_{SF}^2}\,\Bigr] + A_F\, p^{\gamma_F}
  + \kappa_{QF}\,\frac{s - 1}{v}.
\]
The fit gave $\lambda_F = -10.65 \pm 1.99$, $\kappa_{SF} = 1.14 \pm 0.21$,
$s_* \approx 40$~words (from 22 to 70), $A_F = 6.95 \pm 1.56$, $\gamma_F = 0.42 \pm 0.16$ and
$\kappa_{QF} = 3.19 \pm 0.59$; at the same time $s_*$ estimates the position of the corner within the
accepted law, not a sharp threshold established by the measurement.

\begin{figure}[htbp]
\centering
\includegraphics[width=0.85\linewidth]{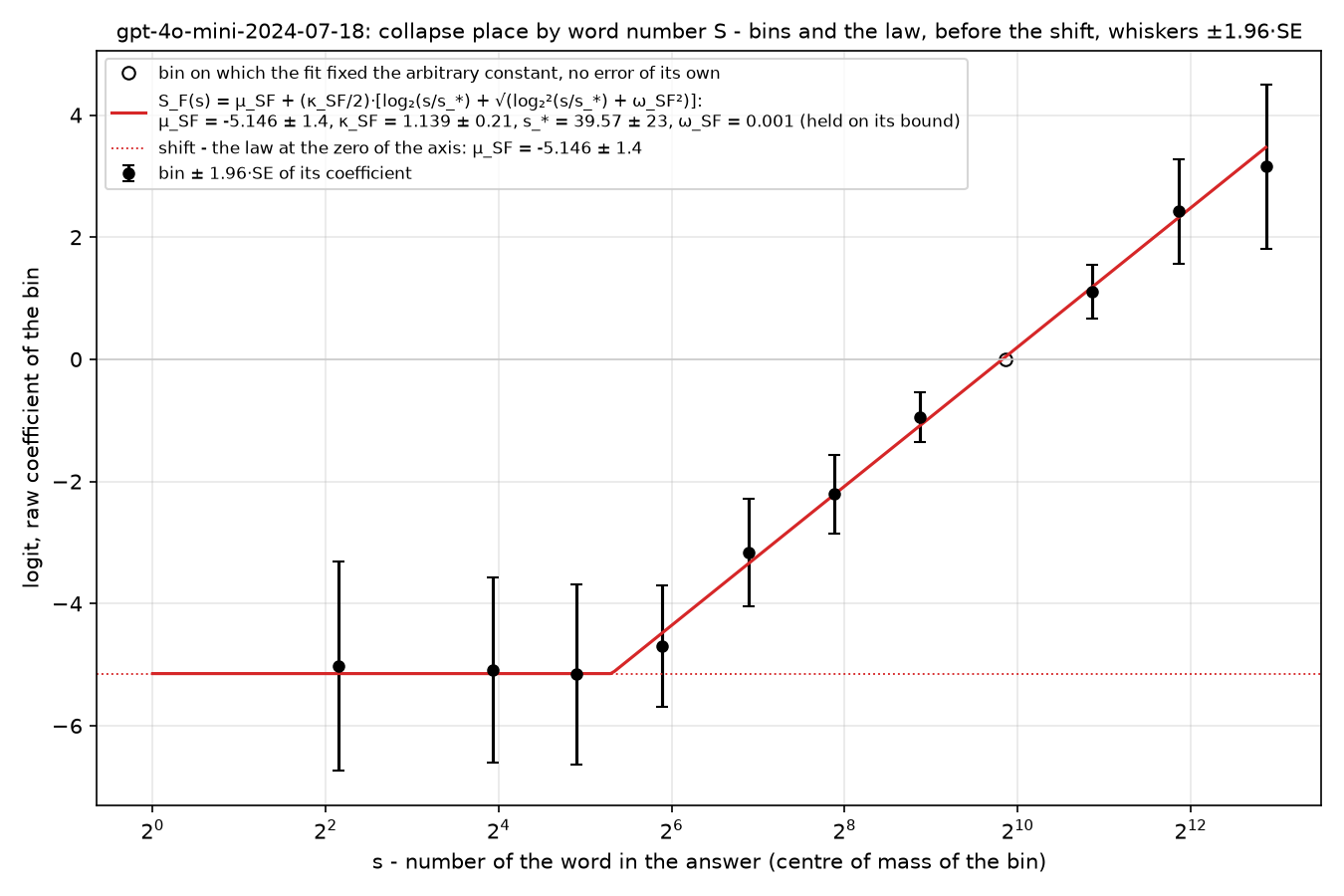}
\caption{Coefficients of the word number bins in the logit of the place of the collapse and the word
number law before the shift correction. The open marker is the bin on which the fit pinned the
arbitrary constant; the dashed line is the level of the shift $\mu_{SF}$. Whiskers are the 95\%
confidence interval of the coefficient of a bin.}
\label{fig:place_s_raw}
\end{figure}

\begin{figure}[htbp]
\centering
\includegraphics[width=0.85\linewidth]{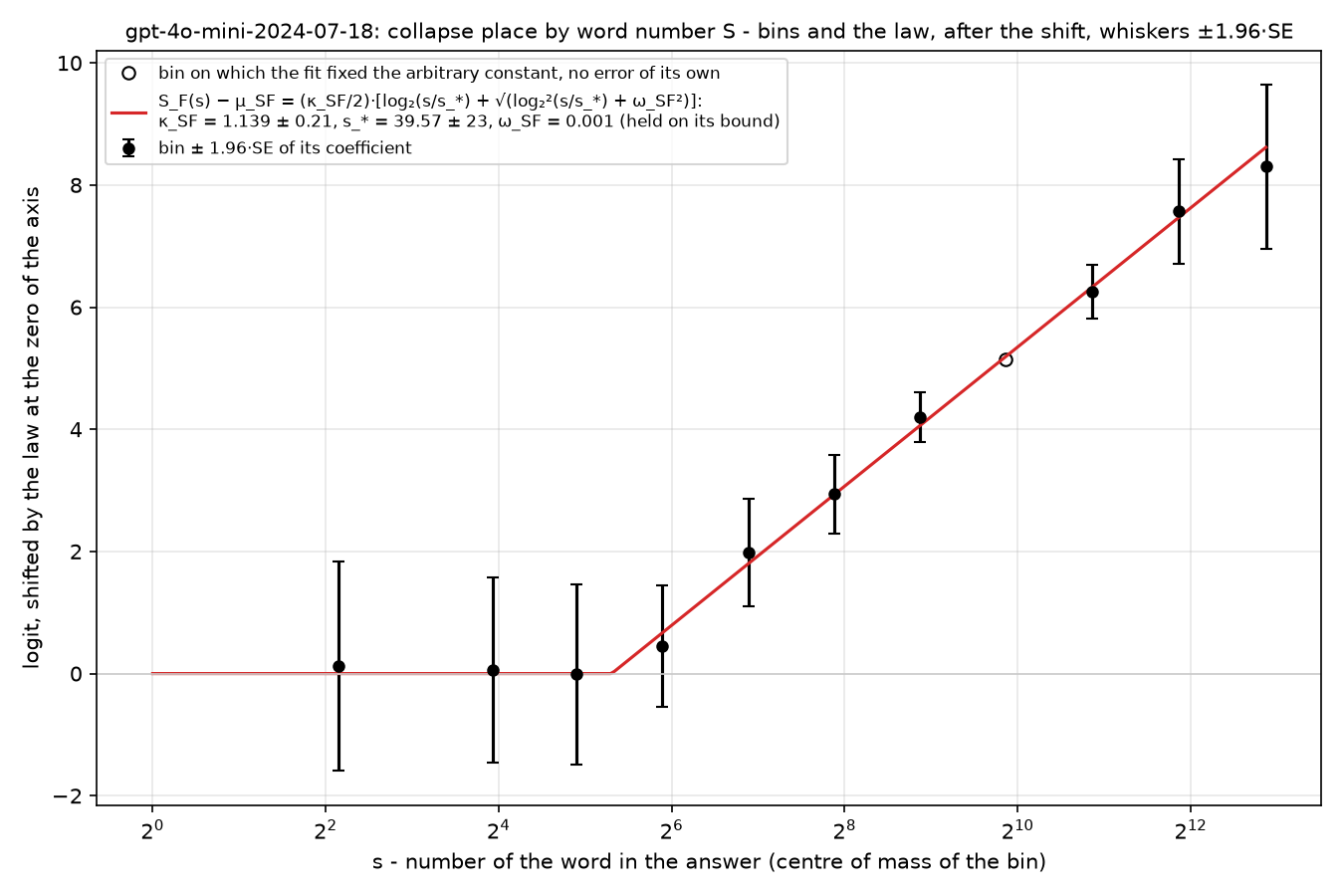}
\caption{Word number bins and the word number law after the shift correction: $\mu_{SF}$ is
subtracted from all bins. The notation is the same as in Figure~\ref{fig:place_s_raw}.}
\label{fig:place_s_corrected}
\end{figure}

\begin{figure}[htbp]
\centering
\includegraphics[width=0.85\linewidth]{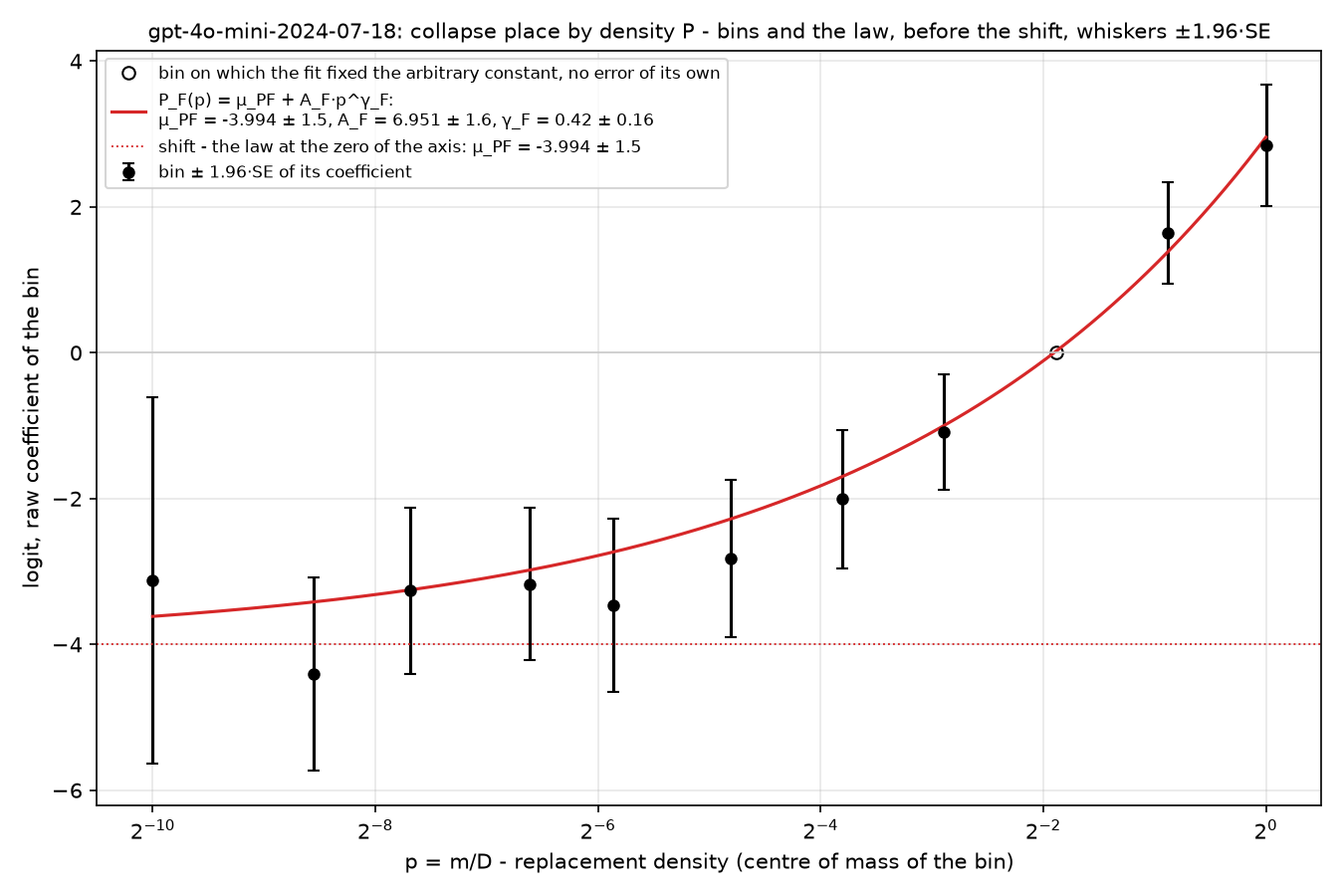}
\caption{Coefficients of the replacement density bins in the logit of the place of the collapse and
the density law before the shift correction. The notation is the same as in
Figure~\ref{fig:place_s_raw}; the dashed line is the level of the shift
$\mu_{PF}$.}
\label{fig:place_p_raw}
\end{figure}

\begin{figure}[htbp]
\centering
\includegraphics[width=0.85\linewidth]{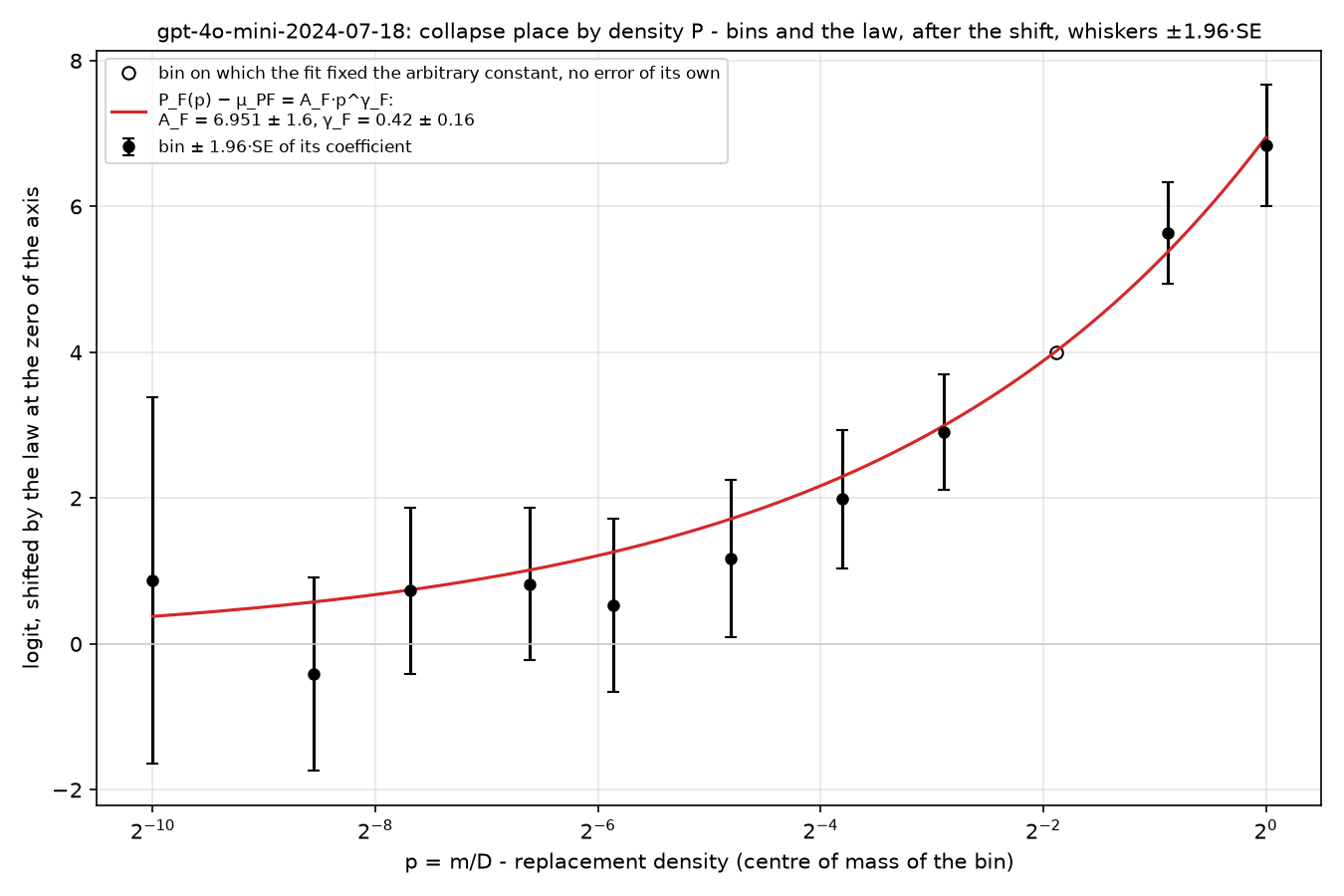}
\caption{Replacement density bins and the density law after the shift correction: $\mu_{PF}$ is
subtracted from all bins. The notation is the same as in Figure~\ref{fig:place_s_raw}.}
\label{fig:place_p_corrected}
\end{figure}

\begin{figure}[htbp]
\centering
\includegraphics[width=0.85\linewidth]{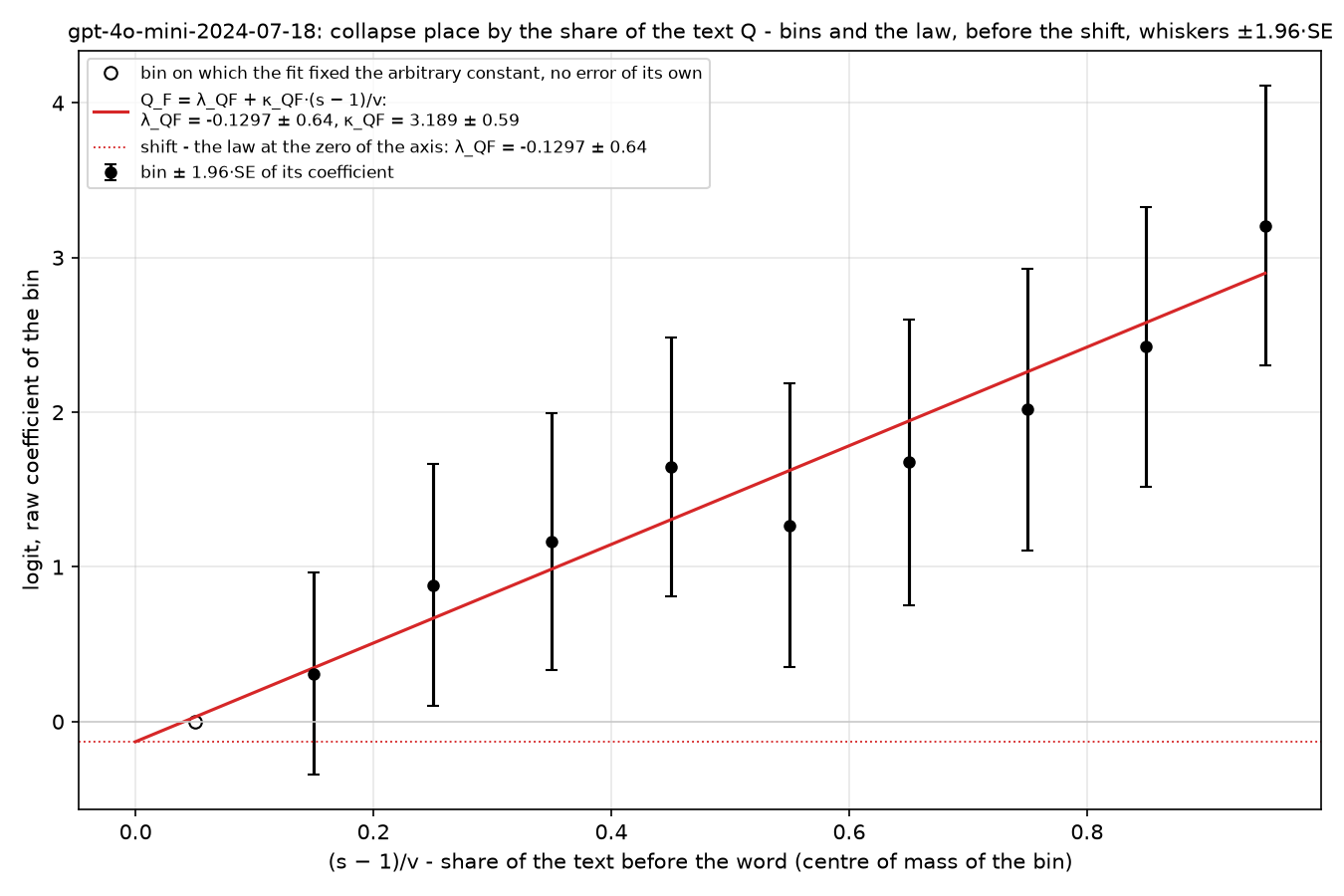}
\caption{Coefficients of the bins of the share of the text passed before the word in the logit of the
place of the collapse and the line of the text share before the shift correction. The notation is the
same as in Figure~\ref{fig:place_s_raw}; the dashed line is the level of the shift $\lambda_{QF}$.}
\label{fig:place_q_raw}
\end{figure}

\begin{figure}[htbp]
\centering
\includegraphics[width=0.85\linewidth]{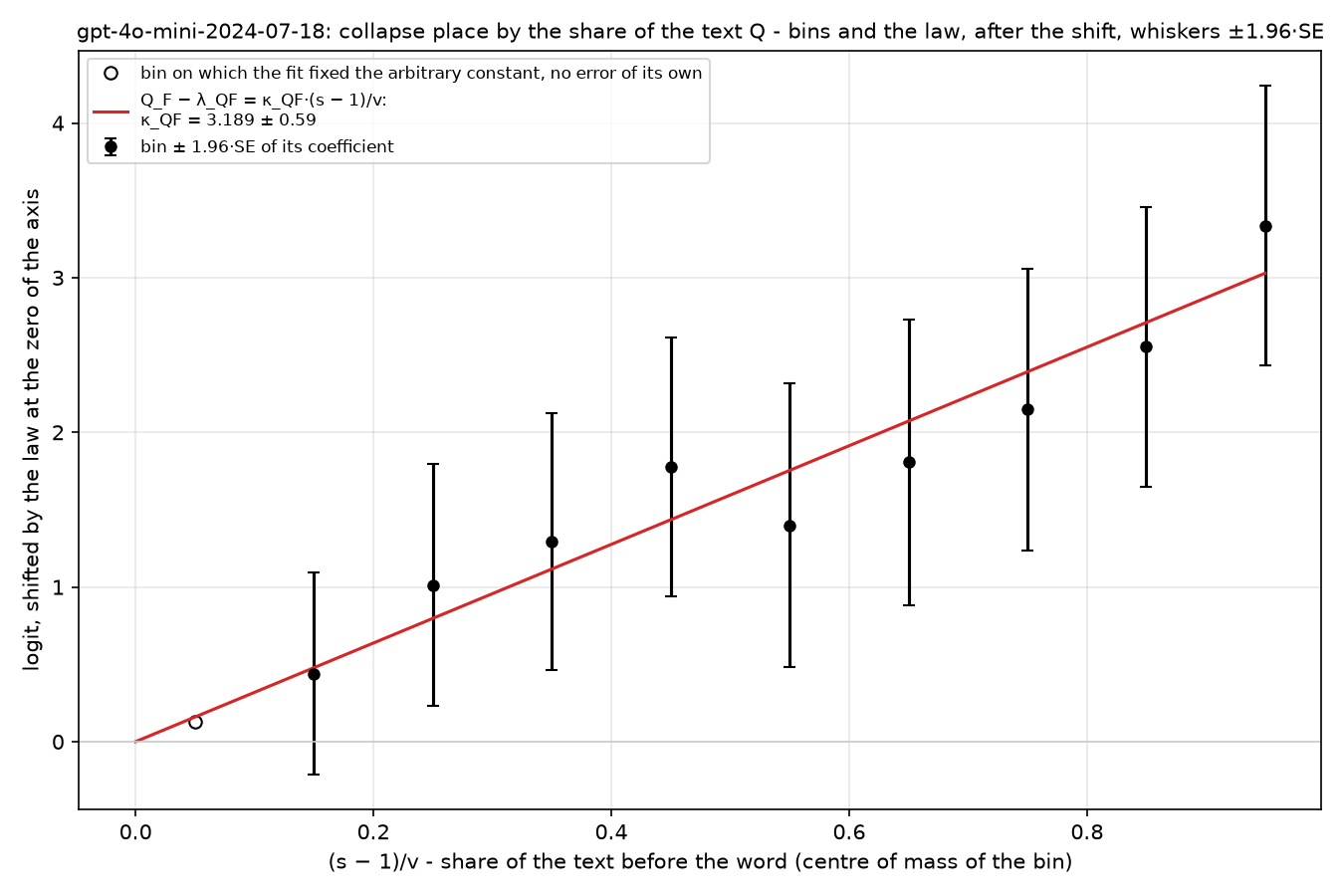}
\caption{Text share bins and the line of the text share after the shift correction: $\lambda_{QF}$ is
subtracted from all bins. The notation is the same as in Figure~\ref{fig:place_s_raw}.}
\label{fig:place_q_corrected}
\end{figure}

The coefficients of the bins and the laws of the axes are shown before the shift correction in
Figures~\ref{fig:place_s_raw}, \ref{fig:place_p_raw} and~\ref{fig:place_q_raw}, and after the
correction in Figures~\ref{fig:place_s_corrected}, \ref{fig:place_p_corrected}
and~\ref{fig:place_q_corrected}. Before the corner, around the 40th word, that is, on the first four
lines of the answer, the profile of the word number lies on the plateau, and after the corner each
doubling of the word number increases the logit of $F$ by $\kappa_{SF}$: the odds that the text is
already lost by a given word grow approximately in proportion to the word number to the power
$\kappa_{SF}/\ln 2 \approx 1.6$. The contribution of the density is about 0.5 logits at the density
1/512 and about 2.2 logits at the density 1/16 and reaches $A_F = 7.0$ logits at full
density: in a dense text the model loses the text earlier. The line of the text share passes through
the beginning of the text---its shift does not differ from zero---and by the end of the text adds
$\kappa_{QF}$ to the logit of $F$.

The text length and the number of rules do not serve as axes of the place of the collapse. If bins of
the text length are added to the model, the test of the hypothesis that they are all zero by the
cluster covariance gives a Wald statistic of
13.9 with 7 degrees of freedom, $p = 0.053$: beyond the word number and the passed share of the text,
the length does not determine the place of the collapse. The bins of the number of rules added to the
model differ from zero ($p < 0.001$), but in one way only: tasks with one rule lose the text earlier
than the others at the same word number, density and text share, while the bins from two to 128 rules
lie 0.5--1.9 logits below the bin of one rule with no trend in the number of rules. The profiles of
the word number, the density and the text share change by no more than 1.1 of their uncertainty when
the bins of the number of rules are added, so the law of the place of the collapse does not depend on
this check.

The law of the place of the collapse answers the question left by the law of the rate
(Section~\ref{sec:collapse_rate}): does the risk of losing the text on the next line grow as the
answer progresses or with the volume of the whole text. The place of the loss is determined by the
word number and the passed share of the text, while the text length beyond them does not determine
it. Consequently, the risk grows as the answer progresses: the further the model has advanced in the
answer, the more likely it is that it has already lost the text. Apparently, in a long text the
collapse is more frequent primarily because the answer to it is longer.

The plateau at the beginning of the answer admits an interpretation within the copying mechanism
proposed for the collapse rate. At the beginning of the answer the place in the text is set by the
very beginning of the input: the first lines of the answer repeat the first lines of the text, and the
model has almost nothing to confuse them with. As the answer progresses, the place in the text is held
only by the coincidence of the last words of the answer with a fragment of the input, and each next
line becomes a new opportunity to lose this coincidence. The density acts on the place of the collapse
in the same direction as on its rate: in a dense text there are more replacement words breaking the
coincidence of the answer with the input, and the text is lost earlier. The instrument measures the
form of the dependence, but not its cause, and the interpretation proposed requires a separate test.

Thus, for the requests that lost the text, up to the 40th word of the answer the contribution of the
word number to the odds that the text is already lost is constant, and after the 40th word these odds
grow with the word number; in addition, they grow with the passed share of the text and with the
replacement density. The risk of losing the text is determined by the progress through the answer,
not by the volume of the text as such.

\subsection{The passport}\label{sec:passport}

\subsubsection{Instruction capacity}\label{sec:passport_iec}

Since in the VS-PMR model the rules are represented by two variables, the meaning of the instruction
capacity changes as well. The rules of a task are described by two axes: the number of rules~$m$ in
the table and the number~$r$ of the checked rule. An added rule loads the language model twice: the
table becomes one line longer, which is accounted for by the axis of the number of rules, and the new
line is executed at its place in the list, which is accounted for by the axis of the rule number.
Therefore the capacity in the passport is not the number of rules at which the probability of a right
replacement in an individual trial falls to~$q$, but the number of rules for which the reserve of the
level suffices if each rule is paid for along both axes.

The price of the rule with number~$r$ is the decrease of the logit brought by adding this rule to the
table:
\[
a(r) = -\Delta M(r) - \widetilde{R}(r),
\]
where $\Delta M(r)$ is the change of the contribution of the number of rules when the table grows from
$r - 1$ to $r$ rules, and $\widetilde{R}(r)$ is the contribution of the rule number~$r$. The
cumulative price of a table of~$m$ rules
\[
C(m) = \sum_{r=2}^{m} a(r) = -\widetilde{M}(m) - \sum_{r=2}^{m} \widetilde{R}(r)
\]
consists of the price of the size of the table and the prices of executing its lines: the sum of the
increments of the contribution of the number of rules from the second rule to the $m$-th equals the
contribution of the number of rules in a table of~$m$ rules. The first rule is present in any task
and therefore has no price. The reserve
$\Delta_q = \alpha_{\mathrm{phys}} - \eta^*(q, \sigma^2)$ is counted from the reference level at the
zero point, where the grid and the density are taken at their horizontal asymptotes
(Section~\ref{sec:bias}), and the prices of rules are taken from the profiles of the number of rules
and of the rule number after the shift correction, which vanish at the first rule. The instruction
capacity is the number of rules at which the cumulative price exhausts the reserve:
\[
C(\mathrm{IEC}_q) = \Delta_q.
\]

By the bins the contribution of the number of rules is given stepwise: within a bin it is constant,
and at the boundary of bins it changes by a jump. In the cumulative sum the term of such a
contribution would be zero when a rule is added within a bin and equal to the whole jump at its
boundary. Therefore the axis of the number of rules enters the price by an increment per rule: the
jump between the corrected coefficients of neighboring bins is distributed equally over the rules of
the bin,
\[
\Delta M(r) = \frac{\widetilde{M}_k - \widetilde{M}_{k-1}}{|B_k|}, \qquad r \in B_k,
\]
where $\widetilde{M}_k$ is the corrected coefficient of the $k$-th bin of the number of rules, and
$|B_k|$ is the number of rules in this bin. The sum of the increments over a bin equals the jump, so
this notation does not change the cumulative price at the boundaries of bins. The axis of the rule
number enters by the corrected coefficient of the bin into which the rule number falls:
$\widetilde{R}(r) = \widetilde{R}_k$ at $r \in B_k$. The capacity by the bins is computed as in
Section~\ref{sec:quantities}:
\[
\mathrm{IEC}_q = m_0 + \frac{\Delta_q - C(m_0)}{a(m_0 + 1)}, \qquad
m_0 = \max\{\, m : C(m) \le \Delta_q \,\}.
\]

By the smooth laws the prices are given by the laws of the number of rules and of the rule number
after the shift correction (Sections~\ref{sec:laws_m} and~\ref{sec:laws_r}):
$\widetilde{M}(m) = \kappa_M \log_2 m$ and $\widetilde{R}(r) = -A_R\bigl(r^{\gamma} - 1\bigr)$. The
sum over the rules is replaced by an integral from~1 to~$m$, and the cumulative price takes the form
\[
C(m) = -\kappa_M \log_2 m + A_R\Bigl[\frac{m^{\gamma + 1} - 1}{\gamma + 1} - (m - 1)\Bigr].
\]
The capacity is the root of the equation $C(m) = \Delta_q$, which is found numerically. The
uncertainties of both estimates are computed by the delta method from the full covariance of the fit,
extended by the variance of the shift: $\sigma^2$ enters the capacity through the threshold
$\eta^*(q, \sigma^2)$ (Section~\ref{sec:quantities}).

For GPT-4o mini the reference level is $\alpha_{\mathrm{phys}} = 18.81 \pm 1.44$, and the variance of
the shift of a request is $\sigma^2 = 5.90 \pm 0.71$. The spread of request difficulty raises the
threshold of the level 0.95 from $\ln 19 \approx 2.94$ to $\eta^* = 4.97$, and the reserve of this
level is $\Delta_{0.95} = 13.84$ logits. The threshold of the level 0.5 is zero, and the whole
reference level serves as its reserve, $\Delta_{0.5} = 18.81$. By the bins the reserve of the level
0.95 is exhausted at the 37th rule, and
$\mathrm{IEC}_{0.95} = 36.58 \pm 5.41$; by the smooth laws
$\mathrm{IEC}_{0.95} = 38.06 \pm 6.13$. For the level 0.5 the capacity is
$43.90 \pm 6.62$ by the bins and $46.09 \pm 6.61$ by the laws. Both capacities lie within the
measured region, which extends to 128 rules, and the discrepancy between the two computations, 1.5
and 2.2 rules, is smaller than their uncertainties: the smooth laws preserve the measured capacity.

The capacity refers to the zero point---the beginning of the answer on a short text with sparse
replacements. A longer text, distant word positions and denser replacements spend the same reserve,
and under such conditions the language model holds fewer rules. The reserve itself is spent mainly on
executing the rules at their places in the list, not on the size of the table: by the laws, by the
37th rule the size of the table costs about 2.2 logits, and the rule numbers about 11 logits.

Thus at the zero point GPT-4o mini holds about 37--38 rules at a share of right replacements of 0.95
and about 44--46 rules at a share of 0.5, and the main part of the reserve is spent on executing the
late rules of the list.

\subsubsection{Fatigue capacity}\label{sec:passport_fatigue}

The fatigue execution capacity (Fatigue Execution Capacity, FEC) $\mathrm{FEC}_q$ is the capacity
along the word position
in the answer (Section~\ref{sec:quantities}): the word number at which the loss of the logit due to
progress through the answer exhausts the reserve of the level~$q$. This loss is set by the slanted
asymptote of the grid law $\tilde\lambda_S + \kappa_S \log_2 s$ (Section~\ref{sec:laws_vs}).

The grid law describes each slice by a hyperbola---a smooth transition between two regimes: the
plateau and fatigue, that is, the decrease of the logit with the word number. The hyperbolic
transition between two straight lines was proposed by Griffiths and Miller, and Ferreira and Zocchi
showed that such a hyperbola is the expectation of a broken line of two straight lines with fixed
slopes and intercepts over the distribution of a random transition point, namely a non-standard
$t$-distribution with two degrees of freedom \citep{ferreira2020smooth}. In this reading the
smoothness of the transition reflects the scatter of the transition point between requests with the
straight lines of both regimes common to them, and the law of the fatigue regime is the slanted
asymptote. It is one for all slices and does not depend on the volume of the text.

The fatigue capacity is counted, like the instruction capacity, from the reference level
$\alpha_{\mathrm{phys}}$ (Section~\ref{sec:passport_iec}), and the word number~$\mathrm{FEC}_q$ is
found from the condition
\[
\alpha_{\mathrm{phys}} + \tilde\lambda_S + \kappa_S \log_2 \mathrm{FEC}_q = \eta^*(q, \sigma^2),
\]
whence
\[
\log_2 \mathrm{FEC}_q = \frac{\eta^*(q, \sigma^2) - \alpha_{\mathrm{phys}} - \tilde\lambda_S}{\kappa_S}.
\]
The uncertainty is computed in the same way as the uncertainty of the instruction capacity.

For GPT-4o mini the slope of the asymptote is $\kappa_S = -2.67 \pm 0.31$ logits per doubling of the
position, and its value at the first word after the shift correction is
$\tilde\lambda_S = 17.39 \pm 3.37$. The reserve of the level 0.95, $\Delta_{0.95} = 13.84$ logits, is
exhausted at the word $\mathrm{FEC}_{0.95} = 3277 \pm 571$, within the measured region, whose texts
extend to 10\,240~words. The reserve of the level 0.5, $\Delta_{0.5} = 18.81$, is exhausted at the
word $\mathrm{FEC}_{0.5} = 11\,890 \pm 1839$, beyond the edge of the measured region, and this
estimate is obtained by extrapolating the
law.

The reference level corresponds to the zero point, so the fatigue capacity describes the decrease of
the quality of replacements with the word number in the answer without a load by replacement rules.
Thus GPT-4o
mini keeps the share of right replacements at 0.95 over approximately 3300 words of the answer, and
at 0.5, by extrapolation of the law, over approximately 12 thousand words.

\subsubsection{Guaranteed instruction capacity}\label{sec:passport_igc}

The guaranteed instruction capacity (Instructional Guaranteed Capacity, IGC) $\mathrm{IGC}_q$
is the instruction capacity at full replacement density $p = 1$, when every word of the text is
replaced; the other axes are taken at the zero point. At $p = 1$ every rule comes with its own word of
the text, $D = m$, and as rules are added the density stays equal to one. Therefore the contribution
of the density does not depend on the number of rules and enters the cumulative price of the table
once, not the price of each rule:
\[
C_P(m) = -\widetilde{M}(m) - \sum_{r=2}^{m} \widetilde{R}(r) - \widetilde{P}(1),
\]
where $\widetilde{P}(1)$ is the contribution of the density at $p = 1$ after the shift correction
(Section~\ref{sec:laws_p}). The capacity is defined by the same condition as the instruction
capacity:
\[
C_P(\mathrm{IGC}_q) = \Delta_q, \qquad \Delta_q = \alpha_{\mathrm{phys}} - \eta^*(q, \sigma^2).
\]
The contribution of the density decreases as $p$ grows, and at full density the loss is largest, so
$\mathrm{IGC}_q$ serves as a lower bound of the instruction capacity over the replacement density.
The capacity exists if the reserve of the level is larger than this loss,
$\Delta_q > -\widetilde{P}(1)$; otherwise the language model does not hold the level~$q$ already with
one rule.

By the bins the contribution of the density is the corrected coefficient of the last density bin~$K$,
which contains only full density: $\widetilde{P}_K = \theta_{PK} - \delta_P$. This bin belongs to the
density axis and is common to all requests, so the computation by the bins here, as for the
instruction capacity, does not depend on the law. The cumulative price and the capacity by the bins
are
\[
C_P(m) = \sum_{r=2}^{m} a(r) - \widetilde{P}_K, \qquad
\mathrm{IGC}_q = m_0 + \frac{\Delta_q - C_P(m_0)}{a(m_0 + 1)}, \qquad
m_0 = \max\{\, m : C_P(m) \le \Delta_q \,\},
\]
where the price of a rule
\[
a(r) = -\Delta M(r) - \widetilde{R}(r), \qquad
\Delta M(r) = \frac{\widetilde{M}_k - \widetilde{M}_{k-1}}{|B_k|}, \qquad
\widetilde{R}(r) = \widetilde{R}_k, \qquad r \in B_k,
\]
is the same as in Section~\ref{sec:passport_iec}. By the smooth laws the contribution of the density
is $\widetilde{P}(1) = -A_P\bigl(1 - e^{-1/p_c}\bigr)$, and the cumulative price takes the form
\[
C_P(m) = -\kappa_M \log_2 m + A_R\Bigl[\frac{m^{\gamma + 1} - 1}{\gamma + 1} - (m - 1)\Bigr]
  + A_P\bigl(1 - e^{-1/p_c}\bigr);
\]
the capacity is the root of the equation $C_P(m) = \Delta_q$, which is found numerically. The
uncertainties of both estimates are computed in the same way as the uncertainties of the instruction
capacity and include the uncertainties of the parameters of the density law.

For GPT-4o mini the corrected coefficient of the full density bin is
$\widetilde{P}_K = -9.22 \pm 1.04$, and the loss by the law is
$A_P\bigl(1 - e^{-1/p_c}\bigr) = 9.35 \pm 1.01$ logits. For the level 0.95
$\mathrm{IGC}_{0.95} = 18.69 \pm 3.62$ by the bins and $17.55 \pm 4.04$ by the laws, for the
level 0.5 $29.51 \pm 5.92$ and $29.76 \pm 5.11$. The discrepancy between the two computations, 1.1
and 0.3 rules, is smaller than their uncertainties, and both capacities lie within the measured
region, which extends to 128 rules.

After deducting the loss from full density, the reserve of the level 0.95 decreases threefold, from
13.84 to 4.6 logits by the bins and 4.5 by the laws, while the capacity decreases only twofold. The
reason lies in the form of the cumulative price: late rules of the list cost more than early ones, the
cumulative price grows faster than the number of rules, and the remaining reserve is spent on the
first, cheapest rules.

Thus at full replacement density GPT-4o mini holds about 18--19 rules at a share of right
replacements of 0.95 and about 30 rules at a share of 0.5, that is, half as many and a third fewer
than at the zero point.

\subsubsection{Summary of the passport}\label{sec:passport_summary}

After the shift corrections the statistical model of the passport is written as
\[
\eta = \alpha_{\mathrm{phys}} + \widetilde{VS}(v, s) + \widetilde{P}(p) + \widetilde{M}(m)
  + \widetilde{R}(r) + b, \qquad
\alpha_{\mathrm{phys}} = \delta_{VS} + \delta_P + \delta_M + \delta_R,
\]
where $b \sim N(0, \sigma^2)$ is the shift of a request (Sections~\ref{sec:bernoulli}
and~\ref{sec:bias}).

The grid law is computed in two steps (Section~\ref{sec:laws_vs}). In the first step each text length
slice is described by a hyperbola with its own plateau level~$\tilde\mu_v$ and the parameters
$\kappa_S$, $\tilde\lambda_S$ and~$\omega_S$ common to all slices:
\[
\widetilde{VS}_v(s) = \frac{1}{2}\Bigl[\tilde\lambda_S + \kappa_S \log_2 s + \tilde\mu_v
  - \sqrt{(\tilde\lambda_S + \kappa_S \log_2 s - \tilde\mu_v)^2 + \kappa_S^2\,\omega_S^2}\,\Bigr].
\]
In the second step the plateau law is drawn through the plateau levels of the slices, and
$\tilde\mu_v = \tilde\mu(v)$:
\[
\tilde\mu(v) = \frac{\kappa_V}{2}\Bigl[\log_2\frac{v}{v_*}
  + \sqrt{\log_2^2\frac{v}{v_*} + \omega_V^2}\,\Bigr].
\]
The laws of the replacement density, the number of rules and the rule number
(Sections~\ref{sec:laws_p}--\ref{sec:laws_r}) have the form
\[
\widetilde{P}(p) = -A_P\bigl(1 - e^{-p/p_c}\bigr), \qquad
\widetilde{M}(m) = \kappa_M \log_2 m, \qquad
\widetilde{R}(r) = -A_R\bigl(r^{\gamma} - 1\bigr).
\]

The laws of the collapse of the answer describe the logit of the probability of the collapse~$\psi$
(Section~\ref{sec:collapse_rate}) and the logit of the probability~$F$ that a request that lost the
text has lost it by the word with number~$s$ (Section~\ref{sec:collapse_place}):
\[
\ln\frac{\psi}{1 - \psi} = \lambda_\psi + \kappa_\psi \log_2\frac{v}{20} + A_\psi\, p^{\gamma_\psi},
\]
\[
\ln\frac{F}{1 - F} = \lambda_F + \frac{\kappa_{SF}}{2}\Bigl[\log_2\frac{s}{s_*}
  + \sqrt{\log_2^2\frac{s}{s_*} + \omega_{SF}^2}\,\Bigr] + A_F\, p^{\gamma_F}
  + \kappa_{QF}\,\frac{s - 1}{v}.
\]

The coefficients of the laws of the model are given in Table~\ref{tab:passport_laws}, the
coefficients of the collapse laws in Table~\ref{tab:collapse_laws}. The capacities computed at the
reserves of the level $\Delta_{0.95} = 13.84$ and $\Delta_{0.5} = 18.81$ logits are collected in
Table~\ref{tab:capacities}.

\begin{table}[htbp]
\centering
\small
\caption{Coefficients of the laws of the VS-PMR model for GPT-4o mini after the shift correction.
The uncertainty is the half-width of the 95\% confidence interval.}\label{tab:passport_laws}
\begin{tabular}{@{}p{2.4cm}p{2.3cm}p{2.8cm}p{7.3cm}@{}}
\toprule
Block & Coefficient & Value & Meaning \\
\midrule
Grid & $\kappa_S$ & $-2.67 \pm 0.31$ & slope of the slanted asymptote, logits per doubling of the
  word position \\
 & $\tilde\lambda_S$ & $17.39 \pm 3.37$ & value of the slanted asymptote at the first word of the
  answer \\
 & $\omega_S$ & $2.23 \pm 0.53$ & width of the transition from the plateau to the slanted asymptote,
  doublings of the position \\
 & $\kappa_V$ & $-1.60 \pm 0.43$ & slope of the slanted asymptote of the plateau, logits per doubling
  of the text length \\
 & $v_*$ & 274 (146--517) & text length at which the asymptotes of the plateau intersect, words \\
 & $\omega_V$ & --- & width of the transition of the plateau; not determined by the data \\
\midrule
Density & $A_P$ & $12.07 \pm 2.63$ & limit of the loss of the logit as the replacement density
  grows \\
 & $p_c$ & $0.67 \pm 0.27$ & characteristic density: at $p = p_c$ the loss is
  63\% of~$A_P$ \\
\midrule
Number of rules & $\kappa_M$ & $-0.43 \pm 0.10$ & change of the logit when the number of rules
  doubles \\
\midrule
Rule number & $A_R$ & $0.041 \pm 0.040$ & scale of the degradation of late rules, logits \\
 & $\gamma$ & $0.74 \pm 0.20$ & exponent of the growth of degradation: a doubling of the number
  increases it by a factor of $2^{\gamma}$ \\
\midrule
Level & $\delta_{VS}$ & $13.86 \pm 1.37$ & shift of the grid---the horizontal asymptote of the
  plateau~$\mu_0$ \\
 & $\delta_P$ & $4.45 \pm 0.70$ & shift of the density---the law at $p \to 0$ \\
 & $\delta_M$ & $0.44 \pm 0.58$ & shift of the number of rules---the law at $m = 1$ \\
 & $\delta_R$ & $0.07 \pm 0.13$ & shift of the rule number---the law at $r = 1$ \\
 & $\alpha_{\mathrm{phys}}$ & $18.81 \pm 1.44$ & reference level---the logit at the zero point, the
  sum of the shifts \\
 & $\sigma^2$ & $5.90 \pm 0.71$ & variance of the shift of a request \\
\bottomrule
\end{tabular}
\end{table}

\begin{table}[htbp]
\centering
\small
\caption{Coefficients of the laws of the collapse of the answer for GPT-4o mini. The uncertainty is
the half-width of the 95\% confidence interval.}\label{tab:collapse_laws}
\begin{tabular}{@{}p{2.4cm}p{2.3cm}p{2.8cm}p{7.3cm}@{}}
\toprule
Law & Coefficient & Value & Meaning \\
\midrule
Collapse rate & $\lambda_\psi$ & $-20.00 \pm 1.58$ & logit of the collapse at a text length of
  20~words and $p \to 0$ \\
 & $\kappa_\psi$ & $2.40 \pm 0.19$ & slope in logits per doubling of the text length \\
 & $A_\psi$ & $12.25 \pm 1.01$ & contribution of the density to the logit of the collapse at
  $p = 1$ \\
 & $\gamma_\psi$ & $0.71 \pm 0.09$ & exponent of the contribution of the density \\
\midrule
Place of the collapse & $\lambda_F$ & $-10.65 \pm 1.99$ & logit~$F$ at the beginning of the answer at
  $p \to 0$ \\
 & $\kappa_{SF}$ & $1.14 \pm 0.21$ & slope of the slanted asymptote, logits per doubling of the word
  number \\
 & $s_*$ & about 40 (22--70) & word number at which the asymptotes intersect \\
 & $\omega_{SF}$ & --- & width of the corner; not determined by the data \\
 & $A_F$ & $6.95 \pm 1.56$ & contribution of the density to the logit~$F$ at $p = 1$ \\
 & $\gamma_F$ & $0.42 \pm 0.16$ & exponent of the contribution of the density \\
 & $\kappa_{QF}$ & $3.19 \pm 0.59$ & contribution of the passed share of the text by the end of the
  text \\
\bottomrule
\end{tabular}
\end{table}

\begin{table}[htbp]
\centering
\small
\caption{Capacities of GPT-4o mini: the instruction capacity $\mathrm{IEC}_q$, the guaranteed
instruction capacity $\mathrm{IGC}_q$ and the fatigue capacity $\mathrm{FEC}_q$. The uncertainty is
the half-width of the 95\% confidence interval.}\label{tab:capacities}
\begin{tabular}{@{}lccccl@{}}
\toprule
Capacity & $q$ & By bins & By laws & Unit & Condition \\
\midrule
$\mathrm{IEC}_q$ & 0.95 & $36.58 \pm 5.41$ & $38.06 \pm 6.13$ & rules & zero point \\
 & 0.5 & $43.90 \pm 6.62$ & $46.09 \pm 6.61$ & rules & zero point \\
$\mathrm{IGC}_q$ & 0.95 & $18.69 \pm 3.62$ & $17.55 \pm 4.04$ & rules & $p = 1$ \\
 & 0.5 & $29.51 \pm 5.92$ & $29.76 \pm 5.11$ & rules & $p = 1$ \\
$\mathrm{FEC}_q$ & 0.95 & --- & $3277 \pm 571$ & words & grid asymptote \\
 & 0.5 & --- & $11\,890 \pm 1839$ & words & grid asymptote, extrapolation \\
\bottomrule
\end{tabular}
\end{table}

%% file: sections/06-outlook.tex
% !TEX root = ../main.tex
\section{Outlook}\label{sec:outlook}

\subsection{Passports of existing models}\label{sec:outlook-passports}

The nearest direction of research is a comparative study of the behavior of language models under
load on the widest possible sample of models of different classes and manufacturers. The main
interest lies not only in the quantitative estimates of the instruction capacity, but also in the
regularities of the degradation of the quality of instruction execution: the composition of the
significant load factors, the character of their action and the mathematical form of the
corresponding dependencies. The instruction capacity is only one of the characteristics derived
from these laws, whereas the laws themselves give a fuller description of the behavior of a model
under load.

There are no grounds to assume that the regularities established for GPT-4o mini are universal.
Larger models or models of other architectures may show qualitatively different dependencies: the
appearance of extrema instead of a monotonic change, the replacement of a linear dependence on the
logarithm of the load by a power one, or a change of the very set of significant axes. A systematic
comparison of such regularities across different models and their generations will make it possible
to study the possible connection between the character of the degradation and the features of the
internal structure of the models. Thus a catalog of passports has a double value: as an
experimental basis of a comparative study of language models and as a tool for the quantitative
evaluation of their capabilities.

The proposed methodology provides the flexibility needed for such studies: the composition of the
load axes and the form of the dependencies are determined from experimental data rather than set a
priori (Sections~\ref{sec:analysis} and~\ref{sec:predictor_axes}). The main practical limitations in
building the catalog remain the availability of tokenizers and the cost of measurements.

\paragraph{Access to tokenizers.}
A replacement remains an atomic operation only if both words of the rule are encoded by one token of
the tokenizer of the measured model (Section~\ref{sec:vocabulary}). A common vocabulary for all
models is not needed: the single-token property is required only for the tokenizer of the model
under study, and if the vocabulary of each model is selected by the same rules, the comparability of
measurements is ensured by averaging over random sets of words and rules. The difficulty lies
elsewhere: not all manufacturers publish their tokenizers; for example, Anthropic does not publish
the tokenizer of the Claude models. For such models the single-token property of a word has to be
checked indirectly, through the token count provided by the provider, if it provides one: the number
of tokens of a word is the difference between the counts of the request with the word and without
it.

\paragraph{Number of requests and tokens.}
The cost of a measurement is determined primarily by the limit of the answer length of the
model~$W$. The answer of the instrument repeats the text of the task in full, so the limit of the
answer length restricts the text length and, through it, all three parameters of the task: the
number of distinct words~$D$, the number of repetitions~$n$ and, since $m \le D$, the number of
rules~$m$. Doubling the limit of the answer length allows each of the three parameters to be
doubled, that is, it adds one bin to each axis of the load cube $D \times n \times m$. If an axis
has $k$ bins, the number of cells of the cube grows from $k^3$ to $(k+1)^3$, and at the same average
number of requests per cell the number of requests of a measurement grows in the same proportion.

The estimate rests on the measurement of GPT-4o mini with an answer limit of 16\,384 tokens
(Section~\ref{sec:measurements}). The axes $D$, $n$ and $m$ have 9, 8 and 9 bins in it. Without the
restriction of the vocabulary, the number of requests at an answer limit of
$W = 2^j \cdot 16$~thousand tokens is
\[
N(j) \approx N_0\, \frac{(9 + j)(8 + j)(9 + j)}{9 \cdot 8 \cdot 9}, \qquad N_0 = 2958.
\]
The number of tokens grows faster: the tokens of a request are proportional to the text length, and
the average text length in the cube is proportional to its limit, so each doubling of the limit
also doubles the average tokens of a request:
\[
T(j) \approx T_0\, 2^{j}\, \frac{(9 + j)(8 + j)(9 + j)}{9 \cdot 8 \cdot 9}, \qquad
T_0 = 15.94~\text{M}.
\]
The estimates by these formulas are given in Table~\ref{tab:outlook_volume}.

\begin{table}[htbp]
\centering
\small
\caption{Estimate of the volume of a measurement by the limit of the answer length of the model,
without the restriction of the vocabulary. The reference row is the measurement of GPT-4o
mini.}\label{tab:outlook_volume}
\begin{tabular}{@{}lrrr@{}}
\toprule
Answer limit, thousand tokens & Requests & Token multiplier & Tokens, million \\
\midrule
16 & 2958 & 1.0 & 16 \\
32 & $\approx 4100$ & 2.8 & $\approx 44$ \\
64 & $\approx 5500$ & 7.5 & $\approx 119$ \\
128 & $\approx 7200$ & 19.6 & $\approx 312$ \\
256 & $\approx 9300$ & 50.1 & $\approx 800$ \\
\bottomrule
\end{tabular}
\end{table}

\paragraph{Time and cost.}
The generation time is proportional to the number of output tokens, so at the same speed of the
model and the same number of parallel calls the time of a measurement grows with the same multiplier
as the number of tokens: the measurement of GPT-4o mini took 5~h 43~min, and at an answer limit of
128~thousand tokens it would take about 112~h. The cost of a measurement is
$C = T_{\text{in}}\, p_{\text{in}} + T_{\text{out}}\, p_{\text{out}}$, where $p$ is the price of the
model per million input and output tokens. The prices of large models are higher, and this
multiplies the cost beyond the growth of the number of tokens. The cost of the measurement of GPT-4o
mini at the provider's price is \$6.03; the estimates for the models available in October 2026 are
given in Table~\ref{tab:outlook_cost}.

\begin{table}[htbp]
\centering
\small
\caption{Estimate of the cost and time of a measurement by the formulas of this subsection. Prices
are in dollars per million input and output tokens, without discounts; the time is at the
generation speed and parallelism of the measurement of GPT-4o mini. An asterisk marks answer limits
not confirmed by official documentation.}\label{tab:outlook_cost}
\begin{tabular}{@{}lrrrr@{}}
\toprule
Model & Answer limit & Price & Cost, USD & Time \\
\midrule
GPT-4o mini (measurement) & 16 thousand & 0.15 / 0.60 & 6 & 5.7 h \\
gpt-4.1-mini & 32 thousand & 0.40 / 1.60 & $\approx 45$ & $\approx 16$ h \\
gpt-4.1 & 32 thousand & 2 / 8 & $\approx 220$ & $\approx 16$ h \\
Gemini 2.5 Flash & 64 thousand & 0.30 / 2.50 & $\approx 170$ & $\approx 43$ h \\
Gemini 2.5 Pro & 64 thousand & 1.25 / 10 & $\approx 680$ & $\approx 43$ h \\
Gemini 3.1 Pro & 64 thousand* & 2 / 12 & $\approx 840$ & $\approx 43$ h \\
Claude Haiku 5.5 & 128 thousand & 0.10 / 0.50 & $\approx 95$ & $\approx 112$ h \\
GPT-6 Luna & 128 thousand* & 0.10 / 0.50 & $\approx 95$ & $\approx 112$ h \\
gpt-5.4-mini & 128 thousand & 0.75 / 4.50 & $\approx 830$ & $\approx 112$ h \\
Claude Sonnet 5.5 & 128 thousand & 2 / 10 & $\approx 1900$ & $\approx 112$ h \\
GPT-6 Sol & 128 thousand* & 2 / 10 & $\approx 1900$ & $\approx 112$ h \\
Claude Opus 5.5 & 128 thousand & 4 / 20 & $\approx 3800$ & $\approx 112$ h \\
Claude Fable 5.1 & 128 thousand & 10 / 50 & $\approx 9500$ & $\approx 112$ h \\
GPT-6 Astra & 128 thousand* & 10 / 50 & $\approx 9500$ & $\approx 112$ h \\
\bottomrule
\end{tabular}
\end{table}

The data on the maximum answer length and the prices of the Claude models are given according to
the Anthropic documentation\footnote{\url{https://platform.claude.com/docs/en/about-claude/models/overview},
25.10.2026.}, and the prices of the other models according to price summaries for October
2026\footnote{\url{https://www.morphllm.com/llm-api},
\url{https://mem0.ai/library/llms-and-models/openai-api-pricing}, 25.10.2026.}. Models are appearing
that allow the generation of up to a million tokens in one answer. Thus, for Gemini 4 Argon with the
stated limit, a full measurement by the existing methodology would require about 5~billion tokens,
which is approximately 300 times the volume of the measurement of GPT-4o mini. These estimates are
approximate and are upper bounds, since they do not take into account the restriction of the
vocabulary and the loss of the structure of the answer, which reduce the available region of the
load cube.

Such significant costs are characteristic primarily of the initial measurement of a model, whose
goal is to reveal the laws of degradation without prior assumptions about their form. This requires
exploring the whole load cube. Once the corresponding regularities are established, repeated
measurements, needed when the parameters of the model or the conditions of load change, can be
carried out selectively, within individual trajectories or the most significant regions of the
cube. Using the already known laws should make it possible to reduce the volume of measurements and
their cost by at least an order of magnitude. Developing a strategy for such measurements is a
separate task.

The possibility of passing from a full study of a model to substantially less costly measurements is
of fundamental importance for the planned weighing of prompts (Section~\ref{sec:outlook-prompts}).
The initial measurement provides the calibration of the meter, after which the instruction weight of
individual prompts is expected to be determined at costs amounting to only a small fraction of the
cost of the full calibration.

\subsection{Measuring prompts}\label{sec:outlook-prompts}

Prompts have become the main means of programming the behavior of language models. Their
development, complication and optimization have acquired the character of an independent
engineering activity, yet there is still no generally accepted quantitative measure of the
instruction load they create. The present work proposes to consider a prompt as an object of
metrological measurement and to express its weight in equivalents of elementary
instructions---replacement rules. The goal of such a measurement is not to evaluate the quality of
the prompt itself or the success of the task set in it, but to determine the magnitude of the
instruction load that it creates for a language model. The literature review
(Section~\ref{sec:related-work}) did not reveal studies in which the load of arbitrary prompts was
measured in common comparable units. The closest to this formulation is the observation that the
quality of instruction execution decreases when a parallel task is added
\citep{mittal2026prospective}, but the cited work does not introduce a quantitative measure of this
effect.

The instruction weight of a prompt differs fundamentally from its formal characteristics---the
length of the text and the number of instructions it contains. Many simple independent requirements
may create a smaller load than a few complex interrelated conditions. Therefore counting
instructions or tokens does not by itself make it possible to evaluate the demands of a prompt on the
capabilities of a language model. The proposed measurement aims to determine its actual instruction
load, expressed on the common scale of elementary replacement rules.

The proposed method of weighing prompts is based on an indirect measurement of the instruction load.
Instead of evaluating the success of executing the prompt under study, one measures its effect on
the quality of executing an independent control task with a reference answer known in advance and a
calibrated dependence on the load. The word replacement test serves as such a task. The prompt under
study is added to the task of the instrument next to the table of rules, after which the decrease of
the share of right replacements is measured. The laws of the passport of the model make it possible
to express the corresponding change of the logit through an equivalent number of additional
replacement rules, which defines the instruction weight of the prompt. The fundamental advantage of
the method lies in the possibility of measuring the load of arbitrary prompts without developing
special tests to evaluate the quality of executing the tasks they contain. The conversion into
equivalents of rules rests on the assumption that the additional load of a prompt acts on the logit
of a right replacement in the same way as an increase of the number of rules. The validity of this
representation is yet to be established experimentally.

The instruction capacity of a model and the weight of a prompt, expressed in the same units, form a
common scale for comparing the capabilities of a model and the demands of the task set before it.
This will make it possible to evaluate quantitatively not only the ability of a model to execute
instructions, but also the load created by a specific prompt, the available reserve of capacity and
its change when the prompt is optimized. Thereby the proposed approach opens the possibility of
passing from the experimental study of the limitations of language models to quantitative methods of
designing applied systems based on them.

In this formulation the passport of a model is not only an independent result of the study, but
also a necessary calibration basis of the future prompt meter. It is the established laws of the
dependence of the quality of instruction execution on the load that make it possible to convert the
decrease of the results of the control task into an equivalent number of rules. A full measurement
of a model (Section~\ref{sec:outlook-passports}) requires significant costs, but after the
calibration each subsequent weighing is expected to be carried out using the known laws, spending
only a small fraction of the resources needed for the initial measurement. Thus the developed
methodology creates the prerequisites not only for a systematic study of the instruction capacity of
language models, but also for creating a new class of measuring instruments intended for the
quantitative evaluation of the load of applied prompts.

%% file: sections/07-conclusion.tex
% !TEX root = ../main.tex
\section{Conclusion}\label{sec:conclusion}

This work proposes a way to measure the behavior of a language model under controlled instruction
load and builds an instrument that implements it. The load is formed by rules replacing single-token
words in a text without meaning: such an instruction is atomic, the rules are independent by the
construction of the task, and the outcome of each application of a rule is determined by a
computable reference answer without an evaluator (Section~\ref{sec:instrument}). One rule serves as
a conventional unit of load, and every occurrence of a word to be replaced becomes a Bernoulli trial
with a known position in the answer and a known rule number.

The statistical model describes the logit of a right replacement by the sum of the contributions of
the load axes and the shift of the request, and contains no common level that depends on the choice
of reference bins. The form of each contribution is first measured by bins without assumptions and
then described by a smooth law, and the approximation shift correction restores a reference level
that does not depend on the coding (Section~\ref{sec:analysis}). The instruction capacity is defined
as the solution of an inverse problem---the load at which the share of right replacements falls to
a given level of quality---and is expressed in rules with an uncertainty.

The measurement of GPT-4o mini showed that the form measured by bins without assumptions is regular
for each load factor and is described by a law with a small number of parameters
(Section~\ref{sec:results}). The text length and the word position act not as a sum of independent
losses, but as two interacting constraints: the length sets the level of quality at the beginning of
the answer, and the word position sets the regularity of its subsequent decrease, common to all
lengths. The replacement density acts as an independent saturating factor, and the size of the rule
list and the position of a rule in it act as two different factors from which the price of each
next rule is composed. The collapse of the answer obeys its own laws: it is determined by the load
of the text and the progress through the answer, not by the number of instructions held. From these
laws the instruction, fatigue and guaranteed instruction capacities of the model are computed
(Section~\ref{sec:passport}), and every number of the passport is recomputed from the published
points (Section~\ref{sec:availability}).

The laws found for GPT-4o mini describe one model, and the nearest task is to build the passports of
many models and to compare the very laws of their behavior (Section~\ref{sec:outlook-passports}).
The ultimate goal of the instrument is the weighing of prompts: the passport of a model serves as
the calibration of a meter that expresses the instruction load of an arbitrary prompt in the same
units as the capacity of the model (Section~\ref{sec:outlook-prompts}).

%% file: sections/08-availability.tex
% !TEX root = ../main.tex
\section{Code and data availability}\label{sec:availability}

{\setlength{\emergencystretch}{5em}%
The public part of the instrument is published in an open repository
\url{https://github.com/kutomanovhennadii/instruction-capacity}. It contains the code that
executes one point---generating a task from three seeds, calling the model and validating the
answer---the code that builds the passport and checks it, the vocabulary together with the code that
produced it, the documents of the method and the published measurement of GPT-4o mini: the records
of all points with prompts, answers and token counters, the description of the measurement with the
checksums of the files, and the passport.\par}

Any number of the passport given in the article is recomputed from the published points by the
command
\begin{verbatim}
python scripts/build_passport.py artefacts/pilot_gpt4o_mini_v34 \
    --out-dir my_passport
\end{verbatim}
and the integrity and reproducibility of the published measurement are checked by the command
\begin{verbatim}
python scripts/verify.py artefacts/pilot_gpt4o_mini_v34
\end{verbatim}
The check compares the checksums of the files, compares the vocabulary with its reference version,
regenerates the rule list and the text of each request from its seeds and recomputes the passport,
comparing it with the published one within the precision of the computations; it does not call the
language model. With the same code one can measure points on another model and build its passport.

The points of the published measurement were chosen by an adaptive planning technology developed by
the author: it decides which points to measure from the results of the points already measured. This
technology forms the closed part of the instrument and is not disclosed: neither its design nor its
rules for choosing points are published. The passport does not depend on it: it is computed from the
published points alone, whoever chose them, and therefore every number of the passport can be
checked without the closed part.

The code is distributed under the MIT license, the data and the documents under the Creative Commons
Attribution 4.0 International license.

%% file: sections/09-acknowledgments.tex
% !TEX root = ../main.tex
\section*{Acknowledgments and disclosure of AI use}

In developing the code of the instrument and the statistical model, in processing the data and in
preparing the text of the article, the author used the Claude models of Anthropic. The formulation
of the problem, the design of the instrument and of the method, and the decisions on the composition
and form of the results belong to the author; the author has verified and bears responsibility for
the entire content of the article, including the code and the numbers.

The author thanks OpenAI for its program of free daily usage for traffic shared with
OpenAI\footnote{\url{https://help.openai.com/en/articles/10306912}.}: organizations that allow their
API traffic to be shared with OpenAI receive up to 10~million free tokens per day for small models,
including gpt-4o-mini and gpt-5.4-mini, and up to 1~million for large ones. Thanks to this program,
all the tokens spent in the measurements were provided free of charge.

%% file: sections/90-appendix.tex
% !TEX root = ../main.tex
\section{Example of a request}\label{sec:appendix}

The published request at the point $(D, n, m) = (4, 5, 2)$, used as an example in
Section~\ref{sec:task}, is given below in full: a text of 20 words and two rules, the order seed is
zero, and the lines of the rules follow in the order of the choice. The prompt is assembled from the
template of protocol version~0.1.

Prompt:
\begin{verbatim}
Rewrite the text below by applying the substitution rules.

Rules (replace every occurrence of the left word with the right word):
james -> hang
reign -> steal

Strict requirements:
- Replace only the words listed in the rules.
- Do not change, add, or remove any other words.
- Keep the exact word order, spaces, and line breaks.
- Output only the rewritten text, nothing else.

Text:
reign cab reign luck cab luck reign reign cab reign
james luck james james luck james cab james cab luck
\end{verbatim}

Answer of GPT-4o mini:
\begin{verbatim}
steal cab steal luck cab luck steal steal cab steal
hang luck hang hang luck hang cab hang cab luck
\end{verbatim}

The request contains 10 trials---five occurrences of each of the two words to be replaced. The
structure of the answer is preserved, and all ten replacements are made correctly.